\documentclass[10pt,journal,compsoc]{IEEEtran}
\usepackage{graphicx}
\usepackage[compress]{cite} 
\usepackage{subfig}
\usepackage{url}
\usepackage{amsmath,amssymb,amsfonts}
\usepackage{colortbl}
\usepackage{xcolor}
\usepackage{bm}
\usepackage{varwidth}
\usepackage{multirow}
\usepackage{graphicx}
\usepackage{algorithm}
\usepackage{algorithmic}
\usepackage{verbatim}
\usepackage{upgreek}
\usepackage{longtable}
\usepackage{textcomp}
\usepackage{booktabs}
\usepackage{ragged2e}
\usepackage{ulem}

\newcommand\redsout{\bgroup\markoverwith{\textcolor{red}{\rule[0.5ex]{2pt}{0.4pt}}}\ULon}
\newcommand{\rev}[1]{{\textcolor{black}{#1}}}
\ifCLASSINFOpdf
\else
\fi
\begin{document}
%
\title{
A Survey on Self-supervised Learning: Algorithms, Applications, and Future Trends
}
%
%
%
%

\author{Jie~Gui,~\IEEEmembership{Senior Member,~IEEE,}
        Tuo~Chen,
        Jing~Zhang,~\IEEEmembership{Senior Member,~IEEE,}
        Qiong~Cao,
        Zhenan~Sun,~\IEEEmembership{Senior Member,~IEEE,}
        Hao~Luo,
        Dacheng~Tao,~\IEEEmembership{Fellow,~IEEE}
\IEEEcompsocitemizethanks{\IEEEcompsocthanksitem J. Gui is with the School of Cyber Science and Engineering, Southeast University and with Purple Mountain Laboratories, Nanjing 210000, China (e-mail: guijie@seu.edu.cn).}
\IEEEcompsocitemizethanks{\IEEEcompsocthanksitem T. Chen is with the School of Cyber Science and Engineering, Southeast University (e-mail: 230219309@seu.edu.cn).}
\IEEEcompsocitemizethanks{\IEEEcompsocthanksitem Jing Zhang is with the School of Computer Science, The University of Sydney, Camperdown, NSW 2050, Australia (e-mail: jing.zhang1@sydney.edu.au).}
\IEEEcompsocitemizethanks{\IEEEcompsocthanksitem D. Tao is with the College of Computing \& Data Science at Nanyang Technological University, \#32 Block N4 \#02a-014, 50 Nanyang Avenue, Singapore 639798 (email: dacheng.tao@gmail.com).}
\IEEEcompsocitemizethanks{\IEEEcompsocthanksitem Q. Cao is with JD Explore Academy (e-mail: mathqiong2012@gmail.com).}
\IEEEcompsocitemizethanks{\IEEEcompsocthanksitem Z. Sun is with the Center for Research on Intelligent Perception and Computing, Chinese Academy of Sciences, Beijing 100190, China (e-mail: znsun@nlpr.ia.ac.cn).}
\IEEEcompsocitemizethanks{\IEEEcompsocthanksitem H. Luo is with Alibaba Group, Hangzhou 310052, China (e-mail: michuan.lh@alibaba-inc.com).}

}

%
%

\markboth{Journal of \LaTeX\ Class Files,~Vol.~14, No.~8, August~2015}%
{Shell \MakeLowercase{\textit{et al.}}: Bare Demo of IEEEtran.cls for Computer Society Journals}
%



\IEEEtitleabstractindextext{%
\begin{abstract}
\justifying
Deep supervised learning algorithms typically require a large volume of labeled data to achieve satisfactory performance. However, the process of collecting and labeling such data can be expensive and time-consuming. Self-supervised learning (SSL), a subset of unsupervised learning, aims to learn discriminative features from unlabeled data without relying on human-annotated labels. SSL has garnered significant attention recently, leading to the development of numerous related algorithms. However, there is a dearth of comprehensive studies that elucidate the connections and evolution of different SSL variants. This paper presents a review of diverse SSL methods, encompassing algorithmic aspects, application domains, three key trends, and open research questions. Firstly, we provide a detailed introduction to the motivations behind most SSL algorithms and compare their commonalities and differences. Secondly, we explore representative applications of SSL in domains such as image processing, computer vision, and natural language processing. Lastly, we discuss the three primary trends observed in SSL research and highlight the open questions that remain. A curated collection of valuable resources can be accessed at \url{https://github.com/guijiejie/SSL}.
\end{abstract}

}

\maketitle

\IEEEdisplaynontitleabstractindextext

%
\IEEEpeerreviewmaketitle

\IEEEraisesectionheading{\section{Introduction}\label{sec:introduction}}

%
%
%
%
\IEEEPARstart{D}{eep} supervised learning algorithms have demonstrated impressive performance in various domains, including computer vision (CV) and natural language processing (NLP). To address this, models pre-trained on large-scale datasets like ImageNet~\cite{deng2009imagenet} are commonly employed as a starting point and subsequently fine-tuned for specific downstream tasks (Table \ref{tab:0}). This practice is motivated by two primary reasons. Firstly, the parameters acquired from large-scale datasets offer a favorable initialization, enabling faster convergence of models trained on other tasks \cite{radford2021learning}. Secondly, a network trained on a large-scale dataset has already learned discriminative features, which can be easily transferred to downstream tasks and mitigate the overfitting issue arising from limited training data in such tasks \cite{ericsson2021well,liu2021self}.

\begin{table*}[htbp]
\caption{Comparison between supervised and self-supervised pre-training and fine-tuning.}
    \resizebox{\linewidth}{!}{%
  \begin{tabular}{cccc}
  \hline
  Pre-training& Data & Pre-training Tasks                                  & Downstream Tasks                                      \\
   \hline
   \multirow{3}{*}{Supervised} & \multirow{3}{*}{extensive labeled data}   & \multirow{2}{*}{image categorization\cite{dosovitskiy2020image}}         & detection / segmentation / \\
   &      &                                                    & pose estimation / depth estimation, etc.\\
                                                              \cline{3-4}
   &      & video action categorization\cite{tran2015learning}                       & action recognition / object tracking, etc.\\
   \hline
   \multirow{5}{*}{SSL} & \multirow{5}{*}{extensive unlabeled data} & \multirow{2}{*}{Image: rotation \cite{gidaris2018unsupervised}, jigsaw \cite{noroozi2016unsupervised}, etc.} & detection / segmentation / \\
   &      &                                                    & pose estimation / depth estimation, etc.\\
                                                              \cline{3-4}
   &      & Video: the order of frames \cite{misra2016shuffle}, playing direction \cite{wei2018learning}, etc. & action recognition / object tracking, etc.\\
                                                              \cline{3-4} 
   &      & \multirow{2}{*}{NLP: masked language modeling\cite{devlin2018bert}}               & question answering / textual entailment recognition / \\
   &      &                                                    & natural language inference, etc.\\
   \hline
  \end{tabular}}
  \label{tab:0}
  \end{table*}
  
Unfortunately, numerous real-world data mining and machine learning applications face a common challenge where an abundance of unlabeled training instances coexists with a limited number of labeled ones. The acquisition of labeled examples is frequently costly, arduous, or time-consuming due to the requirement of skilled human annotators with sufficient domain expertise \cite{zeng2020realistic,miech2020end}. To illustrate, consider the analysis of web user profiles, where a substantial amount of data can be readily collected. However, the labeling of non-profitable or profitable users necessitates thorough scrutiny, judgment, and sometimes even time-intensive tracing tasks performed by experienced human assessors, resulting in significant expenses. Another instance pertains to the medical field, where unlabeled examples can be easily obtained through routine medical examinations. Nevertheless, assigning diagnoses individually to such a large number of cases places a substantial burden on medical experts. For example, in the case of breast cancer diagnosis, radiologists must label each focus in a vast collection of easily attainable, high-resolution mammograms. This process often proves to be highly inefficient and time-consuming. Additionally, supervised learning methods are susceptible to spurious correlations and generalization errors, and vulnerable to adversarial attacks.

\begin{figure}
  \begin{center}
  \scalebox{0.54}{\includegraphics{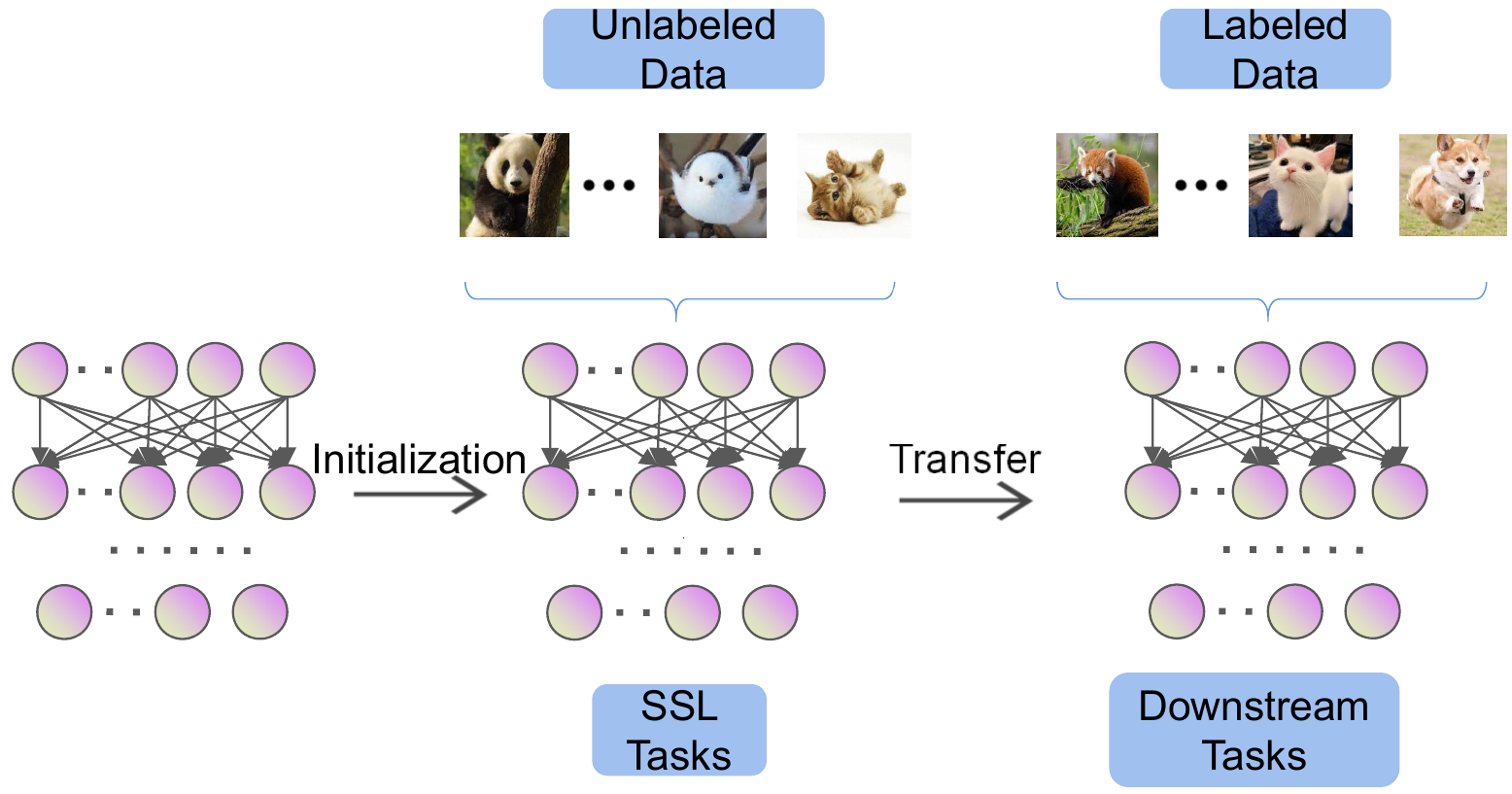}}
  \end{center}
     \caption{The general pipeline of applying SSL methods to downstream tasks. The SSL models are first pre-trained on the unlabeled data and then fine-tuned, or directly evaluated, on the labeled data of the downstream tasks.}
  \label{fig:1_1}
  \end{figure}

To address the aforementioned limitations of supervised learning, various machine learning paradigms have been introduced, including active learning, semi-supervised learning, and self-supervised learning (SSL). This paper specifically emphasizes SSL. SSL algorithms aim to learn discriminative features from vast quantities of unlabeled instances without relying on human annotations. The general pipeline of SSL is depicted in Fig. \ref{fig:1_1}. In the self-supervised pre-training phase, a pre-defined pretext task is formulated for the deep learning algorithm to solve. Pseudo-labels for the pretext task are automatically generated based on specific attributes of the input data. 
Once the self-supervised pre-training process is completed, the acquired model can be transferred to downstream tasks.

  \begin{figure}
    \begin{center}
    \scalebox{0.54}{\includegraphics{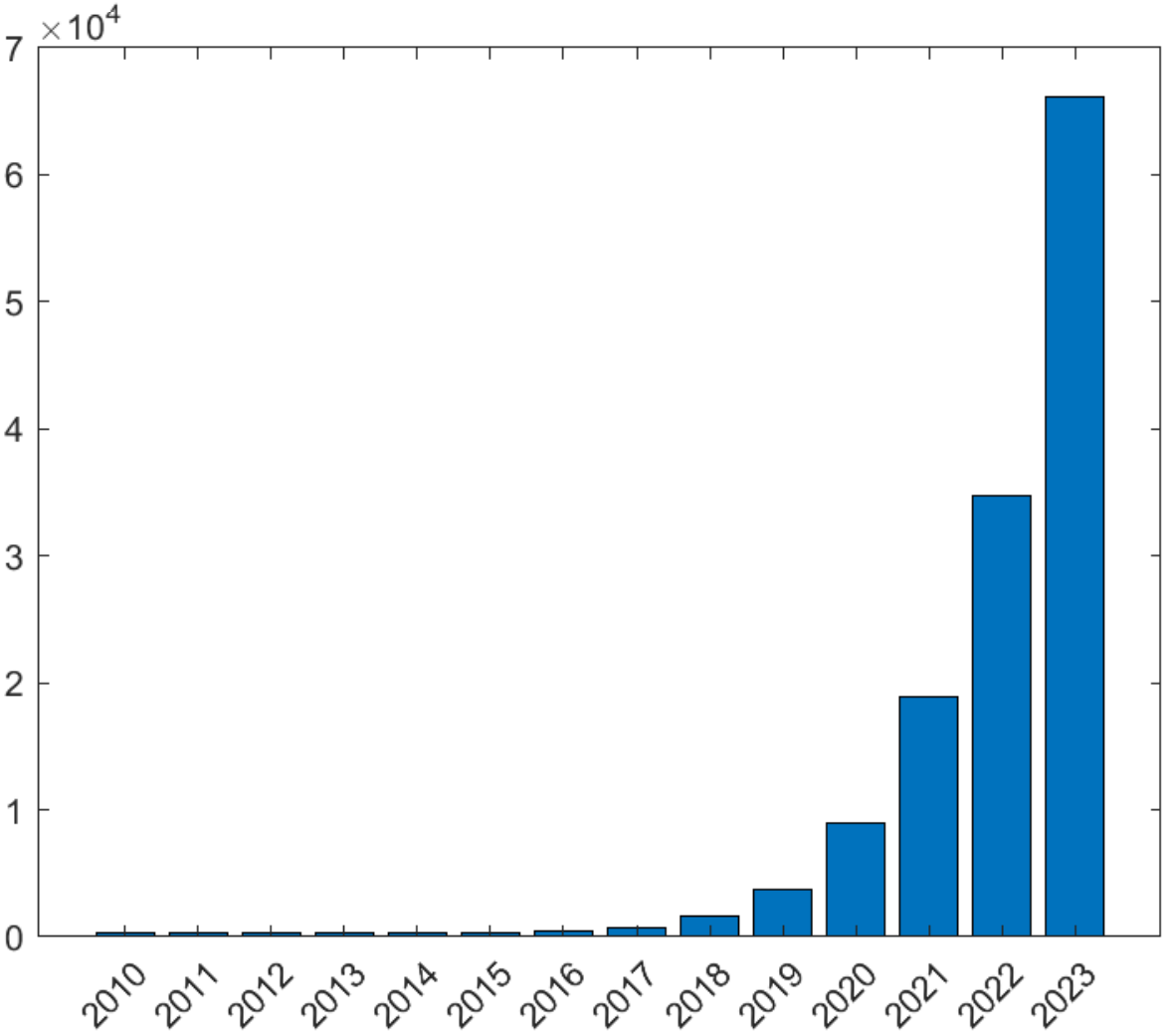}}
    \end{center}
       \caption{Google Scholar search results for ``self-supervised learning''. The vertical and horizontal axes denote the number of SSL publications and the year, respectively.}
    \label{fig:1}
    \end{figure}

One notable advantage of SSL algorithms is their ability to leverage extensive unlabeled data since the generation of pseudo-labels does not necessitate human annotations. By utilizing these pseudo-labels during training, self-supervised algorithms have demonstrated promising outcomes, resulting in a reduced performance disparity compared to supervised algorithms in downstream tasks. Asano et al. \cite{asano2019critical} demonstrated that SSL can produce generalizable features that exhibit robust generalization even when applied to a single image.

The advancement of SSL \cite{hinton2006reducing,vincent2008extracting,pinto2016supersizing,li2016unsupervised,li2016unsupervisedV,lee2019rethinking,zoph2020rethinking,orhan2020self,mitrovic2020representation,ericsson2021well,liu2021self,hua2021feature} has exhibited rapid progress, capturing significant attention within the research community (Fig. \ref{fig:1}), and is recognized as a crucial element for achieving human-level intelligence \cite{timmurphy.org}. Google Scholar reports a substantial volume of SSL-related publications, with approximately 18,900 papers published in 2021 alone. This accounts for an average of 52 papers per day or more than two papers per hour (Fig. \ref{fig:1}). To assist researchers in navigating this vast number of SSL papers and to consolidate the latest research findings, we aim to provide a timely and comprehensive survey on this subject.

\textbf{Differences from previous work}: Previous works have provided reviews on SSL that cater to specific applications such as recommender systems \cite{yu2022self}, graphs \cite{liu2022graph}, sequential transfer learning \cite{mao2020survey}, videos \cite{schiappa2022self}, and adversarial pre-training of self-supervised deep networks \cite{qi2022adversarial}. Besides, Liu et al. \cite{liu2021self} primarily focused on papers published before 2020, lacking the latest advancements. Jaiswal et al. \cite{jaiswal2020survey} centered their survey on contrastive learning (CL). Notably, recent breakthroughs in SSL research within the CV domain are of significant importance. Thus, this review predominantly encompasses recent SSL research derived from the CV community, particularly those influential and classic findings. The primary objectives of this review are to elucidate the concept of SSL, its categories and subcategories, its differentiation and relationship with other machine learning paradigms, as well as its theoretical foundations. We present an extensive and up-to-date review of the frontiers of visual SSL, dividing it into four key areas: context-based, CL, generative, and contrastive generative algorithms, aiming to outline prominent research trends for scholars.

\section{Algorithms}
This section begins by providing an introduction to SSL, followed by an explanation of the pretext tasks associated with SSL and their integration with other learning paradigms.
\subsection{What is SSL?}

\begin{figure}
\begin{center}
\scalebox{0.33}{\includegraphics{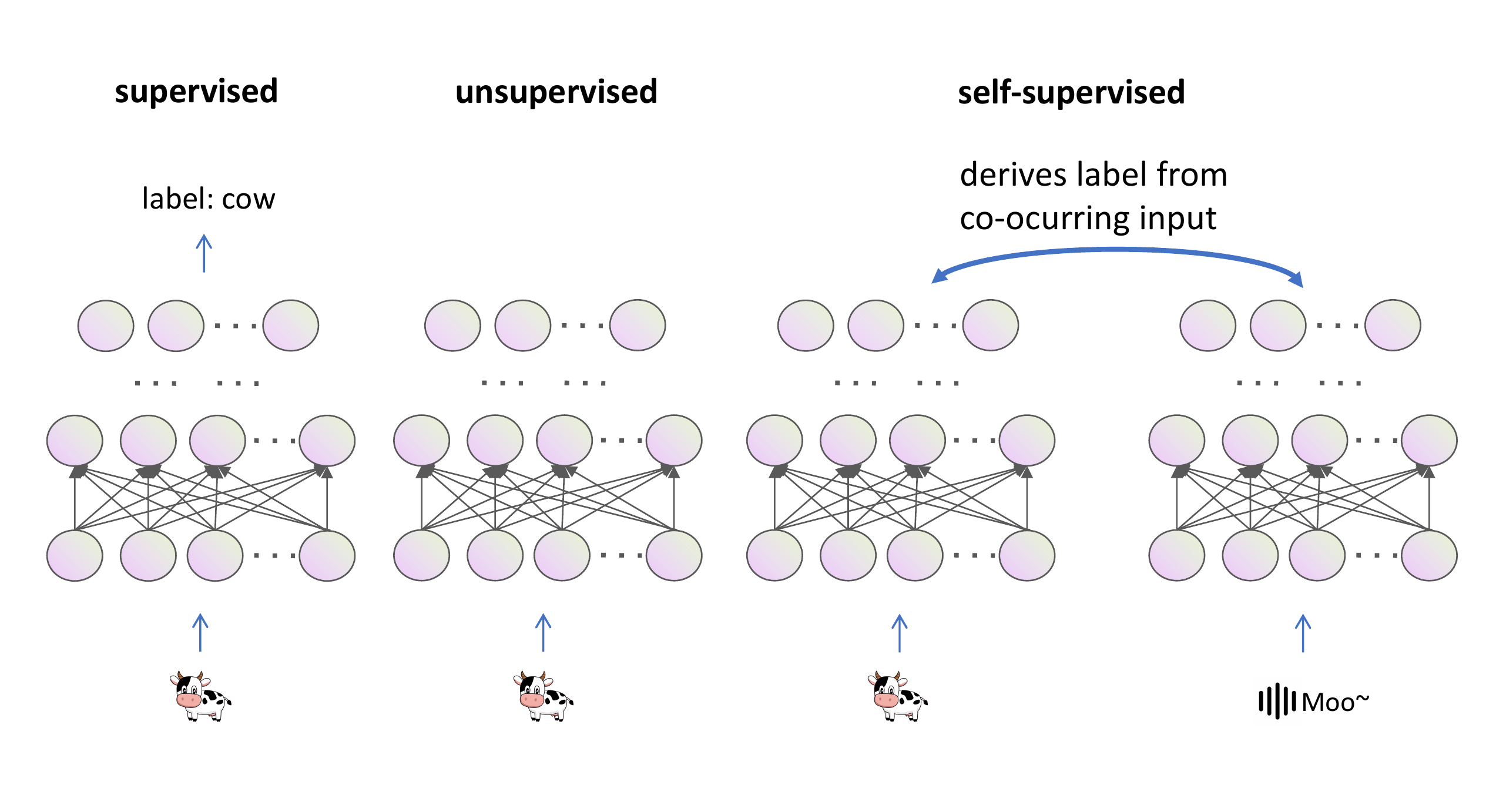}}
\end{center}
   \caption{The differences among supervised learning, unsupervised learning, and SSL. The image is reproduced from \cite{de1994learning}. SSL utilizes freely derived labels as supervision instead of manually annotated labels.}
\label{fig:2_1}
\end{figure}

The introduction of SSL is attributed to \cite{de1994learning} (Fig. \ref{fig:2_1}), who employed this architecture to learn in natural environments featuring diverse modalities. Although the cow image may not warrant a cow label, it is frequently associated with a ``moo'' sound. The crux lies in the co-occurrence relationship between them.

Subsequently, the machine learning community has advanced the concept of SSL, which falls within the realm of unsupervised learning. SSL involves generating output labels ``intrinsically'' from input data examples by revealing the relationships between data components or various views of the data. These output labels are derived directly from the data examples. According to this definition, an autoencoder (AE) can be perceived as a type of SSL algorithms, where the output labels correspond to the data itself. AEs have gained extensive usage across multiple domains, including dimensionality reduction and anomaly detection.

In the keynote talk at ICLR 2020~\cite{lecun2020reflections}, Yann LeCun elucidated the concept of SSL as an analogous process to completing missing information (reconstruction). He presented multiple variations as follows: 1) Predict any part of the input from any other part; 2) Predict the future from the past; 3) Predict the invisible from the visible; and 4) Predict any occluded, masked, or corrupted part from all available parts.
In summary, a portion of the input is unknown in SSL, and the objective is to predict that particular segment.

Jing et al. \cite{Jing2021Self-Supervised} expanded the definition of SSL to encompass methods that operate without human-annotated labels. Consequently, any approach devoid of such labels can be categorized under SSL, effectively equating SSL with unsupervised learning. This categorization includes generative adversarial networks (GANs) \cite{gui2020review}, thereby positioning them within the realm of SSL.

Pretext tasks, also referred to as surrogate or proxy tasks, are a fundamental concept in the field of SSL. The term ``pretext'' denotes that the task being solved is not the primary objective but serves as a means to generate a robust pre-trained model. Prominent examples of pretext tasks include rotation prediction and instance discrimination, among others. Each pretext task necessitates the use of distinct loss functions to achieve its intended goal. Given the significance of pretext tasks in SSL, we proceed to introduce them in further detail.

\subsection{Pretext tasks} \label{subsection_Pretext}
This section provides a comprehensive overview of the pretext tasks employed in SSL. A prevalent approach in SSL involves devising pretext tasks for networks to solve, where the networks are trained by optimizing the objective functions associated with these tasks. Pretext tasks typically exhibit two key characteristics. Firstly, deep learning methods are employed to learn features that facilitate the resolution of pretext tasks. Secondly, supervised signals are derived from the data itself, a process known as self-supervision. Commonly employed techniques encompass four categories of pretext tasks: context-based methods, CL, generative algorithms, and contrastive generative methods. In our paper, generative algorithms primarily refer to masked image modeling (MIM) methods.

\subsubsection{Context-based methods}
\begin{figure}
\begin{center}
\scalebox{0.58}{\includegraphics{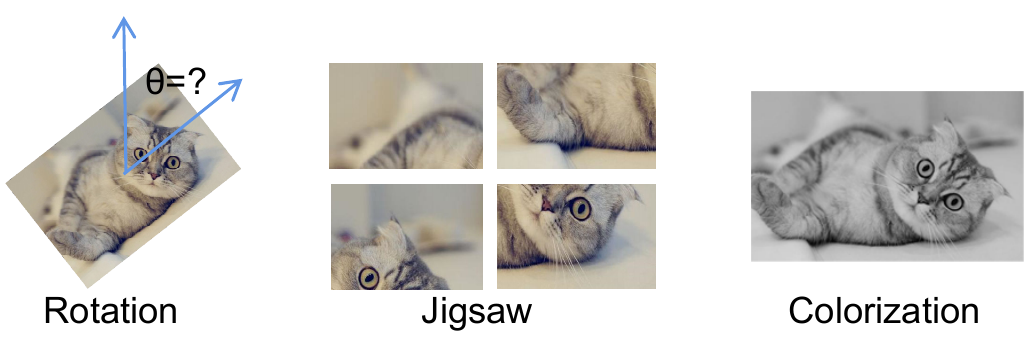}}
\end{center}
   \caption{Illustration of three common context-based methods: rotation, jigsaw, and colorization.}
\label{fig:2_2}
\end{figure}

Context-based methods rely on the inherent contextual relationships among the provided examples, encompassing aspects such as spatial structures and the preservation of both local and global consistency. We illustrate the concept of context-based pretext tasks using rotation as a simple example \cite{nathan2018improvements}. Subsequently, we progressively introduce additional tasks (Fig. \ref{fig:2_2}).

\textbf{Rotation}: Gidaris et al. \cite{gidaris2018unsupervised} trained deep neural networks (DNNs) to learn image representations by recognizing the random geometric transformations. They streamlined image augmentation by introducing rotations of $0^\circ$, $90^\circ$, $180^\circ$, and $270^\circ$ to generate three additional images from each original. This method employs rotation angles as self-supervised labels, using a set of $K=4$ geometric transformations $G = \{ g(\cdot | y) \}_{y = 1}^K$. Here, $g(\cdot | y)$ applies a geometric transformation labeled $y$ to an image $X$, resulting in a transformed image ${X^y} = g(X | y)$.

Gidaris et al. utilized a deep convolutional neural network (CNN), \(\mathcal{F}(\cdot)\), to perform rotation prediction through a four-class categorization task. This CNN processes an input image \(X^{y^*}\), with \(y^*\) being unknown to \(\mathcal{F}(\cdot)\), and outputs a probability distribution over possible geometric transformations, expressed as

  \begin{eqnarray}\label{equ:1}
    \mathcal{F}\left( {{X^{{y^*}}}|\theta } \right) = \left\{ {{\mathcal{F}^y}\left( {{X^{{y^*}}}|\theta } \right)} \right\}_{y = 1}^K.
    \end{eqnarray}
Here, $\mathcal{F}^y\left( {{X^{{y^*}}}|\theta } \right)$ represents the predicted probability for the geometric transformation labeled as $y$, while $\theta$ denotes the learnable parameters of \(\mathcal{F}(\cdot)\).

Given training instances $D = \{ {X_i}\} _{i = 1}^N$, the training objective can be formulated as
  \begin{eqnarray}\label{equ:2}
    \mathop {\min }\limits_\theta  \,\frac{1}{N}\sum\limits_{i = 1}^N {\mathcal{L}({X_i},\theta )}.
    \end{eqnarray}
  Here, the loss function is defined as
\begin{eqnarray}\label{equ:3}
\mathcal{L}({X_i},\theta ) = - \frac{1}{K}\sum\limits_{y = 1}^K {\log ({\mathcal{F}^y}\left( {g\left( {{X_i}|y} \right)|\theta } \right))}.
\end{eqnarray}

In \cite{agrawal2015learning}, the relative rotation angle was confined to the interval of $[-30^o, 30^o]$. These rotations were discretized into bins of $3^o$ each, leading to a total of 20 classes (or bins).

\textbf{Colorization}: The concept of colorization was initially introduced in \cite{zhang2016colorful}, and subsequent studies \cite{larsson2016learning,zhang2017real,larsson2017colorization} demonstrated its effectiveness as a pretext task for SSL. Color prediction offers the advantageous feature of requiring freely available training data. In this context, a model can utilize the lightness channel of any color image as input and utilize the corresponding $ab$ color channels in the CIE $Lab$ color space as self-supervised signals. The objective is to predict the $ab$ color channels $Y \in {R^{H \times W \times 2}}$ given an input lightness channel $X \in {R^{H \times W \times 1}}$
. A commonly employed learning objective is  
\begin{eqnarray}\label{equ:3_1}
    \mathcal{L} = \left\| {\hat Y - Y} \right\|_F^2,
    \end{eqnarray}
where $Y$ and $\hat Y$ denote the ground truth and predicted values, respectively.

Besides, \cite{zhang2016colorful} utilized the multinomial cross-entropy loss instead of \eqref{equ:3_1} to enhance robustness. Upon completing the training process, the $ab$ color channels would be predicted for any grayscale image. Consequently, the lightness channel and the $ab$ color channels can be concatenated to restore the original grayscale image to a colorful representation.

\textbf{Jigsaw}: The jigsaw approach leverages jigsaw puzzles as surrogate tasks, operating under the assumption that a model accomplishes these tasks by comprehending the contextual information embedded within the examples. Specifically, images are fragmented into discrete patches, and their positions are randomly rearranged, with the objective of reconstructing the original order. In \cite{goyal2019scaling}, the impact of scaling two self-supervised methods, namely jigsaw \cite{noroozi2016unsupervised,ahsan2019video} and colorization, was investigated along three dimensions: data size, model capacity, and problem complexity. The results indicated that transfer performance exhibits a log-linear growth pattern in relation to data size. Furthermore, representation quality was found to improve with higher-capacity models and increased problem complexity. 

\textbf{Others}: The pretext task employed in \cite{zhan2019self,wang20193d} involved a conditional motion propagation problem. To enforce a specific constraint on the feature representation process, Noroozi et al. \cite{noroozi2017representation} introduced an additional requirement where the sum of feature representations of all image patches should approximate the feature representation of the entire image. While many pretext tasks yield representations that exhibit covariance with image transformations, \cite{misra2020self} argued for the importance of semantic representations being invariant to such transformations. In response, they proposed a pretext-invariant representation learning approach that enables the learning of invariant representations through pretext tasks.

\subsubsection{Contrastive Learning}
\begin{figure*}
\begin{center}
\scalebox{0.54}{\includegraphics{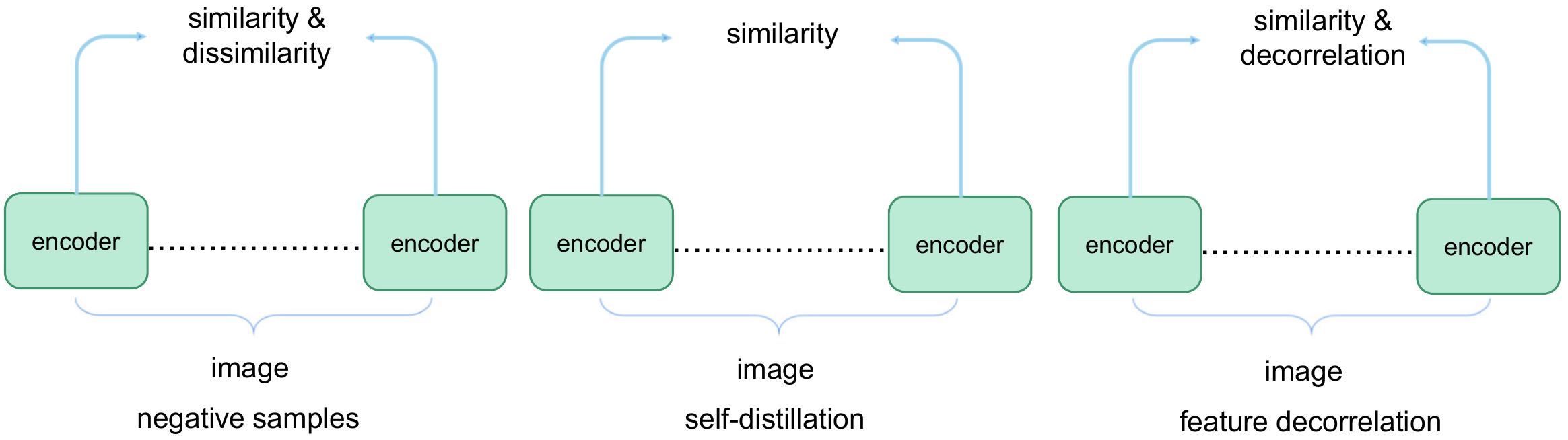}}
\end{center}
   \caption{Illustration of different CL methods: CL based on negative examples (left), CL based on self-distillation (middle), and CL based on feature decorrelation (right). For a demonstration of the concepts of similarity and dissimilarity, one can refer to \cite{wang2020understanding,chen2020simple}, while for insights into decorrelation, \cite{zbontar2021barlow,bardes2021vicreg} provide a comprehensive overview.}
\label{fig:3}
\end{figure*}

Numerous SSL methods based on CL have emerged, building upon the foundation of simple instance discrimination tasks \cite{wu2018unsupervised,zhao2020makes}. Notable examples include MoCo v1 \cite{he2020momentum}, MoCo v2 \cite{chen2020improved}, SimCLR v1 \cite{chen2020simple} and SimCLR v2 \cite{chen2020big}. Pioneering algorithms, such as MoCo, have significantly enhanced the performance of self-supervised pre-training, reaching a level comparable to that of supervised learning, thus rendering SSL highly pertinent for large-scale applications. Early CL approaches were built upon the concept of utilizing negative examples. However, as CL has progressed, a range of methods have emerged that eliminate the need for negative examples. These methods embrace distinct ideas such as self-distillation and feature decorrelation, yet all adhere to the principle of maintaining positive example consistency. The following section outlines the various CL methods currently available (Fig. \ref{fig:3}).

\paragraph{Negative example-based CL} 
Negative examples-based CL adheres to a pretext task known as instance discrimination, which involves generating distinct views of an instance. In negative examples-based CL, views originating from the same instance are treated as positive examples for an anchor sample, while views from different instances serve as negative examples. The underlying principle is to promote proximity between positive examples and maximize the separation between negative examples within the latent space. The definition of positive and negative examples varies depending on factors such as the modality being considered and specific requirements, including spatial and temporal consistency in video understanding or the co-occurrence of modalities in multi-modal learning scenarios. In the context of conventional 2D image CL, image augmentation techniques are utilized to generate diverse views from a single image.

\textbf{MoCo}: 
He et al. \cite{he2020momentum} framed CL as a dictionary look-up task. In this framework, a query $q$ exists and a set of encoded examples $\left\{ {{k_0},{k_1},{k_2}, \cdots } \right\}$ serve as the keys in a dictionary. Assuming a single key, denoted as ${k_ + }$ in the dictionary, matches the query $q$, a contrastive loss \cite{hadsell2006dimensionality} function is employed. The value of this function is low when $q$ is similar to its positive key ${k_ + }$ and dissimilar to all other negative keys. In the MoCo v1 \cite{he2020momentum} framework, the InfoNCE loss function \cite{oord2018representation}, a form of contrastive loss, is utilized, \textit{i.e.},
  \begin{eqnarray}\label{equ:4}
    {\mathcal{L}_q} = - \log \frac{{\exp ({{q \cdot {k_ + }} \mathord{\left/
    {\vphantom {{q \cdot {k_ + }} \tau }} \right.
    \kern-\nulldelimiterspace} \tau })}}{{\sum\nolimits_{i = 0}^K {\exp ({{q \cdot {k_i}} \mathord{\left/
    {\vphantom {{q \cdot {k_i}} \tau }} \right.
    \kern-\nulldelimiterspace} \tau })} }},
    \end{eqnarray}
where $\tau$ represents the temperature hyper-parameter and ($\cdot$) denotes vector product. The summation is computed over one positive example and $K$ negative examples. InfoNCE is derived from noise contrastive estimation (NCE) \cite{gutmann2010noise}.

MoCo v2 \cite{chen2020improved} builds upon MoCo v1 \cite{he2020momentum} and SimCLR v1 \cite{chen2020simple}, incorporating a multilayer perceptron (MLP) projection head and more data augmentations. 

\textbf{SimCLR}:
SimCLR v1 \cite{chen2020simple} employs a mini-batch sampling strategy with $N$ instances, wherein a contrastive prediction task is formulated on pairs of augmented instances from the mini-batch, generating a total of 2$N$ instances. Notably, SimCLR v1 does not explicitly select negative instances. Instead, for a given positive pair, the remaining 2$(N-1)$ augmented instances in the mini-batch are treated as negatives. Let $sim(u,v) = {{{u^T}v} \mathord{\left/
 {\vphantom {{{u^T}v} {\left( {\left\| u \right\|\left\| v \right\|} \right)}}} \right.
 \kern-\nulldelimiterspace} {\left( {\left\| u \right\|\left\| v \right\|} \right)}}$ represent the cosine similarity between two instances $u$ and $v$. The loss function of SimCLR v1 for a positive instance pair $(i,j)$ is defined as
  \begin{equation}\label{equ:5}
    {\mathcal{L}_{i,j}} = - \log \frac{{\exp ({{sim({z_i},{z_j})} / \tau })}}{{\sum\nolimits_{k = 1}^{2N} {{1_{\left[ {k \ne i} \right]}}\exp ({{sim({z_i},{z_k})} / \tau })} }},
    \end{equation}
where ${1_{\left[ {k \ne i} \right]}} \in \{ 0,1\}$ is an indicator function equal to 1 if $k \ne i$, and $\tau$ denotes the temperature hyper-parameter. The overall loss is computed across all positive pairs, including both $(i,j)$ and $(j,i)$, within the mini-batch.

In MoCo, the features generated by the momentum encoder are stored in a feature queue as negative examples. These negative examples do not undergo gradient updates during backpropagation. Conversely, SimCLR utilizes negative examples from the current mini-batch, and all of them are subjected to gradient updates during backpropagation. Both MoCo and SimCLR rely on data augmentation techniques, including cropping, resizing, and color distortion. Notably, SimCLR made a significant contribution by highlighting the crucial role of strong data augmentation in CL, a finding subsequently confirmed by MoCo v2. Additional augmentation methods have also been explored \cite{zheng2021ressl}. For instance, in \cite{zhao2020distilling}, foreground saliency levels were estimated in images, and augmentations were created by selectively copying and pasting image foregrounds onto diverse backgrounds, such as grayscale images with random grayscale levels, texture images, and ImageNet images. Furthermore, views can be derived from various sources, including different modalities such as photos and sounds \cite{arandjelovic2018objects}, as well as coherence among different image channels \cite{tian2019contrastive}.

Minimizing the contrastive loss is known to effectively maximize a lower bound of the mutual information $I(\mathbf{x}_1 ; \mathbf{x}_2)$ between the variables $\mathbf{x}_1$ and $\mathbf{x}_2$ \cite{oord2018representation}. Building upon this understanding, \cite{tian2020makes} proposes principles for designing diverse views based on information theory. These principles suggest that the views should aim to maximize $I(\mathbf{v}_1 ; \mathbf{y})$ and $I(\mathbf{v}_2 ; \mathbf{y})$ ($\mathbf{v}_1$, $\mathbf{v}_2$, and $\mathbf{y}$ denoting the first view, the second view, and the label, respectively), representing the amount of information contained about the task label, while simultaneously minimizing $I(\mathbf{v}_1 ; \mathbf{v}_2)$, indicating the shared information between inputs encompassing both task-relevant and irrelevant details. Consequently, the optimal data augmentation method is contingent on the specific downstream task. In the context of dense prediction tasks, \cite{xie2020propagate} introduces a novel approach for generating different views. This study reveals that commonly employed data augmentation methods, as utilized in SimCLR, are more suitable for categorization tasks rather than dense prediction tasks such as object detection and semantic segmentation. Consequently, the design of data augmentation methods tailored to specific downstream tasks has emerged as a significant area of exploration.

Given the observed benefits of strong data augmentation in enhancing CL performance \cite{chen2020simple}, there has been a growing interest in leveraging more robust augmentation techniques. However, it is worth noting that solely relying on strong data augmentation can actually lead to a decline in performance \cite{tian2020makes}. The distortions introduced by strong data augmentation can alter the image structure, resulting in a distribution that differs from that of weakly augmented images. This discrepancy poses optimization challenges. To address the overfitting issue arising from strong data augmentation, \cite{Wang2022Contrastive} proposes an alternative approach. Instead of employing a one-hot distribution, they suggest using the distribution generated by weak data augmentation as a mimic. This mitigates the negative impact of strong data augmentation by aligning the distribution of augmented examples with that of weakly augmented examples.

\paragraph{Self-distillation-based CL}
Bootstrap Your Own Latent (BYOL) \cite{grill2020bootstrap} is a prominent self-distillation algorithm designed specifically for self-supervised image representation learning, eliminating the need for negative pairs. BYOL employs two identical DNNs, known as Siamese networks, with the same architecture but different weights. One serves as the online network, while the other is the target network. Similar to MoCo \cite{he2020momentum}, BYOL enhances the target network through a gradual averaging of the online network. Siamese networks have emerged as prevalent architectures in contemporary self-supervised visual representation learning models, including SimCLR, BYOL, and SwAV \cite{caron2020unsupervised}. These models aim to maximize the similarity between two augmented versions of a single image while incorporating specific conditions to mitigate the risk of collapsing solutions.

Simple Siamese (SimSiam) networks, introduced by \cite{chen2020exploring}, offers a straightforward approach to learning effective representations in SSL without the need for negative example pairs, large batches, or momentum encoders. Given a data point $x$ and two randomly augmented views $x_1$ and $x_2$, an encoder $f$ and an MLP prediction head $h$ process these views. The resulting outputs are denoted as ${p_1} = h\left( {f\left( {{x_1}} \right)} \right)$ and ${z_2} = f\left( {{x_2}} \right)$. The objective of \cite{chen2020exploring} is to minimize their negative cosine similarity:
\begin{eqnarray}\label{equ:5_1}
D\left( {{p_1},{z_2}} \right) =  - \frac{{{p_1}}}{{{{\left\| {{p_1}} \right\|}_2}}}\frac{{{z_2}}}{{{{\left\| {{z_2}} \right\|}_2}}}.
 \end{eqnarray}
Here, ${\left\| {} \right\|_2}$ represents the $l_2$-norm. Similar to \cite{grill2020bootstrap}, a symmetric loss \cite{chen2020exploring} is defined as
  \begin{eqnarray}\label{equ:5_2}
    \mathcal{L} = \frac{1}{2}\left( {D\left( {{p_1},{z_2}} \right) + D\left( {{p_2},{z_1}} \right)} \right).
     \end{eqnarray}
This loss is defined based on the example $x$, and the overall loss is the average of all examples. Notably, \cite{chen2020exploring} employs a stop-gradient ($stopgrad$) operation by modifying Eq.~\eqref{equ:5_1} as
$D\left( {{p_1},stopgrad\left( {{z_2}} \right)} \right)$. This implies that $z_2$ is treated as a constant. Similarly, Eq.~\eqref{equ:5_2} is revised as
  \begin{eqnarray}\label{equ:5_4}
    \mathcal{L} = \frac{1}{2}\left( {D\left( {{p_1},stopgrad\left( {{z_2}} \right)} \right) + D\left( {{p_2},stopgrad\left( {{z_1}} \right)} \right)} \right).
    \end{eqnarray}

\begin{figure}
\begin{center}
\scalebox{0.5}{\includegraphics{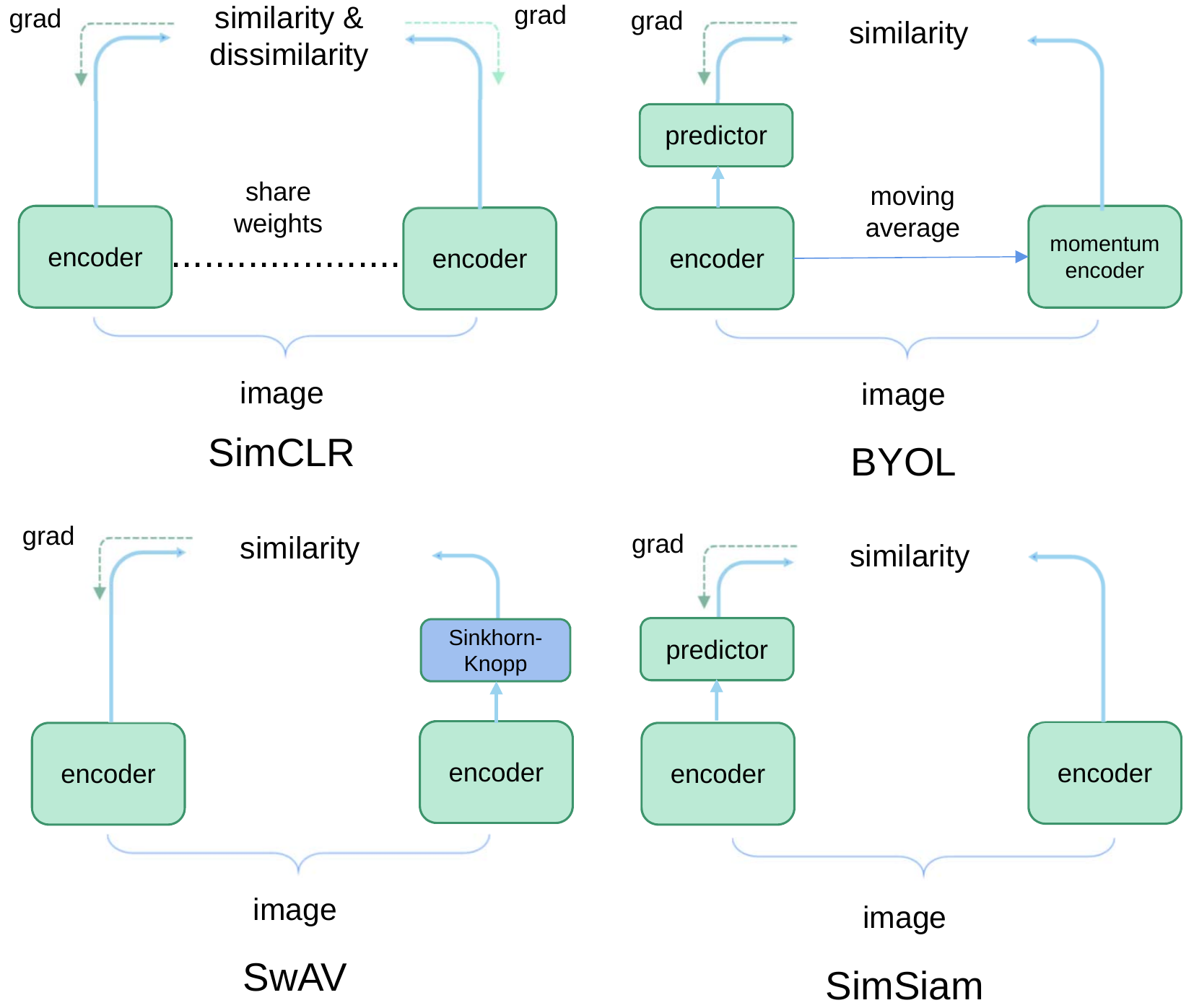}}
\end{center}
   \caption{Comparison among different Siamese architectures. The image is reproduced from \cite{chen2020exploring}.}
\label{fig:4}
\end{figure}

 Figure \ref{fig:4} illustrates the distinctions among SimCLR, BYOL, SwAV, and SimSiam. The categorization of BYOL and SimSiam as CL methods is a subject of debate due to their exclusion of negative examples. However, to be consistent with \cite{he2021masked}, this paper considers BYOL and SimSiam to belong to CL methods.

\paragraph{Feature decorrelation-based CL} 
The objective of feature decorrelation is to learn decorrelated features.

\textbf{Barlow Twins}:

Barlow Twins \cite{zbontar2021barlow} introduced a novel loss function that encourages the similarity of embedding vectors from distorted versions of an example while minimizing redundancy between their components. Similar to other SSL methods such as MoCo \cite{he2020momentum} and SimCLR \cite{chen2020simple}, Barlow Twins generates two distorted views $Y^{A}$ and $Y^{B}$ via a distribution of data augmentations $\mathcal{T}$ for each image in a data batch sampled from a dataset, 
resulting in batches of embeddings $Z^A$ and $Z^B$. The loss function of Barlow Twins is defined as
\begin{eqnarray}\label{equ:6}
  {\mathcal{L}_{BT}} = \sum\limits_i {{{\left( {1 - {C_{ii}}} \right)}^2}}  + \lambda \sum\limits_i {\sum\limits_{j \ne i} {C_{ij}^2} }.
  \end{eqnarray}
Here, $\lambda$ is a hyper-parameter, and $C$ represents the cross-correlation matrix computed between the two batches of embeddings $Z^A$ and $Z^B$, defined as

\begin{eqnarray}\label{equ:7}
{C_{ij}} = \frac{{\sum\nolimits_b {z_{b,i}^Az_{b,j}^B} }}{{\sqrt {\sum\nolimits_b {{{\left( {z_{b,i}^A} \right)}^2}} } \sqrt {\sum\nolimits_b {{{\left( {z_{b,j}^B} \right)}^2}} } }},
\end{eqnarray}
  where $b$ indexes batch samples and $i, j$ index the vector dimension of the networks' outputs. $C$ is a square matrix that measures the correlation between the two batches of embeddings $Z^A$ and $Z^B$. The first term in Eq. (\ref{equ:6}) encourages the diagonal elements of $C$ to be close to 1, while the second term encourages the off-diagonal elements to be close to 0.

\textbf{Variance-Invariance-Covariance Regularization}:
  Borrowing the covariance regularization from the Barlow Twins method, Variance-invariance-covariance regularization (VICReg) \cite{bardes2021vicreg} proposes a new self-supervised method for training joint embedding architectures that simultaneously considers variance, invariance, and covariance.
    Similar to Barlow Twins, VICReg generates two distorted views $Y^{A}$ and $Y^{B}$ via a distribution of the data augmentation $\mathcal{T}$ and gets their embeddings \(Z^A \in \mathbb{R}^{n \times d}\) and \(Z^B\in\mathbb{R}^{n \times d}\).
    Let the subscript $j$ index the embedding in the batch and $d$, $n$ represent the dimensionality of the vectors in $Z^A$ and the batch size, respectively.
    The main contribution of VICReg is the variance preservation term, which explicitly prevents a collapse due to a shrinkage of the embedding vectors toward zero.
    The variance regularization term $v$ in VICReg is defined as a hinge loss function applied to the standard deviation of the embeddings along the batch dimension:
    \begin{eqnarray}\label{equ:7_1}
      v\left(Z^A\right) = \frac{1}{d}\sum_{j=1}^{d}{\max(0,{\gamma}-S\left(z_j^A,{\varepsilon}\right))}.
      \end{eqnarray}
      Here, $z_{j}^{A}$ represents the vector composed of each value at dimension $j$ in \(Z^A\) and $S$ represents the regularized standard deviation, defined as
    \begin{eqnarray}\label{equ:7_2}
      S(y,{\varepsilon}) = \sqrt{\text{Var}(y)+{\varepsilon}}.
      \end{eqnarray}
      The constant $\gamma$ determines the standard deviation and is set to 1 in the experiments, while $\varepsilon$ is a small scalar used to prevent numerical instabilities. This criterion encourages the variance within the current batch to be equal to or greater than $\gamma$ for every dimension, thereby preventing collapse scenarios where all data are mapped to the same vector.

      The invariance criterion $s$ in VICReg, which captures the similarity between $Z^A$ and $Z^B$, is defined as the mean-squared Euclidean distance between each pair of data without any normalization:
\begin{eqnarray}\label{equ:7_3}
  s\left( {{Z^A},{Z^B}} \right) = \frac{1}{n}\sum\limits_{b = 1}^n {\left\| {z_b^A - z_b^B} \right\|_2^2} .
\end{eqnarray}
In addition, the covariance criterion $c(Z)$ in VICReg is defined as
\begin{eqnarray}\label{equ:7_4}
  {c}\left({Z}\right)=\frac{1}{d}\sum_{i \neq j}\left[C(Z)\right]_{i,j}^2,
\end{eqnarray}
where $C(Z)$ represents the covariance matrix of $Z$. The overall loss of VICReg is a weighted sum of the variance, invariance, and covariance:
  \begin{eqnarray}\label{equ:7_5}
    \begin{array}{l}
      \mathcal{L} = s\left( {{Z^A},{Z^B}} \right) + \alpha \left( {v\left( {{Z^A}} \right) + v\left( {{Z^B}} \right)} \right) \\ 
      \quad  + \beta \left( {C\left( {{Z^A}} \right) + C\left( {{Z^B}} \right)} \right), \\ 
      \end{array} 
    \end{eqnarray} 
where $\alpha$ and $\beta$ are two hyper-parameters. Note that both regularization terms — the variance regularization term and the covariance regularization term — are applied independently to each branch of the architecture. This differs from the Barlow Twins, which uses a cross-correlation matrix between the two branches of the Siamese architecture.

\paragraph{Analysis of CL}

Despite the impressive results achieved by contrastive SSL, the underlying mechanisms remain obscure and not fully understood. Several studies have delved into this area \cite{tschannen2019mutual,saunshi2019theoretical,yang2020rethinking,tsai2020demystifying,wang2020understanding,chuang2020debiased,lee2020predicting,chen2021large,haochen2021provable,tosh2021contrastive,wei2020theoretical,tian2022deep}. Theoretical investigations by \cite{saunshi2019theoretical,lee2020predicting,tosh2021contrastive} have provided support for the value of feature representations generated through CL. In the Appendix, we also provide explanations of the connections between contrastive learning and other concepts, such as Principal Component Analysis, Spectral Clustering, and Supervised Learning.


\paragraph{Others}

Besides the aforementioned works, several other approaches have employed CL. Among them, \cite{chen2021empirical,caron2021emerging} investigated the utilization of vision transformers (ViTs) as the backbone for contrastive SSL, employing multi-crop and cross-entropy loss \cite{caron2021emerging}. Notably, \cite{caron2021emerging} discovered that the resultant features exhibited exceptional performance as $K$-nearest neighbors ($K$-NN) classifiers and effectively encoded explicit information regarding the semantic segmentation of images. These desirable properties have also motivated specific downstream tasks \cite{wang2022self}.

In a different study, \cite{hoffer2016deep} adopted patches extracted from the same image as a positive pair, while patches from different images served as negative pairs. A mixing operation is further explored in RegionCL~\cite{xu2022regioncl} to diversify the contrastive pairs. Yang et al. \cite{yang2022reading} integrated CL and MIM in the context of text recognition, utilizing a weighted objective function.

Numerous CL-based methods are available in the literature \cite{zhu2021improving,yang2021partially,islam2021broad,li2021learning,jing2021understanding,zhang2021video,hu2020adco,kalantidis2020hard,purushwalkam2020demystifying}. It should be noted that CL is not restricted solely to SSL, as it can also be used in supervised learning \cite{khosla2020supervised}.

\subsubsection{Generative algorithms}

\begin{figure}
  \begin{center}
  \scalebox{0.7}{\includegraphics{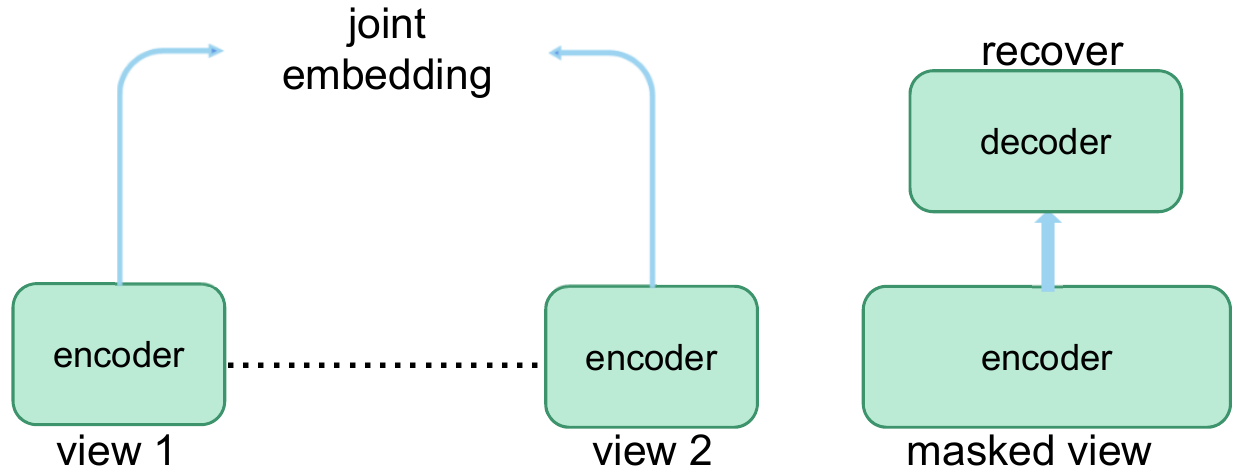}}
  \end{center}
     \caption{The broad differences between CL and MIM. Note that the actual differences between their pipelines are not limited to what is shown.}
  \label{fig:4_1}
  \end{figure}

For the category of generative algorithms, this study primarily focuses on MIM methods. MIM methods \cite{zhou2021ibot} (Fig. \ref{fig:4_1})—namely, bidirectional encoder representation from image transformers (BEiT) \cite{bao2021beit}, masked AE (MAE) \cite{he2021masked}, context AE (CAE) \cite{chen2022context}, and a simple framework for MIM (SimMIM) \cite{xie2021simmim}—have gained significant popularity and pose a considerable challenge to the prevailing dominance of CL. MIM leverages co-occurrence relationships among image patches as supervision signals.

MIM is a variant of the denoising AE (DAE) \cite{vincent2008extracting}. Notably, the Bidirectional Encoder Representations from Transformers (BERT) \cite{devlin2018bert} and Generative Pre-trained Transformer (GPT) \cite{brown2020language} have emerged as a renowned variant of the DAE and achieved remarkable success in NLP. Researchers aspire to extend this success to CV by employing BERT-like pre-training strategies. However, it is crucial to acknowledge that BERT's success in NLP can be attributed not only to its large-scale self-supervised pre-training but also to its scalable network architecture. A notable distinction between the NLP and CV communities is their use of different primary models, with transformers being prevalent in NLP and CNNs being widely adopted in CV.

The landscape changed significantly with the introduction of the original ViT \cite{dosovitskiy2020image}, which marked a pivotal moment. Alexey Dosovitskiy et al. conducted pioneering research on applying MIM to CV, drawing inspiration from BERT's masked image prediction paradigm. Their smaller ViT-B/16 model achieved 79.9\% accuracy on ImageNet~\cite{deng2009imagenet} through self-supervised pre-training, an impressive 2\% improvement over training from scratch. However, it still fell short of the accuracy attained by supervised pre-training. iGPT \cite{chen2020generative} further employs the GPT-style next token prediction, but it received limited attention due to its subpar accuracy and computational efficiency. Beyond ViTs, a separate early investigation adopted context encoders \cite{pathak2016context}, employing a concept akin to MAE, \textit{i.e.}, image inpainting.

However, the differences between natural language and visual signals limit the effectiveness of naive paradigm transfer. BEiT introduces a tailored MIM task for visual pre-training, \textit{i.e.}, an extra tokenization procedure which breaks down the input image into visual tokens, and then predicts randomly masked subset of the image tokens. To address the challenge of tokenization, the authors leveraged a discrete variational autoencoder (dVAE) \cite{ramesh2021zero} to create a predefined visual vocabulary. In contrast to BEiT, MAE does not utilize image tokens; instead, it approaches the problem from the perspective of image signal sparsity. MAE identifies a significant amount of redundancy in image signals, necessitating a higher masking rate, such as 75\%.

Here, we define
\begin{equation}
    \operatorname{MIM} :=\mathcal{L}\left(\mathcal{D}\left(\mathcal{E}\left(\mathcal{T}_{1}\left(I\right)\right)\right), \mathcal{T}_{2}\left(I\right)\right),
\end{equation} where $\mathcal{E}$ denotes the encoder, $\mathcal{D}$ denotes the decoder, $\mathcal{T}_{1}$ represents the transformation applied to the input before it is fed into the network, and $\mathcal{T}_{2}$ represents the transformation used to derive the target label. It is noteworthy that this representation is provided for the sake of clarity and ease of understanding rather than serving as a strict definition.

The primary distinction between BEiT and MAE lies in their choice of $\mathcal{T}$. While BEiT employs the token output from the pre-trained tokenizer as its target, MAE directly uses the original pixels as its target. BEiT adopts a two-stage approach, initially training a tokenizer to convert images into visual tokens, followed by BERT-style training. On the other hand, MAE is a one-stage end-to-end approach, incorporating a decoder to decode the encoder-derived representation into the original pixels. The two representative MIM approaches BEiT and MAE, showcase different architectural designs, with subsequent MIM methods often following one of these techniques. A central challenge in MIM lies in the selection of the target representation $\mathcal{T}_{2}$, which leads to the categorization of MIM methods, as presented in Table \ref{MIM-classification}.

\begin{table*}[htbp]
  \centering
  \caption{Categorization of MIM methods based on the reconstruction target. The second and third rows denote MIM methods and reconstructing targets, respectively.}
  \resizebox{\linewidth}{!}{%
  \begin{tabular}{c|cccc|ccc|cc|cc}
  \hline
   & \multicolumn{4}{c|}{Low-Level Targets} & \multicolumn{3}{c|}{High-Level Targets} & \multicolumn{2}{c|}{Self-Distillation} & \multicolumn{2}{c}{Contrastive / Multi-modal Teacher} \\
 \hline
  Algorithm & ViT \cite{dosovitskiy2020image} & MAE \cite{he2021masked} & SimMIM \cite{xie2021simmim} & Maskfeat \cite{wei2022masked} & BEiT \cite{bao2021beit} & CAE \cite{chen2022context} & PeCo \cite{dong2021peco} & data2vec \cite{baevski2022data2vec} & SdAE \cite{chen2022sdae} & MimCo \cite{zhou2022mimco} & BEiT v2 \cite{peng2022beitv2}\\
  \hline
  Target & \multicolumn{3}{c|}{Raw Pixel} & HOG & \multicolumn{2}{c|}{VQ-VAE} & VQ-GAN & \multicolumn{2}{c|}{self} & \multicolumn{1}{c|}{MoCo v3} & CLIP \\
\hline
\end{tabular}}
  \label{MIM-classification}  
\end{table*}

\begin{figure*}
  \begin{center}
  \includegraphics[width=\linewidth]{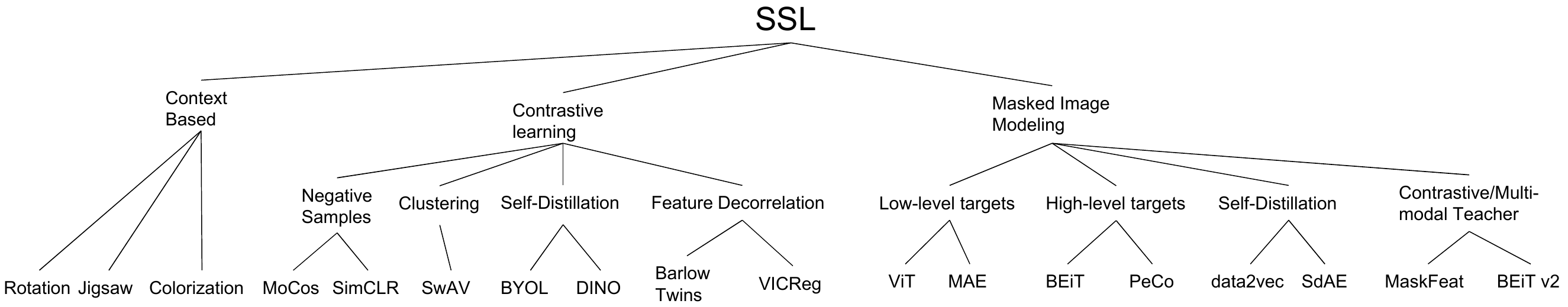}
  \end{center}
     \caption{Several representative pretext tasks of SSL.}
  \label{fig:5_1}
  \end{figure*}

Following the introduction of BEiT and MAE, several variants have been proposed. iBOT \cite{zhou2021ibot} is an ``online tokenizer'' adaptation of BEiT, aiming to address the limitation of dVAE in capturing only low-level semantics within local details. The CAE introduces an alignment constraint to encourage masked patch representations (predicted by a ``latent contextual regressor'') to lie in the encoded representation space. This decoupling of the representation learning task and pretext task enhances the model's capacity for representation learning. Furthermore, MAE has been extended to other modalities beyond images \cite{feichtenhofer2022masked,liang2022meshmae,pang2022masked}.

Generative pre-training has also evolved in the video domain. BEVT \cite{wang2022bevt} decouples video representation learning into spatial representation learning and temporal dynamics learning. It first undertakes masked image modeling on image data, followed by a joint approach of masked image modeling and masked video modeling on video data. This accelerates training and achieves results comparable to those of strongly-supervised baselines. Similarly, VideoMAE \cite{tong2022videomae} extends the MAE to videos and discovers that an extremely high proportion of masking ratio (90\% to 95\%) is permissible in video mask modeling. Moreover, it remains effective even on very small datasets, consisting of only 3,000 to 4,000 videos. OmniMAE \cite{Girdhar2023OmniMAE} demonstrates that a unified model can be concurrently trained across multiple visual modalities, breaking the paradigm of previously studying different modes in isolation. This significantly streamlines the training process, enabling more efficient development of large-scale model architectures. SiamMAE \cite{gupta2023Siamese} indicates that, contrary to images that are (approximately) isotropic, the temporal dimension is unique, necessitating an asymmetric approach to processing temporal and spatial information, as not all spatiotemporal orientations are equally probable.

MIM has demonstrated significant potential in pre-training vision transformers~\cite{liu2022swin,li2022exploring,xu2022vitpose}. However, in prior works, the random masking of image patches led to an underutilization of valuable semantic information essential for effective visual representation learning. Liu et al. \cite{Gui2023Good} introduced an attention-driven masking strategy to explore improvements over random masking for insufficient semantic utilization.

\subsubsection{Contrastive Generative Methods}
As stated in \cite{qi2023contrast}, contrastive models tend to be data-hungry and vulnerable to overfitting issues, whereas generative models encounter data-filling challenges and exhibit inferior data scaling capabilities when compared to contrastive models. While contrastive models often focus on global views \cite{caron2021emerging}, overlooking internal structures within images, MIM primarily models local relationships. The divergent characteristics and challenges encountered in contrastive self-supervised learning and generative self-supervised learning have motivated researchers to explore the combination of these two kinds of approaches.

To elaborate further, let us compare the challenges faced by contrastive self-supervised methods and generative self-supervised methods. Generative self-supervised methods are characterized as data-filling approaches \cite{xie2023data}. For a model of a certain size, when the dataset reaches a certain magnitude, further scaling of the data does not lead to significant performance gains in generative self-supervised methods. In contrast, recent studies have revealed the potential of data scaling to enhance the performance of CL \cite{oquab2023dinov2}. As data increases, CL shows substantial performance improvements, demonstrating remarkable generalization without additional fine-tuning on downstream tasks. However, the scenario differs in low-data regimes. Contrastive models may find shortcuts with trivial representations that overfit the limited data \cite{he2020momentum}, thus leading to inconsistent improvements in generalization performance for downstream tasks using pre-trained models with contrastive self-supervised methods \cite{qi2023contrast}. On the other hand, generative methods are more adept at handling low-data scenarios and can even achieve notable performance improvements when data is extremely scarce, such as with only 10 images \cite{kong2023understanding}.

Several endeavors have sought to integrate both types of algorithms \cite{chen2021joint,qi2023contrast}. In \cite{chen2021joint}, GANs are employed for online data augmentation in CL. The study devises a contrastive module that learns view-invariant features for generation and introduces a view-invariant loss function to facilitate learning between original and generated views. On the other hand, \cite{zhou2021ibot} draws inspiration from both BEiT and DINO \cite{caron2021emerging}. It modifies the tokenizer of BEiT to an online distilled teacher while integrating cross-view distillation from the DINO framework. As a result, iBOT \cite{zhou2021ibot} significantly enhances linear probing accuracy compared to the MIM method. RePre \cite{wang2022RePre} integrates local feature learning into self-supervised vision transformers through reconstructive pre-training, an approach that enhances contrastive frameworks. This is achieved by incorporating an additional branch dedicated to reconstructing raw image pixels, which operates concurrently with the established contrastive objective. CMAE \cite{huang2023Contrastive} concurrently performs CL and MIM tasks. To align CL with MIM effectively, CMAE introduces two novel components: pixel shifting for generating plausible positive views, and a feature decoder for enhancing the features of contrastive pairs. This approach significantly improves the quality of representation and transfer performance compared to its MIM-only counterparts. 
SiameseIM \cite{tao2023Siamese} does not simply merge the objectives of CL and MIM, but rather utilizes the views generated by CL as the target for MIM reconstruction in the latent space.

Despite attempts to combine both types of approaches, naive combinations may not always yield performance gains and can even perform worse than the generative model baseline, thereby exacerbating the issue of representation over-fitting \cite{qi2023contrast}. The performance degradation could be attributed to the disparate properties of CL and generative methods. For instance, CL methods typically exhibit longer attention distances, whereas generative methods tend to favor local attention \cite{xie2023revealing}. In light of this challenge, RECON \cite{qi2023contrast} emerges as a solution by training generative modeling to guide CL, thereby leveraging the benefits of both paradigms.

\subsubsection{Summary}

As described above, numerous pretext tasks for SSL have been devised, with several significant milestone variants depicted in Fig. \ref{fig:5_1}. 

Several other pretext tasks are available \cite{dosovitskiy2014discriminative,dosovitskiy2015discriminative}, encompassing diverse approaches such as relative patch location \cite{doersch2015unsupervised}, noise prediction \cite{bojanowski2017unsupervised}, feature clustering \cite{xie2016unsupervised,yang2016joint,caron2018deep}, cross-channel prediction \cite{zhang2017split}, and combining different cues \cite{wang2017transitive}. 
Kolesnikov et al. \cite{kolesnikov2019revisiting} conducted a comprehensive investigation of previously proposed SSL pretext tasks, yielding significant insights. Besides, Kr{\"a}henb{\"u}hl et al. \cite{krahenbuhl2018free} proposed an alternative approach to pretext tasks and demonstrated the ease of obtaining data from video games.

It has been observed that context-based approaches exhibit limited applicability due to their inferior performance. In the realm of visual SSL, two dominant types of algorithms are CL and MIM. While visual CL may encounter overfitting issues, CL algorithms that incorporate multi-modality, exemplified by CLIP \cite{radford2021learning}, have gained popularity.

\subsection{Combinations with other learning paradigms}\label{Connections}

It is essential to acknowledge that the advancements in SSL did not occur in isolation; instead, they have been the result of continuous development over time. In this section, we provide a comprehensive list of relevant learning paradigms that, when combined with SSL, contribute to a clearer understanding of their collective impact.

\subsubsection{GANs}
GANs represent classical unsupervised learning methods and were among the most successful approaches in this domain before the surge of SSL techniques. The integration of GANs with SSL offers various avenues, with self-supervised GANs (SS-GAN) serving as one such example. The GANs' objective function \cite{gui2020review,goodfellow2014generative} is given as
\begin{eqnarray}\label{equ:8}
\begin{array}{l}
 \mathop {\min }\limits_G {\rm{ }}\mathop {\max }\limits_D \;V\left( {G,D} \right) = {E_{x \sim {p_{data}}\left( x \right)}}\left[ {\log D\left( x \right)} \right] \\ 
 \quad  + {E_{z \sim {p_z}\left( z \right)}}\left[ {\log \left( {1 - D\left( {G\left( z \right)} \right)} \right)} \right]. \\ 
 \end{array}
\end{eqnarray}

The SS-GAN \cite{chen2019self} is defined by combining the objective functions of GANs with the concept of rotation \cite{gidaris2018unsupervised}:
\begin{align}\label{equ:9}
    L_G(G,D) &=  - V(G,D) \notag \\
             &\quad - \alpha \mathbb{E}_{x \sim p_G} \mathbb{E}_{r \sim R}[\log Q_D\left( R = r|x^r \right)],
\end{align}
\begin{align}\label{equ:9_1}
    L_D(G,D) &= V(G,D) \notag \\
             &\quad - \beta \mathbb{E}_{x \sim p_{data}} \mathbb{E}_{r \sim R}[\log Q_D\left( R = r|x^r \right)],
\end{align}
where $V(G,D)$ represents the objective function of GANs as given in Eq.~\eqref{equ:8}, and $r \sim R$ refers to a rotation selected from a set of possible rotations, similar to the concept presented in \cite{gidaris2018unsupervised}. Here, $x^r$ denotes an image $x$ rotated by $r$ degrees, and $Q\left( {R|{x^r}} \right)$ corresponds to the discriminator's predictive distribution over the angles of rotation for a given example $x$. Notably, rotation \cite{gidaris2018unsupervised} serves as a classical SSL method. The SS-GAN incorporates rotation invariance into the GANs' generation process by integrating the rotation prediction task during training.

\subsubsection{Semi-supervised learning}
SSL and semi-supervised learning are contrasting paradigms that can be effectively combined. One notable example of this combination is self-supervised semi-supervised learning (S$^4$L) \cite{zhai2019s4l}. In S$^4$L, the objective function is given by
  \begin{eqnarray}\label{equ:10}
    \mathcal{L} = \mathop {\min }\limits_\theta  \;{\mathcal{L}_l}\left( {{D_l},\theta } \right) + w{\mathcal{L}_u}\left( {{D_u},\theta } \right).
    \end{eqnarray}
This means optimizing the corresponding loss objectives on a labeled dataset $D_l$ and an unlabeled dataset $D_u$. $\mathcal{L}_l$ is the categorization loss (e.g., cross-entropy) and $\mathcal{L}_u$ stands for the self-supervised loss (\textit{e.g.}, rotation task in Eq.~\eqref{equ:3}). $\theta$ is the learnable parameters.

Incorporating SSL as an auxiliary task is a well-established approach in semi-supervised learning. Another classical method to leverage SSL within this context involves implementing SSL on unlabeled data, followed by fine-tuning the resultant model on labeled data, as demonstrated in the SimCLR.

To demonstrate the robustness of self-supervision against adversarial perturbations, Hendrycks et al. \cite{hendrycks2019using} proposed an overall loss function as a linear combination of supervised and self-supervised losses:
  \begin{eqnarray}\label{equ:10_1}
    \begin{array}{l}
     \mathcal{L}(x,y,\theta ) = {\mathcal{L}_{CE}}\left( {y,p\left( {y|PGD(x)} \right),\theta } \right) \\ 
     \quad  + \lambda {\mathcal{L}_{SS}}\left( {PGD(x),\theta } \right), \\ 
     \end{array}
     \end{eqnarray}
where $x$ is the example, $y$ is the one-hot vector of ground-truth and $\theta$ denotes the model parameters. The adversarial example is generated from $x$ by 
projected gradient descent (PGD) and adversarial training is implemented by cross-entropy loss $\mathcal{L}_{CE}$. $\mathcal{L}_{SS}$ is the self-supervised loss.

\subsubsection{Multi-instance learning (MIL)}
Miech et al. \cite{miech2020end} introduced an extension of the InfoNCE loss (\ref{equ:4}) for MIL and termed it MIL-NCE:
\begin{eqnarray}\label{equ:10_2}
\mathop {\max }\limits_{f,g} \,\sum\limits_{i = 1}^n {\log \left( {\frac{{\sum\limits_{(x,y) \in {P_i}} {{e^{f{{\left( x \right)}^T}g\left( y \right)}}} }}{{\sum\limits_{(x,y) \in {P_i}} {{e^{f{{\left( x \right)}^T}g\left( y \right)}}}  + \sum\limits_{(x',y') \in {N_i}} {{e^{f{{\left( {x'} \right)}^T}g\left( {y'} \right)}}} }}} \right)} ,
 \end{eqnarray}
where $x$ and $y$ represent a video clip and a narration, respectively. The functions $f$ and $g$ generate embeddings of $x$ and $y$, respectively. For a specific example indexed by $i$, $P_i$ denotes the set of positive video/narration pairs, while $N_i$ corresponds to the set of negative video/narration pairs.

\subsubsection{Multi-view/multi-modal(ality) learning}

Observation plays a vital role in infants' acquisition of knowledge about the world. Notably, they can grasp the concept of apples through observational and comparative processes, which distinguishes their learning approach from traditional supervised algorithms that rely on extensive labeled apple data. This phenomenon was demonstrated by Orhan et al. \cite{orhan2020self}, who gathered perceptual data from infants and employed an SSL algorithm to model how infants learn the concept of ``apple''. Moreover, infants' learning about the world extends to multi-view and multi-modal(ality) learning \cite{radford2021learning}, encompassing various sensory inputs such as video and audio. Hence, SSL and multi-view/multi-modal(ality) learning converge naturally in infants' learning mechanisms as they explore and comprehend the workings of the world.

\paragraph{Multiview CL} The objective function in standard multiview CL, as proposed by Tian et al. \cite{tian2020makes}, is given by
  \begin{eqnarray}\label{equ:11}
    {\mathcal{L}_{NCE}} = E\left[ {{L_q}} \right],
    \end{eqnarray}
    where $L_q$ corresponds to Eq. (\ref{equ:4}). Multiview CL treats different views of the same sample as positive examples for contrastive learning. Tian et al. \cite{tian2020makes} introduced both unsupervised and semi-supervised multiview learning based on adversarial learning.  Let $\hat X$ denote $g(X)$, \textit{i.e.}, $\hat X = g(X)$. Two encoders, $f_1$ and $f_2$, were trained to maximize $I_{NCE}(\hat X_1, \hat X_{2:3})$ as stated in Eq.~\eqref{equ:11}. A flow-based model $g$ was trained to minimize $I_{NCE}(\hat X_1, \hat X_{2:3})$ and $\left\{ {{X_1},{X_{2:3}}} \right\}$ is obtained from image splitting over its channels. Formally, the objective function for unsupervised view learning can be expressed as
\begin{eqnarray}\label{equ:12}
\mathop {\min }\limits_g \mathop {\max }\limits_{{f_1},{f_2}} I_{_{NCE}}^{{f_1},{f_2}}(g{\left( X \right)_1},g{\left( X \right)_{2:3}}).
\end{eqnarray}

In the context of semi-supervised view learning, when several labeled examples are available, the objective function is formulated as
\begin{eqnarray}\label{equ:13}
\begin{array}{l}
 \mathop {\min }\limits_{g,{c_1},{c_2}} \mathop {\max }\limits_{{f_1},{f_2}} I_{_{NCE}}^{{f_1},{f_2}}(g{\left( X \right)_1},g{\left( X \right)_{2:3}}) \\ 
  + {L_{ce}}\left( {{c_1}\left( {g{{\left( X \right)}_1}} \right),y} \right) + {L_{ce}}\left( {{c_2}\left( {g{{\left( X \right)}_{2:3}}} \right),y} \right), \\ 
 \end{array}
 \end{eqnarray}
where $y$ represents the labels, $c_1$ and $c_2$ are classifiers, and $L_{ce}$ denotes the cross-entropy. Further relevant works can be found in \cite{tian2020makes, tian2019contrastive, hassani2020contrastive}. Table \ref{tab:2} summarizes different SSL losses.

\begin{table*}
\caption{Different losses of SSL.}
  \begin{center}
  \resizebox{\linewidth}{!}{%
  \begin{tabular}{c c c c c}\hline
  \multicolumn{2}{c}{Category} & Method & Loss & Equation \\\cline{1-5}
   \multirow{6}{*}{\begin{tabular}[c]{@{}c@{}}\\ Pretext\end{tabular}} & Context-Based & Rotation \cite{gidaris2018unsupervised} & Rotation Prediction & (\ref{equ:3}) \\\cline{2-5}
    & \multirow{5}{*}{\begin{tabular}[c]{@{}c@{}}\\ CL\end{tabular}} & MoCo v1 \cite{he2020momentum} & InfoNCE & (\ref{equ:4})
   \\
   & &  SimCLR v1 \cite{chen2020simple} & InfoNCE & (\ref{equ:5}) \\
   &  & SimSiam \cite{chen2020exploring} & Cosine Similarity &(\ref{equ:5_4})\\
   & &  Barlow Twins \cite{zbontar2021barlow} & Invariance, and Covariance &(\ref{equ:6})\\
   & &  VICReg \cite{bardes2021vicreg} & Variance, Invariance, and Covariance & (\ref{equ:7_5})
  \\\cline{1-5}
  \multirow{4}{*}{\begin{tabular}[c]{@{}c@{}} Combinations\\ with Other\\ Learning Paradigms\end{tabular}} &  &  SS-GAN \cite{chen2019self} &
  GAN loss + Rotation Prediction & (\ref{equ:9} \& \ref{equ:9_1})\\
   &  &   S$^4$L \cite{zhai2019s4l} & Supervised and Unsupervised Loss & (\ref{equ:10})\\
   & & SSL improving robustness \cite{hendrycks2019using} & Supervised and Self-supervised Adversarial Training Loss & (\ref{equ:10_1}) \\
  & &unsupervised multi-view learning \cite{tian2020makes} & Self-supervised Loss on Multiple Views & (\ref{equ:12}) \\
  \cline{1-5}
  \end{tabular}}
  \end{center}
  \label{tab:2}
  \end{table*}

\paragraph{Images and text}

In the study conducted by Gomez et al. \cite{gomez2017self}, the authors employed a topic modeling framework to project the text of an article into the topic probability space. This semantic-level representation was then utilized as the self-supervised signal for training CNN models on images. On a similar note, CLIP \cite{radford2021learning} leverages a CL-style pre-training task to predict the correspondence between captions and images. Benefiting from the CL paradigm, CLIP is capable of training models from scratch on an extensive dataset comprising 400 million image-text pairs collected from the internet. Consequently, CLIP's advancements have significantly propelled multi-modal learning to the forefront of research attention.

\paragraph{Point clouds and other modalities}
Several SSL methods have been proposed for joint learning of 3D point cloud features and 2D image features by leveraging cross-modality and cross-view correspondences through triplet and cross-entropy losses \cite{jing2020self}. Additionally, there are efforts to jointly learn view-invariant and mode-invariant characteristics from diverse modalities, such as images, point clouds, and meshes, using heterogeneous networks for 3D data \cite{jing2020selfM}. SSL has also been employed for point cloud datasets, with approaches including CL and clustering based on graph CNNs \cite{zhang2019unsupervised}. Furthermore, AEs have been used for point clouds in works like \cite{yang2018foldingnet,gadelha2018multiresolution,liang2022meshmae,pang2022masked}, while capsule networks have been applied to point cloud data in \cite{zhao20193d}.

\subsubsection{Test time training}
Sun et al. \cite{sun2020test} introduced ``test time training (TTT) with self-supervision'' to enhance the performance of predictive models when the training and test data come from distinct distributions. TTT converts an individual unlabeled test example into an SSL problem, enabling model parameter updates before making predictions. Recently, Gandelsman et al. \cite{gandelsman2022test} combined TTT with MAE for improved performance. They argued that by treating TTT as a one-sample learning problem, optimizing a model for each test input could be addressed using the MAE as
\begin{equation}\label{equ:13_1}
h_0=\arg \min _h \frac{1}{n} \sum_{i=1}^n \mathcal{L}_m\left(h \circ f_0\left(x_i\right), y_i\right),
\end{equation}
\begin{equation}\label{equ:13_2}
f_x, g_x=\arg \min _{f, g} \mathcal{L}_s(g \circ f(\operatorname{mask}(x)), x).
\end{equation}
Here, $f$ and $g$ refer to the encoder and decoder of MAE, and $h$ denotes the main task head, respectively. 

TTT achieves an improved bias-variance tradeoff under distribution shifts. A static model heavily depends on training data that may not accurately represent the new test distribution, leading to bias. On the other hand, training a new model from scratch for each test input, ignoring all training data, is undesirable. This approach results in an unbiased representation for each test input but exhibits high variance due to its singularity.

\subsubsection{Summary}

The evolution of SSL is characterized by its dynamic and interconnected nature. Analyzing the amalgamation of various methods allows for a clearer grasp of SSL's developmental trajectory. An exemplar of this success is evident in CLIP, which effectively combines CL with multi-modal learning, leading to remarkable achievements. SSL has been extensively integrated with various machine learning tasks, showcasing its versatility and potential. It has been combined with clustering \cite{caron2020unsupervised}, semi-supervised learning \cite{zhai2019s4l}, multi-task learning \cite{sun2021task, ren2018cross}, transfer learning \cite{saito2020universal, sun2019unsupervised, noroozi2018boosting}, graph NNs \cite{hu2020gpt,rong2020self, hassani2020contrastive}, reinforcement learning \cite{buchler2018improving, guo2020bootstrap, hansen2020self}, few-shot learning \cite{gidaris2019boosting, su2019boosting}, neural architecture search \cite{li2021bossnas}, robust learning \cite{fan2021does, kim2020adversarial, chen2020adversarial, hendrycks2019using}, and meta-learning \cite{lin2021self, anconditional}. This diverse integration underscores the widespread applicability and impact of SSL in the machine learning domain.

\section{Applications}
SSL initially emerged in the context of vowel class recognition \cite{pal1978computer}, and subsequently, it was extended to encompass object extraction tasks \cite{ghosh1993self}. SSL has found widespread applications in diverse domains, including CV, NLP, medical image analysis, and remote sensing (RS).

\subsection{CV}
Sharma et al. \cite{sharma2016vconv} introduced a fully convolutional volumetric AE for unsupervised deep embeddings learning of object shapes. In addition, SSL has been extensively applied to various aspects of image processing and CV: image inpainting \cite{pathak2016context}, human parsing \cite{gong2017look,liang2018look}, scene deocclusion \cite{zhan2020self}, semantic image segmentation \cite{pathak2017learning,wang2020self}, monocular vision \cite{chen2021aggnet}, person reidentification (re-ID) \cite{chen2021ice,isobe2021towards}, visual odometry \cite{li2020self}, scene flow estimation \cite{wupointpwc}, knowledge distillation \cite{xu2020knowledge}, optical flow prediction \cite{walker2015dense}, vision-language navigation \cite{zhu2020vision}, physiological signal estimation \cite{niu2020rhythmnet,niu2020video}, image denoising \cite{xie2020noise2same,huang2021neighbor2neighbor}, object detection \cite{yang2021instance,croitoru2017unsupervised,xie2021detco}, super-resolution \cite{wu2021practical,menon2020pulse}, voxel prediction from 2D images \cite{girdhar2016learning}, and ego-motion \cite{jayaraman2015learning,yin2018geonet}. These applications highlight the broad impact and relevance of SSL in the realm of image processing and CV.

\subsubsection{SSL models for videos }
SSL has garnered widespread usage across various applications, including video representation learning \cite{huang2021self, hu2021contrast, tschannen2020self} and video retrieval \cite{He2022Learn}.

\paragraph{Temporal information in videos}
Various forms of temporal information in videos can be employed, encompassing frame order, video playback direction, video playback speed, and future prediction information \cite{han2019video,han2020memory}. 
1) The order of the frames. Several studies have explored the significance of frame order in videos. Misra et al. \cite{misra2016shuffle} introduced a method for learning visual representations from raw spatiotemporal signals and determining the correct temporal sequence of frames extracted from videos. Fernando et al. \cite{fernando2017self} proposed a novel self-supervised CNN pre-training approach called ``odd-one-out learning,'' where the objective is to identify the unrelated or odd element within a set of related elements. This odd element corresponds to a video subsequence with an incorrect temporal frame order, while the related elements maintain the correct temporal order. Lee et al. \cite{lee2017unsupervised} employed temporally shuffled frames, presented in a non-chronological order, as inputs to train a CNN for predicting the correct order of the shuffled sequences, effectively using temporal coherence as a self-supervised signal. Building upon this work, Xu et al. \cite{xu2019self} utilized temporally shuffled clips as inputs instead of individual frames, training 3D CNNs to sort these shuffled clips. 
2) Video playback direction. Temporal direction analysis in videos, as studied by Wei et al. \cite{wei2018learning}, involves discerning the arrow of time to determine if a video sequence progresses in the forward or backward direction. 
3) Video playback speed. Video playback speed has been a subject of investigation in several studies. Benaim et al. \cite{benaim2020speednet} focused on predicting the speeds of moving objects in videos, determining whether they moved faster or slower than the normal speed. Yao et al. \cite{yao2020video} leveraged playback rates and their corresponding video content as self-supervision signals for video representation learning. Additionally, Wang et al. \cite{wang2020selfV} addressed the challenge of self-supervised video representation learning through the lens of video pace prediction.

\paragraph{Motions of objects in videos}
Diba et al. \cite{diba2019dynamonet} focused on SSL of motions in videos by employing dynamic motion filters to enhance motion representations, particularly for improving human action recognition. The concept of SSL with videos (CoCLR) \cite{han2020Self} bears similarities to SimCLR \cite{chen2020simple}. 

\paragraph{Multi-modal(ality) data in videos}
The auditory and visual components in a video are intrinsically interconnected. Leveraging this correlation, Korbar et al. \cite{korbar2018cooperative} employed a self-supervised temporal synchronization approach to learn comprehensive and effective models for both video and audio analysis. Similarly, other methodologies \cite{arandjelovic2017look,arandjelovic2018objects} are also founded on joint video and audio modalities while certain studies \cite{sun2019videobert,nagrani2020speech2action,stroud2020learning} incorporated both video and text modalities. Moreover, Alayrac et al. \cite{alayrac2020self} explored a tri-modal approach involving vision, audio, and language in videos. On a different note, Sermanet et al. \cite{sermanet2018time} proposed a self-supervised technique for learning representations and robotic behaviors from unlabeled videos captured from various viewpoints. 

\paragraph{Spatial-temporal coherence of objects in videos}
Wang et al. \cite{wang2019learning} introduced a self-supervised algorithm for learning visual correspondence in unlabeled videos by utilizing cycle consistency in time as a self-supervised signal. Extensions of this work have been explored by Li et al. \cite{li2019joint} and Jabri et al. \cite{jabri2020space}. Lai et al. \cite{lai2020mast} presented a memory-augmented self-supervised method that enables generalizable and accurate pixel-level tracking. Zhang et al. \cite{zhang2020online} employed spatial-temporal consistency of depth maps to mitigate forgetting during the learning process. Zhao et al. \cite{luo2020video} proposed a novel self-supervised algorithm named the ``video cloze procedure (VCP),'' which facilitates learning rich spatial-temporal representations for videos. Feichtenhofer et al. \cite{feichtenhofer2022masked} extended the MAE to video representation learning and demonstrated that leveraging the naive ViT along with the spatiotemporal co-occurrence of videos can outperform the vanilla supervised training. Gupta et al. \cite{gupta2023Siamese} demonstrated the importance of asymmetrically modeling the spatiotemporal information of videos.

\subsubsection{Universal sequential SSL models for image processing and CV}
Contrastive predictive coding (CPC) \cite{oord2018representation} operates on the fundamental concept of acquiring informative representations through latent space predictions of future data using robust autoregressive models. While initially applied to sequential data like speech and text, CPC has also found applicability to images \cite{henaff2019data}.

Drawing inspiration from the accomplishments of GPT \cite{radford2018improving,brown2020language} in NLP, iGPT \cite{chen2020generative} investigates whether similar models can effectively learn representations for images. iGPT explores two training objectives, namely autoregressive prediction and a denoising objective, thereby sharing similarities with BERT \cite{devlin2018bert}. In high-resolution scenarios, this approach \cite{chen2020generative} competes favorably with other self-supervised methods on ImageNet~\cite{deng2009imagenet}. Similar to iGPT, ViT \cite{dosovitskiy2020image} also adopts a transformer architecture for vision tasks. By applying a pure transformer to sequences of image patches, ViT has demonstrated outstanding performance in image recognition tasks. The transformer architecture has been further extended to various vision-related applications, as evidenced by \cite{li2021efficient,he2021masked,bao2021beit,caron2021emerging,chen2021empirical}.

\subsection{NLP}
In the realm of NLP, pioneering works for performing SSL on word embeddings include the continuous bag-of-words model and the continuous skip-gram model \cite{mikolov2013distributed}. They can be considered as belonging to generative self-supervised learning algorithms, which have long dominated the field of NLP. Despite their diverse forms, these algorithms are fundamentally based on language models that employ maximum likelihood estimation. Discriminative algorithms (e.g., contrastive learning) were initially deemed ineffective due to the distinct semantics inherent in language. 
Some discriminative algorithms aim to challenge the conventions, among which ELECTRA \cite{clark2020electra} stands out as a pioneer. ELECTRA employs the Replaced Token Detection (RTD) task and draws upon the structure and ideas of GANs (notably, without adopting GAN's training paradigm) to pre-train a language model. \cite{pappas2019gile} demonstrated that supervised contrastive pretraining enables zero-shot prediction of unseen text classes and enhances few-shot performance. A series of subsequent works have demonstrated that task-agnostic self-supervised contrastive pre-training \rev{has} been shown to improve language modeling \cite{clark2020pre,wu2020clear,giorgi2020declutr}.
However, the automatic creation of textual input augmentation remains a significant challenge, as a single token can reverse the meaning of a sentence. Generative SSL algorithms continue to dominate NLP, from early works such as BERT and GPT to recent trillion-scale large language models.

\subsection{Other fields}
Within the medical field \cite{zhou2021preservational}, the availability of labeled data is typically limited, while a vast amount of unlabeled data exists. This natural scenario makes SSL a compelling approach, which has been effectively employed for various tasks like medical image segmentation \cite{chaitanya2020contrastive} and 3D medical image analysis \cite{zhu2020rubik}. Recently, SSL has also found applications in the remote sensing domain, benefiting from the abundance of large-scale unlabeled data that remains largely unexplored. For example, SeCo~\cite{manas2021seasonal} leverages seasonal changes in RS images to construct positive pairs and perform CL. On the other hand, RVSA~\cite{wang2022advancing} introduces a novel rotated varied-size window attention mechanism that advances the plain vision transformer to serve as a fundamental model for various remote sensing tasks. Notably, it is pre-trained using the generative SSL method MAE~\cite{he2021masked} on the large-scale MillionAID dataset.

\begin{table*}[htbp]
\caption{Experimental results of the tested algorithms for linear classification and transfer learning tasks. DB denotes the default batch size. The symbol ``-'' indicates the absence or unavailability of the data point in the respective paper. The subscripts A, R, and V represent AlexNet, ResNet-50, and ViT-B, respectively. The superscript ``e'' indicates the utilization of extra data, specifically VOC2012.}
\resizebox{\linewidth}{!}{%
  \begin{tabular}{ccccccccc}
      \hline
      Methods                                 & Linear Probe                                                          & Fine-Tuning                             & VOC\_det                                   & VOC\_seg                                 & COCO\_det                           & COCO\_seg                          & ADE20K\_seg                         & DB   \\
      \hline
      Random:                                 & $17.1_{A}$\cite{noroozi2016unsupervised}                              & -                                        & $60.2_{R}^{e}$\cite{chen2020exploring} & $19.8_{A}$\cite{noroozi2016unsupervised} & $36.7_{R}$\cite{he2020momentum}     & $33.7_{R}$\cite{he2020momentum}    & -                                   & -    \\
      R50 Sup                                 & 76.5\cite{caron2020unsupervised}                                      & 76.5\cite{caron2020unsupervised}         & $81.3^{e}$\cite{chen2020exploring}     & 74.4\cite{grill2020bootstrap}            & 40.6\cite{he2020momentum}           & 36.8\cite{he2020momentum}          & -                                   & -    \\
      ViT-B Sup                               & 82.3\cite{he2021masked}                                                                  & 82.3\cite{he2021masked}                  & -                                          & -                                        & 47.9\cite{he2021masked}             & 42.9\cite{he2021masked}            & 47.4\cite{he2021masked}             & -    \\
      \hline
      \multicolumn{9}{l}{\textbf{Context-Based:}} \\
      \hline
      Jigsaw\cite{noroozi2016unsupervised}    & $45.7_{R}$\cite{caron2020unsupervised}                                & $54.7$                               & $61.4_{R}$\cite{goyal2019scaling}          & $37.6$                               & -                                   & -                                  & -                                   & 256  \\
      Colorization\cite{zhang2016colorful}    & $39.6_{R}$\cite{caron2020unsupervised}                                & $40.7$\cite{gidaris2018unsupervised} & $46.9$                                 & $35.6$                               & -                                   & -                                  & -                                   & -   \\
      Rotation\cite{gidaris2018unsupervised}  & $38.7$                                                            & $50.0$                               & $54.4$                                 & $39.1$                               & -                                   & -                                  & -                                   & 128  \\
      \hline
      \multicolumn{9}{l}{\textbf{CL Based on Negative Examples:}}                                                                                                                                                                                                                                                                                               \\
      \hline
      Examplar\cite{dosovitskiy2014discriminative}        & $31.5$\cite{wu2018unsupervised}                                   & -                                        & -                                          & -                                        & -                                   & -                                  & -                                   & -    \\
      Instdisc\cite{wu2018unsupervised}       & $54.0$                                                            & -                                        & $65.4$                                 & -                                        & -                                   & -                                  & -                                   & 256  \\
      MoCo v1\cite{he2020momentum}            & $60.6$                                                            & -                                        & $74.9$                                 & -                                        & $40.8$                          & $36.9$                         & -                                   & 256  \\
      SimCLR\cite{chen2020simple}             & $73.9_{V}$\cite{chen2021empirical} & -                                        & $81.8^{e}$\cite{chen2020exploring} & -                                        & $37.9$\cite{chen2020exploring}  & $33.3$\cite{chen2020exploring} & -                                   & 4096 \\
      MoCo v2\cite{chen2020improved}          & $72.2$\cite{chen2020exploring}                                    & -                                        & $82.5^{e}$                         & -                                        & $39.8$\cite{bardes2021vicreg}   & $36.1$\cite{bardes2021vicreg}  & -                                   & 256  \\
      MoCo v3\cite{chen2021empirical}         & 76.7                                                & 83.2                               & -                                          & -                                        & 47.9\cite{he2021masked}       & 42.7\cite{he2021masked}      & 47.3\cite{he2021masked}       & 4096 \\
      \hline
      \multicolumn{9}{l}{\textbf{CL Based on Clustering:}}                                                                                                                                                                                                                                                                                                     \\
      \hline
      SwAV\cite{caron2020unsupervised}        & $75.3$                         & -                                        & $82.6^{e}$\cite{bardes2021vicreg}  & -                                        & $41.6$                          & $37.8$\cite{bardes2021vicreg}  & -                                   & 4096 \\
      \hline
      \multicolumn{9}{l}{\textbf{CL Based on Self-distillation:}}                                                                                                                                                                                                                                                                                              \\
      \hline
      BYOL\cite{grill2020bootstrap}           & $74.3$                                                            & -                                        & $81.4^{e}$\cite{chen2020exploring} & $76.3$                               & $40.4$\cite{bardes2021vicreg}   & $37.0$\cite{bardes2021vicreg}  & -                                   & 4096 \\
      SimSiam\cite{chen2020exploring}         & $71.3$                                                            & -                                        & $82.4^{e}$\cite{chen2020exploring} & -                                        & $39.2$                          & $34.4$                         & -                                   & 512  \\
      DINO\cite{caron2021emerging}            & 78.2                                              & 83.6\cite{zhou2021ibot}            & -                                          & -                                        & 46.8\cite{chen2022context}        & 41.5\cite{chen2022context}       & 44.1\cite{bao2021beit}        & 1024 \\
      \hline
      \multicolumn{9}{l}{\textbf{CL Based on Feature Decorrelation:}}                                                                                                                                                                                                                                                                                          \\
      \hline
      Barlow Twins\cite{zbontar2021barlow}    & $73.2$                                                            & -                                        & $82.6^{e}$\cite{bardes2021vicreg}  & -                                        & $39.2$                          & $34.3$                         & -                                   & 2048 \\
      VICReg\cite{bardes2021vicreg}           & $73.2$                                                            & -                                        & $82.4^{e}$                        & -                                        & $39.4$                          & $36.4$                         & -                                   & 2048 \\
      \hline
      \multicolumn{9}{l}{\textbf{Masked Image Modeling (ViT-B by default):}}                                                                                                                                                                                                                                                                                                     \\
      \hline
      Context Encoder\cite{pathak2016context} & $21.0_{A}$\cite{gidaris2018unsupervised}                              & -                                        & $44.5_{A}$\cite{gidaris2018unsupervised}   & $30.0_{A}$                               & -                                   & -                                  & -                                   & -    \\
      BEiT v1\cite{bao2021beit}               & 56.7\cite{peng2022beitv2}                                               & 83.4\cite{zhou2021ibot}                  & -                                          & -                                        & 49.8\cite{he2021masked}             & 44.4\cite{he2021masked}            & 47.1\cite{he2021masked}             & 2000 \\
      MAE\cite{he2021masked}                  & 67.8                                                                  & 83.6                                     & -                                          & -                                        & 50.3                                & 44.9                               & 48.1                                & 4096 \\
      SimMIM\cite{xie2021simmim}              & 56.7                                                                  & 83.8                                     & -                                          & -                                        & $52.3_{Swin-B}$\cite{Liu2022MixMIM} & -                                  & $52.8_{Swin-B}$\cite{Liu2022MixMIM} & 2048 \\
      PeCo\cite{dong2021peco} & - & 84.5 & - & - & 43.9 & 39.8 & 46.7 & 2048 \\
      iBOT\cite{zhou2021ibot}                 & 79.5                                                                  & 84.0                                     & -                                          & -                                        & 51.2                                & 44.2                               & 50.0                                & 1024 \\
      MimCo\cite{zhou2022mimco} & - & 83.9 & - & - & 44.9 & 40.7 & 48.91 & 2048 \\
      CAE\cite{chen2022context}                   & 70.4                                                                  & 83.9                                     & -                                          & -                                        & 50                                  & 44                                 & 50.2                                & 2048 \\
      data2vec\cite{baevski2022data2vec} & - & 84.2 & - & - & - & - & - & 2048 \\
      SdAE\cite{chen2022sdae} & 64.9 & 84.1 & - & - & 48.9 & 43.0 & 48.6 & 768 \\
      BEiT v2\cite{peng2022beitv2}              & 80.1                                                                  & 85.5                                     & -                                          & -                                        & -                                   & -                                  & 53.1                                & 2048 \\
      \hline
      \end{tabular}}
  \label{all}
  \end{table*}
  
\section{Performance comparison}
Once a pre-trained model is obtained through SSL, the assessment of its performance becomes necessary. The conventional approach involves gauging the achieved performance on downstream tasks to ascertain the quality of the extracted features. However, this evaluation metric does not provide insights into what the network has specifically learned during self-supervised pre-training. To delve into the interpretability of self-supervised features, alternative evaluation metrics, such as network dissection \cite{bau2017network} and other unsupervised methods \cite{garrido2023RankMe}, can be employed.
In this section, we aim to present a clear demonstration of the performance comparison. We summarize the pre-trained dataset performance and transfer learning efficacy of typical SSL methods on well-established datasets. Note that SSL can technically be applied to diverse modalities. However, for the sake of simplicity, we narrow our focus to SSL in the vision domain.

\subsection{Comprehensive comparison}
We present the results in Table~\ref{all} and \ref{video-performance}. In cases where a method reproduced from another subsequent work achieves superior accuracy compared to the original paper, we report the results with the higher one. Please note that although we have endeavored to align the experimental settings, minor variations in hyper-parameters can still affect the performance. Refer to the original paper if necessary. The experimental results are obtained according to the default backbone specified in the original papers, such as ResNet-50 or ViT-B/16. Additionally, results from alternative backbones were provided in instances where data using the default backbone was not available, and marked accordingly.

\textbf{Setup}. 
Upon the 2D image, the model was pre-trained on ImageNet-1k~\cite{deng2009imagenet} and evaluated on semantic segmentation tasks on PASCAL VOC \cite{everingham2010pascal}, COCO \cite{lin2015microsoft}, and ADE20k \cite{zhou2017scene,zhou2019semantic}, as well as on object detection tasks on VOC and COCO, and on classification tasks on ImageNet-1k.
Upon the video, the model is pre-trained on the Kinetics \cite{kay2017kinetics} or Something Something-v2 (SSv2) \cite{goyal2017something} datasets, and its performance is evaluated 
on action detection tasks on the Kinetics, SSv2, and AVA \cite{gu2018ava} datasets.

The evaluation of object detection on the PASCAL VOC dataset employs mean average precision (mAP), specifically $AP_{50}$. By default, the object detection task on PASCAL VOC employs VOC2007 for training. However, certain methods employ the combined 07+12 dataset and are annotated with a superscript ``e''. As for the object detection and instance segmentation tasks on COCO, we adopt the bounding-box AP ($AP_{bb}$) and mask AP ($AP_{mk}$) metrics, in accordance with \cite{he2020momentum}. The results on video understanding are evaluated using fine-tuned Top-1 accuracy as the metric.

\begin{table*}[ht]
\caption{Performance comparison of SSL methods for video.}
\centering
\label{video-performance}
\begin{tabular}{@{}llllllll@{}}
\toprule
\multicolumn{8}{c}{Contrastive Methods} \\ \midrule
 &  &  &  & \multicolumn{4}{c}{Downstream Dataset} \\
Method & \begin{tabular}[c]{@{}l@{}}Pre-training\\ Dataset\end{tabular} & Backbone & \begin{tabular}[c]{@{}l@{}}Linear\\ Probe\end{tabular}  & \multicolumn{2}{c}{UCF101 \cite{soomro2012ucf101}} & \multicolumn{2}{c}{HMDB51 \cite{kuehne2011hmdb}} \\
       &         &          &              & Linear & Fine-tune          & Linear & Fine-tune \\ \midrule
DSM \cite{wang2021enhancing} &  K400  &  R3D34   &     -    &    -    &      78.2         &    -    &  52.8     \\
TCE \cite{knights2021temporally} &  K400  &  R50   &     -    &    -    &      71.2         &    -    &  36.6     \\
CoCRL \cite{han2020Self} &  K400  &  S3D-G   &    -   &   74.5 \cite{recasens2021broaden}    &      87.9         &  46.1 \cite{recasens2021broaden}      &  54.6     \\
CoCRL  &  K400  &  2×S3D-G   &     -    &    -    &      90.6         &    -    &  62.9     \\
VTHCL \cite{yang2020video} & K400    & R3D50    & 37.8 \cite{feichtenhofer2021large}       &   -  & 82.1               & -  & 49.2      \\
CVRL \cite{qian2021spatiotemporal}  & K400    & R3D50    & 66.1      &   89.2     & 92.2   &   57.3     & 66.7      \\
CVRL  & K600    & R3D50    & 70.4         &   90.6     & 93.4        &   59.7     & 68.0      \\
$\rho$BYOL \cite{feichtenhofer2021large}  & K400    & R3D50    & 71.5         &     -   & 95.5     &   -     & 73.6      \\
$\rho$BYOL & K400    & S3D-G    & -    &   -     & 96.3    &    -    & 75.0      \\ 
BraVe \cite{recasens2021broaden} & K400    & R3D50   & -  & 90.6   & 93.7   &  65.1   &   72.0        \\ 
BraVe & K600    & R3D50    & 69.1   &   91.9  &  94.4   &   67.6    &   73.9    \\ \midrule
\multicolumn{8}{c}{Masked Image Modeling Methods} \\ \midrule
 & & & \multicolumn{5}{c}{Downstream Dataset} \\
Method & \begin{tabular}[c]{@{}l@{}}Pre-training\\ Dataset\end{tabular} & Backbone & \multicolumn{2}{c}{K400 \cite{kay2017kinetics}} & \multicolumn{2}{c}{SSv2 \cite{goyal2017something}} & \multicolumn{1}{c}{AVA \cite{gu2018ava}} \\ \midrule
MaskFeat \cite{wei2022masked} & K400 & MViTv2-L/312 & \multicolumn{2}{c}{86.4} & \multicolumn{2}{c}{74.4} & \multicolumn{1}{c}{37.5} \\
BEVT \cite{wang2022bevt} & K400 & Swin-B & \multicolumn{2}{c}{76.2} & \multicolumn{2}{c}{67.1} & \multicolumn{1}{c}{-} \\
BEVT & IN1K + K400 & Swin-B & \multicolumn{2}{c}{80.6} & \multicolumn{2}{c}{70.6} & \multicolumn{1}{c}{-} \\
VidelMAE \cite{tong2022videomae} & K400 & ViT-B & \multicolumn{2}{c}{80.0} & \multicolumn{2}{c}{68.5} & \multicolumn{1}{c}{26.7} \\
VidelMAE & SSv2 & ViT-B & \multicolumn{2}{c}{69.6} & \multicolumn{2}{c}{79.6} & \multicolumn{1}{c}{-} \\
VidelMAE & SSv2 & ViT-L & \multicolumn{2}{c}{-} & \multicolumn{2}{c}{75.4} & \multicolumn{1}{c}{34.3} \\
MAE-ST \cite{feichtenhofer2022masked} & K400 & ViT-L & \multicolumn{2}{c}{84.8} & \multicolumn{2}{c}{72.1} & \multicolumn{1}{c}{32.3} \\
OmniMAE \cite{Girdhar2023OmniMAE} & IN1K + K400 & ViT-B & \multicolumn{2}{c}{80.8} & \multicolumn{2}{c}{69.0} & \multicolumn{1}{c}{-} \\
OmniMAE & IN1K + SSv2 & ViT-B & \multicolumn{2}{c}{80.6} & \multicolumn{2}{c}{69.5} & \multicolumn{1}{c}{-} \\
OmniMAE & IN1K + SSv2 & ViT-L & \multicolumn{2}{c}{84.0} & \multicolumn{2}{c}{74.2} & \multicolumn{1}{c}{-} \\
\bottomrule
\end{tabular}
\end{table*}

\subsection{Summary}
First, the linear probe performance of contrastive learning models typically surpasses that of other algorithms, and contrastive learning approaches tend to regard the linear probe as a significant performance metric. This superiority is attributed to contrastive learning generating well-structured latent spaces, wherein distinct categories are effectively separated, and similar categories are appropriately clustered.

Secondly, it is observed that pre-trained models using MIM can be fine-tuned to achieve superior performance in most cases. Conversely, pre-trained models based on CL lack this property. One primary reason for this discrepancy lies in the increased susceptibility of CL-based models to overfitting \cite{robinson2021can, Wang2022Contrastive, wei2022contrastive}. This observation also extends to the fine-tuning of pre-trained models for downstream tasks. MIM-based approaches consistently exhibit substantial performance enhancements in downstream tasks, while CL-based methods offer comparatively limited assistance.

Thirdly, CL-based methods tend to employ resource-intensive techniques like momentum encoders, memory queues, and multi-crop, significantly increasing the demands on computing, storage, and communication resources. In contrast, MIM-based methods have a more efficient resource utilization, possibly attributed to the absence of example interactions. This advantageous property allows MIM-based algorithms to easily scale up models and data, efficiently leveraging modern GPUs for high parallel computing. We compared the computational complexity of different SSL methods in Table 1 of the Appendix. Note that the primary sources of time complexity and memory consumption are the neural network other than SSL components, e.g., the calculation of the cross-correlation matrix in Barlow Twins.

\section{Conclusions, Future Trends, and Open Questions}

In summary, this comprehensive review offers essential insights into contemporary SSL research, providing newcomers with an overall picture of the field. The paper presents a thorough survey of SSL from three main perspectives: algorithms, applications, and future trends. We focus on mainstream visual SSL algorithms, classifying them into four major types: context-based methods, generative methods, contrastive methods, and contrastive generative methods. Furthermore, we investigate the correlation between SSL and other learning paradigms. Lastly, we will delve into future trends and open problems as outlined below.

\textbf{Main trends}: 
Firstly, the theoretical cloud still looms over SSL. How can we understand different SSL algorithms and unify them in the same way physics seeks to unify the four fundamental forces? \cite{wang2020understanding} analyzed the key properties of contrastive learning based on negative samples, enhancing the understanding of representation distributions. \cite{haochen2021provable} rethought contrastive learning from the perspective of spectral decomposition, providing a high-level understanding of why contrastive learning is effective. \cite{chen2021Intriguing} showed practical properties, with InfoMin \cite{tian2020makes} indicating that the design of views should consider downstream tasks.
\cite{tian2021understanding} investigated why distillation-based methods do not collapse. \cite{garrido2023Duality} demonstrated the duality between negative example-based contrastive learning and covariance regularization-based methods such as Barlow Twins, indicating the latter can be seen as contrastive between the dimensions of the embeddings instead of between the samples. \cite{lavoie2023simplicial} demonstrated that introducing discrete sparse overcomplete representations for SSL can improve generalization. \cite{tao2022Exploring} presented the connections and distinctions among various SSL methods from the perspective of gradients. We anticipate that new theoretical studies will aid in comprehending and unifying various SSL approaches, particularly in harmonizing CL-based methods with MIM-based methods.

Secondly, a crucial question arises concerning the automatic design of an optimal pretext task to enhance the performance of a fixed downstream task. Various methods have been proposed to address this challenge, including the pixel-to-propagation consistency method \cite{xie2020propagate} and dense contrastive learning \cite{wang2021dense}. However, this problem remains insufficiently resolved, and further theoretical investigations are warranted in this direction.

Thirdly, there is a pressing need for a unified SSL paradigm that encompasses multiple modalities. MIM has demonstrated remarkable progress in vision tasks, akin to the success of masked language model in NLP, suggesting the possibility of unifying learning paradigms. Additionally, the ViT architecture bridges the gap between visual and verbal modalities, enabling the construction of a unified transformer model for both CV and NLP tasks.
Recent endeavors \cite{wang2022image, baevski2022data2vec} have sought to unify SSL models, yielding impressive results in downstream tasks and showing broad applicability. Nevertheless, NLP has advanced further in leveraging SSL models, prompting the CV community to draw inspiration from NLP approaches to effectively harness the potential of pre-trained models. 
 
\textbf{Open problems}: 
Can SSL effectively leverage vast amounts of unlabeled data? How does it consistently benefit from additional unlabeled data, and how can we determine the theoretical inflection point?

Secondly, it is pertinent to explore the interconnection between SSL and multi-modality learning, as both methodologies share resemblances with the cognitive processes observed in infants. Consequently, a critical inquiry arises: how can these two approaches be synergistically integrated to forge a robust and comprehensive learning model?

Thirdly, determining the most optimal or recommended SSL algorithm poses a challenge as there is no universally applicable solution. 
The ideal selection of an algorithm should align with the specific problem structure, yet practical situations often complicate this process. Consequently, the development of a checklist to aid users in identifying the most suitable method under particular circumstances warrants investigation and should be pursued as a promising avenue for future research.

Fourthly, the assumption that unlabeled data invariably leads to improved outcomes warrants scrutiny. Our hypothesis challenges this notion, especially concerning semi-supervised learning methods, as the \textit{no free lunch} theorem comes into play. Performance degradation can arise when model assumptions fail to align effectively with the underlying problem structure. For instance, if a model assumes a substantial separation between decision boundaries and regions of high data density, it may perform poorly when faced with data originating from heavily overlapping Cauchy distributions, as the decision boundary would traverse through dense areas. However, preemptively identifying such mismatches remains intricate and an unresolved matter. Consequently, this topic merits further research to shed light on the matter.


%



\ifCLASSOPTIONcompsoc
  \section*{Acknowledgments}
\else
  \section*{Acknowledgment}
\fi

This work was supported in part by the grant of the National Science Foundation of China under Grant 62172090, U23B2054, 62276263; Start-up Research Fund of Southeast University under Grant RF1028623097. Dr Tao’s research is partially supported by NTU RSR and Start Up Grants. We thank the Big Data Computing Center of Southeast University for providing the facility support on the numerical calculations.

\ifCLASSOPTIONcaptionsoff
  \newpage
\fi



%


\bibliographystyle{ieeetr}
\normalem
\bibliography{ssl_survey}

\begin{thebibliography}{100}

\bibitem{deng2009imagenet}
J.~Deng, W.~Dong, R.~Socher, L.-J. Li, K.~Li, and L.~Fei-Fei, ``Imagenet: A
  large-scale hierarchical image database,'' in {\em IEEE Conf. Comput. Vis.
  Pattern Recognit.}, pp.~248--255, 2009.

\bibitem{radford2021learning}
A.~Radford, J.~W. Kim, C.~Hallacy, A.~Ramesh, G.~Goh, S.~Agarwal, G.~Sastry,
  A.~Askell, P.~Mishkin, J.~Clark, {\em et~al.}, ``Learning transferable visual
  models from natural language supervision,'' in {\em Int. Conf. Mach. Learn.},
  pp.~8748--8763, 2021.

\bibitem{ericsson2021well}
L.~Ericsson, H.~Gouk, and T.~M. Hospedales, ``How well do self-supervised
  models transfer?,'' in {\em IEEE Conf. Comput. Vis. Pattern Recognit.},
  pp.~5414--5423, 2021.

\bibitem{liu2021self}
X.~Liu, F.~Zhang, Z.~Hou, L.~Mian, Z.~Wang, J.~Zhang, and J.~Tang,
  ``Self-supervised learning: Generative or contrastive,'' {\em IEEE T. Knowl.
  Data Eng.}, 2022.

\bibitem{dosovitskiy2020image}
A.~Dosovitskiy, L.~Beyer, A.~Kolesnikov, D.~Weissenborn, X.~Zhai,
  T.~Unterthiner, M.~Dehghani, M.~Minderer, G.~Heigold, S.~Gelly, {\em et~al.},
  ``An image is worth 16x16 words: Transformers for image recognition at
  scale,'' in {\em Int. Conf. Learn. Represent.}, 2021.

\bibitem{tran2015learning}
D.~Tran, L.~Bourdev, R.~Fergus, L.~Torresani, and M.~Paluri, ``Learning
  spatiotemporal features with 3d convolutional networks,'' in {\em IEEE Int.
  Conf. Comput. Vis.}, pp.~4489--4497, 2015.

\bibitem{gidaris2018unsupervised}
S.~Gidaris, P.~Singh, and N.~Komodakis, ``Unsupervised representation learning
  by predicting image rotations,'' in {\em Int. Conf. Learn. Represent.},
  pp.~1--14, 2018.

\bibitem{noroozi2016unsupervised}
M.~Noroozi and P.~Favaro, ``Unsupervised learning of visual representations by
  solving jigsaw puzzles,'' in {\em Eur. Conf. Comput. Vis.}, pp.~69--84, 2016.

\bibitem{misra2016shuffle}
I.~Misra, C.~L. Zitnick, and M.~Hebert, ``Shuffle and learn: unsupervised
  learning using temporal order verification,'' in {\em Eur. Conf. Comput.
  Vis.}, pp.~527--544, 2016.

\bibitem{wei2018learning}
D.~Wei, J.~J. Lim, A.~Zisserman, and W.~T. Freeman, ``Learning and using the
  arrow of time,'' in {\em IEEE Conf. Comput. Vis. Pattern Recognit.},
  pp.~8052--8060, 2018.

\bibitem{devlin2018bert}
J.~Devlin, M.-W. Chang, K.~Lee, and K.~Toutanova, ``Bert: Pre-training of deep
  bidirectional transformers for language understanding,'' {\em arXiv preprint
  arXiv:1810.04805}, 2018.

\bibitem{zeng2020realistic}
X.~Zeng, Y.~Pan, M.~Wang, J.~Zhang, and Y.~Liu, ``Realistic face reenactment
  via self-supervised disentangling of identity and pose,'' in {\em AAAI
  Conf.Artif. Intell.}, pp.~12154--12163, 2020.

\bibitem{miech2020end}
A.~Miech, J.-B. Alayrac, L.~Smaira, I.~Laptev, J.~Sivic, and A.~Zisserman,
  ``End-to-end learning of visual representations from uncurated instructional
  videos,'' in {\em IEEE Conf. Comput. Vis. Pattern Recognit.}, pp.~9879--9889,
  2020.

\bibitem{asano2019critical}
Y.~M. Asano, C.~Rupprecht, and A.~Vedaldi, ``A critical analysis of
  self-supervision, or what we can learn from a single image,'' in {\em Int.
  Conf. Learn. Represent.}, 2020.

\bibitem{hinton2006reducing}
G.~E. Hinton and R.~R. Salakhutdinov, ``Reducing the dimensionality of data
  with neural networks,'' {\em Science}, vol.~313, no.~5786, pp.~504--507,
  2006.

\bibitem{vincent2008extracting}
P.~Vincent, H.~Larochelle, Y.~Bengio, and P.-A. Manzagol, ``Extracting and
  composing robust features with denoising autoencoders,'' in {\em Int. Conf.
  Mach. Learn.}, pp.~1096--1103, 2008.

\bibitem{pinto2016supersizing}
L.~Pinto and A.~Gupta, ``Supersizing self-supervision: Learning to grasp from
  50k tries and 700 robot hours,'' in {\em IEEE Int. Conf. Robot. Autom.},
  pp.~3406--3413, 2016.

\bibitem{li2016unsupervised}
Y.~Li, M.~Paluri, J.~M. Rehg, and P.~Doll{\'a}r, ``Unsupervised learning of
  edges,'' in {\em IEEE Conf. Comput. Vis. Pattern Recognit.}, pp.~1619--1627,
  2016.

\bibitem{li2016unsupervisedV}
D.~Li, W.-C. Hung, J.-B. Huang, S.~Wang, N.~Ahuja, and M.-H. Yang,
  ``Unsupervised visual representation learning by graph-based consistent
  constraints,'' in {\em Eur. Conf. Comput. Vis.}, pp.~678--694, 2016.

\bibitem{lee2019rethinking}
H.~Lee, S.~J. Hwang, and J.~Shin, ``Rethinking data augmentation:
  Self-supervision and self-distillation,'' {\em arXiv preprint
  arXiv:1910.05872}, 2019.

\bibitem{zoph2020rethinking}
B.~Zoph, G.~Ghiasi, T.-Y. Lin, Y.~Cui, H.~Liu, E.~D. Cubuk, and Q.~Le,
  ``Rethinking pre-training and self-training,'' in {\em Neural Inf. Process.
  Syst.}, pp.~1--13, 2020.

\bibitem{orhan2020self}
A.~E. Orhan, V.~V. Gupta, and B.~M. Lake, ``Self-supervised learning through
  the eyes of a child,'' in {\em Neural Inf. Process. Syst.}, pp.~9960--9971,
  2020.

\bibitem{mitrovic2020representation}
J.~Mitrovic, B.~McWilliams, J.~Walker, L.~Buesing, and C.~Blundell,
  ``Representation learning via invariant causal mechanisms,'' in {\em Int.
  Conf. Learn. Represent.}, pp.~1--19, 2021.

\bibitem{hua2021feature}
T.~Hua, W.~Wang, Z.~Xue, S.~Ren, Y.~Wang, and H.~Zhao, ``On feature
  decorrelation in self-supervised learning,'' in {\em IEEE Int. Conf. Comput.
  Vis.}, pp.~9598--9608, 2021.

\bibitem{timmurphy.org}
VentureBeat, ``Yann {LeCun}, {Yoshua} {Bengio}: Self-supervised learning is key
  to human-level intelligence.''
\newblock
  https://cacm.acm.org/news/244720-yann-lecun-yoshua-bengio-self-supervised-learning-is-key-to-human-level-intelligence/fulltext.

\bibitem{yu2022self}
J.~Yu, H.~Yin, X.~Xia, T.~Chen, J.~Li, and Z.~Huang, ``Self-supervised learning
  for recommender systems: A survey,'' {\em arXiv preprint arXiv:2203.15876},
  2022.

\bibitem{liu2022graph}
Y.~Liu, M.~Jin, S.~Pan, C.~Zhou, Y.~Zheng, F.~Xia, and P.~Yu, ``Graph
  self-supervised learning: A survey,'' {\em IEEE T. Knowl. Data Eng.}, 2022.

\bibitem{mao2020survey}
H.~H. Mao, ``A survey on self-supervised pre-training for sequential transfer
  learning in neural networks,'' {\em arXiv preprint arXiv:2007.00800}, 2020.

\bibitem{schiappa2022self}
M.~C. Schiappa, Y.~S. Rawat, and M.~Shah, ``Self-supervised learning for
  videos: A survey,'' {\em arXiv preprint arXiv:2207.00419}, 2022.

\bibitem{qi2022adversarial}
G.-J. Qi and M.~Shah, ``Adversarial pretraining of self-supervised deep
  networks: Past, present and future,'' {\em arXiv preprint arXiv:2210.13463},
  2022.

\bibitem{jaiswal2020survey}
A.~Jaiswal, A.~R. Babu, M.~Z. Zadeh, D.~Banerjee, and F.~Makedon, ``A survey on
  contrastive self-supervised learning,'' {\em Technologies}, vol.~9, no.~1,
  pp.~1--22, 2020.

\bibitem{de1994learning}
V.~R. de~Sa, ``Learning classification with unlabeled data,'' in {\em Neural
  Inf. Process. Syst.}, pp.~112--119, 1994.

\bibitem{lecun2020reflections}
Y.~LeCun and Y.~Bengio, ``Reflections from the turing award winners.''
\newblock https://iclr.cc/virtual\_2020/speaker\_7.html.

\bibitem{Jing2021Self-Supervised}
L.~Jing and Y.~Tian, ``Self-supervised visual feature learning with deep neural
  networks: A survey,'' {\em IEEE Trans. Pattern Anal. Mach. Intell.}, vol.~43,
  no.~11, pp.~4037--4058, 2021.

\bibitem{gui2020review}
J.~Gui, Z.~Sun, Y.~Wen, D.~Tao, and J.~Ye, ``A review on generative adversarial
  networks: Algorithms, theory, and applications,'' {\em IEEE T. Knowl. Data
  Eng.}, 2022.

\bibitem{nathan2018improvements}
T.~Nathan~Mundhenk, D.~Ho, and B.~Y. Chen, ``Improvements to context based
  self-supervised learning,'' in {\em IEEE Conf. Comput. Vis. Pattern
  Recognit.}, pp.~9339--9348, 2018.

\bibitem{agrawal2015learning}
P.~Agrawal, J.~Carreira, and J.~Malik, ``Learning to see by moving,'' in {\em
  IEEE Int. Conf. Comput. Vis.}, pp.~37--45, 2015.

\bibitem{zhang2016colorful}
R.~Zhang, P.~Isola, and A.~A. Efros, ``Colorful image colorization,'' in {\em
  Eur. Conf. Comput. Vis.}, pp.~649--666, 2016.

\bibitem{larsson2016learning}
G.~Larsson, M.~Maire, and G.~Shakhnarovich, ``Learning representations for
  automatic colorization,'' in {\em Eur. Conf. Comput. Vis.}, pp.~577--593,
  2016.

\bibitem{zhang2017real}
R.~Zhang, J.-Y. Zhu, P.~Isola, X.~Geng, A.~S. Lin, T.~Yu, and A.~A. Efros,
  ``Real-time user-guided image colorization with learned deep priors,'' {\em
  arXiv preprint arXiv:1705.02999}, 2017.

\bibitem{larsson2017colorization}
G.~Larsson, M.~Maire, and G.~Shakhnarovich, ``Colorization as a proxy task for
  visual understanding,'' in {\em IEEE Conf. Comput. Vis. Pattern Recognit.},
  pp.~6874--6883, 2017.

\bibitem{goyal2019scaling}
P.~Goyal, D.~Mahajan, A.~Gupta, and I.~Misra, ``Scaling and benchmarking
  self-supervised visual representation learning,'' in {\em IEEE Int. Conf.
  Comput. Vis.}, pp.~6391--6400, 2019.

\bibitem{ahsan2019video}
U.~Ahsan, R.~Madhok, and I.~Essa, ``Video jigsaw: Unsupervised learning of
  spatiotemporal context for video action recognition,'' in {\em Proc. Winter
  Conf. Appl. Comput. Vis.}, pp.~179--189, 2019.

\bibitem{zhan2019self}
X.~Zhan, X.~Pan, Z.~Liu, D.~Lin, and C.~C. Loy, ``Self-supervised learning via
  conditional motion propagation,'' in {\em IEEE Conf. Comput. Vis. Pattern
  Recognit.}, pp.~1881--1889, 2019.

\bibitem{wang20193d}
K.~Wang, L.~Lin, C.~Jiang, C.~Qian, and P.~Wei, ``3d human pose machines with
  self-supervised learning,'' {\em IEEE Trans. Pattern Anal. Mach. Intell.},
  vol.~42, no.~5, pp.~1069--1082, 2019.

\bibitem{noroozi2017representation}
M.~Noroozi, H.~Pirsiavash, and P.~Favaro, ``Representation learning by learning
  to count,'' in {\em IEEE Int. Conf. Comput. Vis.}, pp.~5898--5906, 2017.

\bibitem{misra2020self}
I.~Misra and L.~v.~d. Maaten, ``Self-supervised learning of pretext-invariant
  representations,'' in {\em IEEE Conf. Comput. Vis. Pattern Recognit.},
  pp.~6707--6717, 2020.

\bibitem{wu2018unsupervised}
Z.~Wu, Y.~Xiong, S.~X. Yu, and D.~Lin, ``Unsupervised feature learning via
  non-parametric instance discrimination,'' in {\em IEEE Conf. Comput. Vis.
  Pattern Recognit.}, pp.~3733--3742, 2018.

\bibitem{zhao2020makes}
N.~Zhao, Z.~Wu, R.~W. Lau, and S.~Lin, ``What makes instance discrimination
  good for transfer learning?,'' in {\em Int. Conf. Learn. Represent.},
  pp.~1--11, 2021.

\bibitem{he2020momentum}
K.~He, H.~Fan, Y.~Wu, S.~Xie, and R.~Girshick, ``Momentum contrast for
  unsupervised visual representation learning,'' in {\em IEEE Conf. Comput.
  Vis. Pattern Recognit.}, pp.~9729--9738, 2020.

\bibitem{chen2020improved}
X.~Chen, H.~Fan, R.~Girshick, and K.~He, ``Improved baselines with momentum
  contrastive learning,'' {\em arXiv preprint arXiv:2003.04297}, 2020.

\bibitem{chen2020simple}
T.~Chen, S.~Kornblith, M.~Norouzi, and G.~Hinton, ``A simple framework for
  contrastive learning of visual representations,'' in {\em Int. Conf. Mach.
  Learn.}, pp.~1597--1607, 2020.

\bibitem{chen2020big}
T.~Chen, S.~Kornblith, K.~Swersky, M.~Norouzi, and G.~Hinton, ``Big
  self-supervised models are strong semi-supervised learners,'' in {\em Neural
  Inf. Process. Syst.}, pp.~1--13, 2020.

\bibitem{wang2020understanding}
T.~Wang and P.~Isola, ``Understanding contrastive representation learning
  through alignment and uniformity on the hypersphere,'' in {\em Int. Conf.
  Mach. Learn.}, pp.~9929--9939, 2020.

\bibitem{zbontar2021barlow}
J.~Zbontar, L.~Jing, I.~Misra, Y.~LeCun, and S.~Deny, ``Barlow twins:
  Self-supervised learning via redundancy reduction,'' in {\em Int. Conf. Mach.
  Learn.}, 2021.

\bibitem{bardes2021vicreg}
A.~Bardes, J.~Ponce, and Y.~LeCun, ``Vicreg: Variance-invariance-covariance
  regularization for self-supervised learning,'' in {\em Int. Conf. Learn.
  Represent.}, pp.~1--12, 2022.

\bibitem{hadsell2006dimensionality}
R.~Hadsell, S.~Chopra, and Y.~LeCun, ``Dimensionality reduction by learning an
  invariant mapping,'' in {\em IEEE Conf. Comput. Vis. Pattern Recognit.},
  pp.~1735--1742, 2006.

\bibitem{oord2018representation}
A.~v.~d. Oord, Y.~Li, and O.~Vinyals, ``Representation learning with
  contrastive predictive coding,'' {\em arXiv preprint arXiv:1807.03748}, 2019.

\bibitem{gutmann2010noise}
M.~Gutmann and A.~Hyv{\"a}rinen, ``Noise-contrastive estimation: A new
  estimation principle for unnormalized statistical models,'' in {\em Int.
  Conf. Artif. Intell. Statist.}, pp.~297--304, 2010.

\bibitem{zheng2021ressl}
M.~Zheng, S.~You, F.~Wang, C.~Qian, C.~Zhang, X.~Wang, and C.~Xu, ``Ressl:
  Relational self-supervised learning with weak augmentation,'' {\em arXiv
  preprint arXiv:2107.09282}, 2021.

\bibitem{zhao2020distilling}
N.~Zhao, Z.~Wu, R.~W. Lau, and S.~Lin, ``Distilling localization for
  self-supervised representation learning,'' in {\em AAAI Conf.Artif. Intell.},
  pp.~10990--10998, 2021.

\bibitem{arandjelovic2018objects}
R.~Arandjelovic and A.~Zisserman, ``Objects that sound,'' in {\em Eur. Conf.
  Comput. Vis.}, pp.~435--451, 2018.

\bibitem{tian2019contrastive}
Y.~Tian, D.~Krishnan, and P.~Isola, ``Contrastive multiview coding,'' in {\em
  Eur. Conf. Comput. Vis.}, pp.~776--794, 2020.

\bibitem{tian2020makes}
Y.~Tian, C.~Sun, B.~Poole, D.~Krishnan, C.~Schmid, and P.~Isola, ``What makes
  for good views for contrastive learning,'' in {\em Neural Inf. Process.
  Syst.}, pp.~1--13, 2020.

\bibitem{xie2020propagate}
Z.~Xie, Y.~Lin, Z.~Zhang, Y.~Cao, S.~Lin, and H.~Hu, ``Propagate yourself:
  Exploring pixel-level consistency for unsupervised visual representation
  learning,'' in {\em IEEE Conf. Comput. Vis. Pattern Recognit.},
  pp.~16684--16693, 2021.

\bibitem{Wang2022Contrastive}
X.~Wang and G.-J. Qi, ``Contrastive learning with stronger augmentations,''
  {\em IEEE Trans. Pattern Anal. Mach. Intell.}, pp.~1--12, 2022.

\bibitem{grill2020bootstrap}
J.-B. Grill, F.~Strub, F.~Altch{\'e}, C.~Tallec, P.~H. Richemond,
  E.~Buchatskaya, C.~Doersch, B.~A. Pires, Z.~D. Guo, M.~G. Azar, {\em et~al.},
  ``Bootstrap your own latent: A new approach to self-supervised learning,'' in
  {\em Neural Inf. Process. Syst.}, pp.~1--14, 2020.

\bibitem{caron2020unsupervised}
M.~Caron, I.~Misra, J.~Mairal, P.~Goyal, P.~Bojanowski, and A.~Joulin,
  ``Unsupervised learning of visual features by contrasting cluster
  assignments,'' in {\em Neural Inf. Process. Syst.}, 2020.

\bibitem{chen2020exploring}
X.~Chen and K.~He, ``Exploring simple siamese representation learning,'' in
  {\em IEEE Conf. Comput. Vis. Pattern Recognit.}, pp.~15750--15758, 2021.

\bibitem{he2021masked}
K.~He, X.~Chen, S.~Xie, Y.~Li, P.~Doll{\'a}r, and R.~Girshick, ``Masked
  autoencoders are scalable vision learners,'' in {\em IEEE Conf. Comput. Vis.
  Pattern Recognit.}, pp.~16000--16009, 2022.

\bibitem{tschannen2019mutual}
M.~Tschannen, J.~Djolonga, P.~K. Rubenstein, S.~Gelly, and M.~Lucic, ``On
  mutual information maximization for representation learning,'' in {\em Int.
  Conf. Learn. Represent.}, pp.~1--12, 2020.

\bibitem{saunshi2019theoretical}
N.~Saunshi, O.~Plevrakis, S.~Arora, M.~Khodak, and H.~Khandeparkar, ``A
  theoretical analysis of contrastive unsupervised representation learning,''
  in {\em Int. Conf. Mach. Learn.}, pp.~5628--5637, 2019.

\bibitem{yang2020rethinking}
Y.~Yang and Z.~Xu, ``Rethinking the value of labels for improving
  class-imbalanced learning,'' in {\em Neural Inf. Process. Syst.}, 2020.

\bibitem{tsai2020demystifying}
Y.-H.~H. Tsai, Y.~Wu, R.~Salakhutdinov, and L.-P. Morency, ``Self-supervised
  learning from a multi-view perspective,'' {\em arXiv preprint
  arXiv:2006.05576}, 2020.

\bibitem{chuang2020debiased}
C.-Y. Chuang, J.~Robinson, L.~Yen-Chen, A.~Torralba, and S.~Jegelka, ``Debiased
  contrastive learning,'' in {\em Int. Conf. Learn. Represent.}, 2020.

\bibitem{lee2020predicting}
J.~D. Lee, Q.~Lei, N.~Saunshi, and J.~Zhuo, ``Predicting what you already know
  helps: Provable self-supervised learning,'' {\em arXiv preprint
  arXiv:2008.01064}, 2020.

\bibitem{chen2021large}
S.~Chen, G.~Niu, C.~Gong, J.~Li, J.~Yang, and M.~Sugiyama, ``Large-margin
  contrastive learning with distance polarization regularizer,'' in {\em Int.
  Conf. Mach. Learn.}, pp.~1673--1683, 2021.

\bibitem{haochen2021provable}
J.~Z. HaoChen, C.~Wei, A.~Gaidon, and T.~Ma, ``Provable guarantees for
  self-supervised deep learning with spectral contrastive loss,'' in {\em
  Neural Inf. Process. Syst.}, Nov. 2021.

\bibitem{tosh2021contrastive}
C.~Tosh, A.~Krishnamurthy, and D.~Hsu, ``Contrastive learning, multi-view
  redundancy, and linear models,'' in {\em Algorithmic Learning Theory},
  pp.~1179--1206, 2021.

\bibitem{wei2020theoretical}
C.~Wei, K.~Shen, Y.~Chen, and T.~Ma, ``Theoretical analysis of self-training
  with deep networks on unlabeled data,'' in {\em Int. Conf. Learn.
  Represent.}, pp.~1--15, 2021.

\bibitem{tian2022deep}
Y.~Tian, ``Deep contrastive learning is provably (almost) principal component
  analysis,'' {\em arXiv preprint arXiv:2201.12680}, 2022.

\bibitem{chen2021empirical}
X.~Chen, S.~Xie, and K.~He, ``An empirical study of training self-supervised
  visual transformers,'' in {\em IEEE Int. Conf. Comput. Vis.}, pp.~9640--9649,
  2021.

\bibitem{caron2021emerging}
M.~Caron, H.~Touvron, I.~Misra, H.~J{\'e}gou, J.~Mairal, P.~Bojanowski, and
  A.~Joulin, ``Emerging properties in self-supervised vision transformers,'' in
  {\em IEEE Int. Conf. Comput. Vis.}, pp.~9650--9660, 2021.

\bibitem{wang2022self}
Y.~Wang, X.~Shen, S.~X. Hu, Y.~Yuan, J.~L. Crowley, and D.~Vaufreydaz,
  ``Self-supervised transformers for unsupervised object discovery using
  normalized cut,'' in {\em IEEE Conf. Comput. Vis. Pattern Recognit.},
  pp.~14543--14553, 2022.

\bibitem{hoffer2016deep}
E.~Hoffer, I.~Hubara, and N.~Ailon, ``Deep unsupervised learning through
  spatial contrasting,'' {\em arXiv preprint arXiv:1610.00243}, 2016.

\bibitem{xu2022regioncl}
Y.~Xu, Q.~Zhang, J.~Zhang, and D.~Tao, ``Regioncl: exploring contrastive region
  pairs for self-supervised representation learning,'' in {\em Eur. Conf.
  Comput. Vis.}, pp.~477--494, Springer, 2022.

\bibitem{yang2022reading}
M.~Yang, M.~Liao, P.~Lu, J.~Wang, S.~Zhu, H.~Luo, Q.~Tian, and X.~Bai,
  ``Reading and writing: Discriminative and generative modeling for
  self-supervised text recognition,'' {\em arXiv preprint arXiv:2207.00193},
  2022.

\bibitem{zhu2021improving}
R.~Zhu, B.~Zhao, J.~Liu, Z.~Sun, and C.~W. Chen, ``Improving contrastive
  learning by visualizing feature transformation,'' in {\em IEEE Int. Conf.
  Comput. Vis.}, pp.~10306--10315, 2021.

\bibitem{yang2021partially}
M.~Yang, Y.~Li, Z.~Huang, Z.~Liu, P.~Hu, and X.~Peng, ``Partially view-aligned
  representation learning with noise-robust contrastive loss,'' in {\em IEEE
  Conf. Comput. Vis. Pattern Recognit.}, pp.~1134--1143, 2021.

\bibitem{islam2021broad}
A.~Islam, C.-F. Chen, R.~Panda, L.~Karlinsky, R.~Radke, and R.~Feris, ``A broad
  study on the transferability of visual representations with contrastive
  learning,'' in {\em IEEE Int. Conf. Comput. Vis.}, pp.~8845--8855, 2021.

\bibitem{li2021learning}
J.~Li, C.~Xiong, and S.~C. Hoi, ``Learning from noisy data with robust
  representation learning,'' in {\em IEEE Int. Conf. Comput. Vis.},
  pp.~9485--9494, 2021.

\bibitem{jing2021understanding}
L.~Jing, P.~Vincent, Y.~LeCun, and Y.~Tian, ``Understanding dimensional
  collapse in contrastive self-supervised learning,'' in {\em Int. Conf. Learn.
  Represent.}, pp.~1--11, 2022.

\bibitem{zhang2021video}
J.~Zhang, X.~Xu, F.~Shen, Y.~Yao, J.~Shao, and X.~Zhu, ``Video representation
  learning with graph contrastive augmentation,'' in {\em ACM Int. Conf.
  Multimedia}, pp.~3043--3051, 2021.

\bibitem{hu2020adco}
Q.~Hu, X.~Wang, W.~Hu, and G.-J. Qi, ``Adco: Adversarial contrast for efficient
  learning of unsupervised representations from self-trained negative
  adversaries,'' in {\em IEEE Conf. Comput. Vis. Pattern Recognit.}, 2021.

\bibitem{kalantidis2020hard}
Y.~Kalantidis, M.~B. Sariyildiz, N.~Pion, P.~Weinzaepfel, and D.~Larlus, ``Hard
  negative mixing for contrastive learning,'' in {\em Neural Inf. Process.
  Syst.}, pp.~1--12, 2020.

\bibitem{purushwalkam2020demystifying}
S.~Purushwalkam and A.~Gupta, ``Demystifying contrastive self-supervised
  learning: Invariances, augmentations and dataset biases,'' in {\em Neural
  Inf. Process. Syst.}, pp.~1--12, 2020.

\bibitem{khosla2020supervised}
P.~Khosla, P.~Teterwak, C.~Wang, A.~Sarna, Y.~Tian, P.~Isola, A.~Maschinot,
  C.~Liu, and D.~Krishnan, ``Supervised contrastive learning,'' in {\em Neural
  Inf. Process. Syst.}, pp.~18661--18673, 2020.

\bibitem{zhou2021ibot}
J.~Zhou, C.~Wei, H.~Wang, W.~Shen, C.~Xie, A.~Yuille, and T.~Kong, ``ibot:
  Image bert pre-training with online tokenizer,'' in {\em Int. Conf. Learn.
  Represent.}, pp.~1--12, 2022.

\bibitem{bao2021beit}
H.~Bao, L.~Dong, S.~Piao, and F.~Wei, ``Beit: Bert pre-training of image
  transformers,'' in {\em Int. Conf. Learn. Represent.}, pp.~1--13, 2022.

\bibitem{chen2022context}
X.~Chen, M.~Ding, X.~Wang, Y.~Xin, S.~Mo, Y.~Wang, S.~Han, P.~Luo, G.~Zeng, and
  J.~Wang, ``Context autoencoder for self-supervised representation learning,''
  {\em arXiv preprint arXiv:2202.03026}, 2022.

\bibitem{xie2021simmim}
Z.~Xie, Z.~Zhang, Y.~Cao, Y.~Lin, J.~Bao, Z.~Yao, Q.~Dai, and H.~Hu, ``Simmim:
  A simple framework for masked image modeling,'' in {\em IEEE Conf. Comput.
  Vis. Pattern Recognit.}, pp.~9653--9663, 2022.

\bibitem{brown2020language}
T.~B. Brown, B.~Mann, N.~Ryder, M.~Subbiah, J.~Kaplan, P.~Dhariwal,
  A.~Neelakantan, P.~Shyam, G.~Sastry, A.~Askell, {\em et~al.}, ``Language
  models are few-shot learners,'' {\em arXiv preprint arXiv:2005.14165}, 2020.

\bibitem{chen2020generative}
M.~Chen, A.~Radford, R.~Child, J.~Wu, H.~Jun, P.~Dhariwal, D.~Luan, and
  I.~Sutskever, ``Generative pretraining from pixels,'' in {\em Int. Conf.
  Mach. Learn.}, pp.~1691--1703, 2020.

\bibitem{pathak2016context}
D.~Pathak, P.~Krahenbuhl, J.~Donahue, T.~Darrell, and A.~A. Efros, ``Context
  encoders: Feature learning by inpainting,'' in {\em IEEE Conf. Comput. Vis.
  Pattern Recognit.}, pp.~2536--2544, 2016.

\bibitem{ramesh2021zero}
A.~Ramesh, M.~Pavlov, G.~Goh, S.~Gray, C.~Voss, A.~Radford, M.~Chen, and
  I.~Sutskever, ``Zero-shot text-to-image generation,'' in {\em Int. Conf.
  Mach. Learn.}, pp.~8821--8831, 2021.

\bibitem{wei2022masked}
C.~Wei, H.~Fan, S.~Xie, C.-Y. Wu, A.~Yuille, and C.~Feichtenhofer, ``Masked
  feature prediction for self-supervised visual pre-training,'' in {\em IEEE
  Conf. Comput. Vis. Pattern Recognit.}, pp.~14668--14678, 2022.

\bibitem{dong2021peco}
X.~Dong, J.~Bao, T.~Zhang, D.~Chen, W.~Zhang, L.~Yuan, D.~Chen, F.~Wen, and
  N.~Yu, ``Peco: Perceptual codebook for bert pre-training of vision
  transformers,'' {\em arXiv preprint arXiv:2111.12710}, 2021.

\bibitem{baevski2022data2vec}
A.~Baevski, W.-N. Hsu, Q.~Xu, A.~Babu, J.~Gu, and M.~Auli, ``Data2vec: A
  general framework for self-supervised learning in speech, vision and
  language,'' {\em arXiv preprint arXiv:2202.03555}, 2022.

\bibitem{chen2022sdae}
Y.~Chen, Y.~Liu, D.~Jiang, X.~Zhang, W.~Dai, H.~Xiong, and Q.~Tian, ``Sdae:
  Self-distillated masked autoencoder,'' in {\em Eur. Conf. Comput. Vis.},
  pp.~108--124, 2022.

\bibitem{zhou2022mimco}
Q.~Zhou, C.~Yu, H.~Luo, Z.~Wang, and H.~Li, ``Mimco: Masked image modeling
  pre-training with contrastive teacher,'' in {\em ACM Int. Conf. Multimedia},
  pp.~4487--4495, 2022.

\bibitem{peng2022beitv2}
Z.~Peng, L.~Dong, H.~Bao, Q.~Ye, and F.~Wei, ``Beit v2: Masked image modeling
  with vector-quantized visual tokenizers,'' {\em arXiv preprint
  arXiv:2208.06366}, 2022.

\bibitem{feichtenhofer2022masked}
C.~Feichtenhofer, H.~Fan, Y.~Li, and K.~He, ``Masked autoencoders as
  spatiotemporal learners,'' {\em arXiv preprint arXiv:2205.09113}, 2022.

\bibitem{liang2022meshmae}
Y.~Liang, S.~Zhao, B.~Yu, J.~Zhang, and F.~He, ``Meshmae: Masked autoencoders
  for 3d mesh data analysis,'' in {\em Eur. Conf. Comput. Vis.}, pp.~37--54,
  2022.

\bibitem{pang2022masked}
Y.~Pang, W.~Wang, F.~E. Tay, W.~Liu, Y.~Tian, and L.~Yuan, ``Masked
  autoencoders for point cloud self-supervised learning,'' in {\em Eur. Conf.
  Comput. Vis.}, pp.~604--621, 2022.

\bibitem{wang2022bevt}
R.~Wang, D.~Chen, Z.~Wu, Y.~Chen, X.~Dai, M.~Liu, Y.-G. Jiang, L.~Zhou, and
  L.~Yuan, ``Bevt: Bert pretraining of video transformers,'' in {\em
  Proceedings of the IEEE Conf. Comput. Vis. Pattern Recognit.},
  pp.~14733--14743, 2022.

\bibitem{tong2022videomae}
Z.~Tong, Y.~Song, J.~Wang, and L.~Wang, ``Videomae: Masked autoencoders are
  data-efficient learners for self-supervised video pre-training,'' {\em Neural
  Inf. Process. Syst.}, vol.~35, pp.~10078--10093, 2022.

\bibitem{Girdhar2023OmniMAE}
R.~Girdhar, A.~El-Nouby, M.~Singh, K.~V. Alwala, A.~Joulin, and I.~Misra,
  ``Omnimae: Single model masked pretraining on images and videos,'' in {\em
  Proceedings of the IEEE Conf. Comput. Vis. Pattern Recognit.},
  pp.~10406--10417, June 2023.

\bibitem{gupta2023Siamese}
A.~Gupta, J.~Wu, J.~Deng, and L.~{Fei-Fei}, ``Siamese masked autoencoders,'' in
  {\em Neural Inf. Process. Syst.}, Nov. 2023.

\bibitem{liu2022swin}
Z.~Liu, H.~Hu, Y.~Lin, Z.~Yao, Z.~Xie, Y.~Wei, J.~Ning, Y.~Cao, Z.~Zhang,
  L.~Dong, {\em et~al.}, ``Swin transformer v2: Scaling up capacity and
  resolution,'' in {\em IEEE Conf. Comput. Vis. Pattern Recognit.},
  pp.~12009--12019, 2022.

\bibitem{li2022exploring}
Y.~Li, H.~Mao, R.~Girshick, and K.~He, ``Exploring plain vision transformer
  backbones for object detection,'' in {\em Eur. Conf. Comput. Vis.},
  pp.~280--296, 2022.

\bibitem{xu2022vitpose}
Y.~Xu, J.~Zhang, Q.~Zhang, and D.~Tao, ``Vitpose: Simple vision transformer
  baselines for human pose estimation,'' in {\em Neural Inf. Process. Syst.},
  pp.~38571--38584, 2022.

\bibitem{Gui2023Good}
Z.~Liu, J.~Gui, and H.~Luo, ``Good helper is around you: Attention-driven
  masked image modeling,'' in {\em AAAI Conf.Artif. Intell.}, pp.~1799--1807,
  2023.

\bibitem{qi2023contrast}
Z.~Qi, R.~Dong, G.~Fan, Z.~Ge, X.~Zhang, K.~Ma, and L.~Yi, ``Contrast with
  reconstruct: Contrastive 3d representation learning guided by generative
  pretraining,'' {\em arXiv preprint arXiv:2302.02318}, 2023.

\bibitem{xie2023data}
Z.~Xie, Z.~Zhang, Y.~Cao, Y.~Lin, Y.~Wei, Q.~Dai, and H.~Hu, ``On data scaling
  in masked image modeling,'' in {\em IEEE Conf. Comput. Vis. Pattern
  Recognit.}, pp.~10365--10374, 2023.

\bibitem{oquab2023dinov2}
M.~Oquab, T.~Darcet, T.~Moutakanni, H.~Vo, M.~Szafraniec, V.~Khalidov,
  P.~Fernandez, D.~Haziza, F.~Massa, A.~El-Nouby, {\em et~al.}, ``Dinov2:
  Learning robust visual features without supervision,'' {\em arXiv preprint
  arXiv:2304.07193}, 2023.

\bibitem{kong2023understanding}
X.~Kong and X.~Zhang, ``Understanding masked image modeling via learning
  occlusion invariant feature,'' in {\em IEEE Conf. Comput. Vis. Pattern
  Recognit.}, pp.~6241--6251, 2023.

\bibitem{chen2021joint}
H.~Chen, Y.~Wang, B.~Lagadec, A.~Dantcheva, and F.~Bremond, ``Joint generative
  and contrastive learning for unsupervised person re-identification,'' in {\em
  IEEE Conf. Comput. Vis. Pattern Recognit.}, pp.~2004--2013, 2021.

\bibitem{wang2022RePre}
L.~Wang, F.~Liang, Y.~Li, H.~Zhang, W.~Ouyang, and J.~Shao, ``Repre: Improving
  self-supervised vision transformer with reconstructive pre-training,'' Jan.
  2022.

\bibitem{huang2023Contrastive}
Z.~Huang, X.~Jin, C.~Lu, Q.~Hou, M.-M. Cheng, D.~Fu, X.~Shen, and J.~Feng,
  ``Contrastive masked autoencoders are stronger vision learners,'' {\em IEEE
  Transactions on Pattern Analysis and Machine Intelligence}, pp.~1--13, 2023.

\bibitem{tao2023Siamese}
C.~Tao, X.~Zhu, W.~Su, G.~Huang, B.~Li, J.~Zhou, Y.~Qiao, X.~Wang, and J.~Dai,
  ``Siamese image modeling for self-supervised vision representation
  learning,'' in {\em Proceedings of the IEEE Conf. Comput. Vis. Pattern
  Recognit.}, pp.~2132--2141, 2023.

\bibitem{xie2023revealing}
Z.~Xie, Z.~Geng, J.~Hu, Z.~Zhang, H.~Hu, and Y.~Cao, ``Revealing the dark
  secrets of masked image modeling,'' in {\em IEEE Conf. Comput. Vis. Pattern
  Recognit.}, pp.~14475--14485, 2023.

\bibitem{dosovitskiy2014discriminative}
A.~Dosovitskiy, J.~T. Springenberg, M.~Riedmiller, and T.~Brox,
  ``Discriminative unsupervised feature learning with convolutional neural
  networks,'' in {\em Neural Inf. Process. Syst.}, pp.~766--774, 2014.

\bibitem{dosovitskiy2015discriminative}
A.~Dosovitskiy, P.~Fischer, J.~T. Springenberg, M.~Riedmiller, and T.~Brox,
  ``Discriminative unsupervised feature learning with exemplar convolutional
  neural networks,'' {\em IEEE Trans. Pattern Anal. Mach. Intell.}, vol.~38,
  no.~9, pp.~1734--1747, 2015.

\bibitem{doersch2015unsupervised}
C.~Doersch, A.~Gupta, and A.~A. Efros, ``Unsupervised visual representation
  learning by context prediction,'' in {\em IEEE Int. Conf. Comput. Vis.},
  pp.~1422--1430, 2015.

\bibitem{bojanowski2017unsupervised}
P.~Bojanowski and A.~Joulin, ``Unsupervised learning by predicting noise,'' in
  {\em Int. Conf. Mach. Learn.}, 2017.

\bibitem{xie2016unsupervised}
J.~Xie, R.~Girshick, and A.~Farhadi, ``Unsupervised deep embedding for
  clustering analysis,'' in {\em Int. Conf. Mach. Learn.}, pp.~478--487, 2016.

\bibitem{yang2016joint}
J.~Yang, D.~Parikh, and D.~Batra, ``Joint unsupervised learning of deep
  representations and image clusters,'' in {\em IEEE Conf. Comput. Vis. Pattern
  Recognit.}, pp.~5147--5156, 2016.

\bibitem{caron2018deep}
M.~Caron, P.~Bojanowski, A.~Joulin, and M.~Douze, ``Deep clustering for
  unsupervised learning of visual features,'' in {\em Eur. Conf. Comput. Vis.},
  pp.~132--149, 2018.

\bibitem{zhang2017split}
R.~Zhang, P.~Isola, and A.~A. Efros, ``Split-brain autoencoders: Unsupervised
  learning by cross-channel prediction,'' in {\em IEEE Conf. Comput. Vis.
  Pattern Recognit.}, pp.~1058--1067, 2017.

\bibitem{wang2017transitive}
X.~Wang, K.~He, and A.~Gupta, ``Transitive invariance for self-supervised
  visual representation learning,'' in {\em IEEE Int. Conf. Comput. Vis.},
  pp.~1329--1338, 2017.

\bibitem{kolesnikov2019revisiting}
A.~Kolesnikov, X.~Zhai, and L.~Beyer, ``Revisiting self-supervised visual
  representation learning,'' in {\em IEEE Conf. Comput. Vis. Pattern
  Recognit.}, pp.~1920--1929, 2019.

\bibitem{krahenbuhl2018free}
P.~Kr{\"a}henb{\"u}hl, ``Free supervision from video games,'' in {\em IEEE
  Conf. Comput. Vis. Pattern Recognit.}, pp.~2955--2964, 2018.

\bibitem{goodfellow2014generative}
I.~Goodfellow, J.~Pouget-Abadie, M.~Mirza, B.~Xu, D.~Warde-Farley, S.~Ozair,
  A.~Courville, and Y.~Bengio, ``Generative adversarial nets,'' in {\em Neural
  Inf. Process. Syst.}, pp.~2672--2680, 2014.

\bibitem{chen2019self}
T.~Chen, X.~Zhai, M.~Ritter, M.~Lucic, and N.~Houlsby, ``Self-supervised gans
  via auxiliary rotation loss,'' in {\em IEEE Conf. Comput. Vis. Pattern
  Recognit.}, pp.~12154--12163, 2019.

\bibitem{zhai2019s4l}
X.~Zhai, A.~Oliver, A.~Kolesnikov, and L.~Beyer, ``S4l: Self-supervised
  semi-supervised learning,'' in {\em IEEE Int. Conf. Comput. Vis.},
  pp.~1476--1485, 2019.

\bibitem{hendrycks2019using}
D.~Hendrycks, M.~Mazeika, S.~Kadavath, and D.~Song, ``Using self-supervised
  learning can improve model robustness and uncertainty,'' in {\em Neural Inf.
  Process. Syst.}, pp.~15663--15674, 2019.

\bibitem{hassani2020contrastive}
K.~Hassani and A.~H. Khasahmadi, ``Contrastive multi-view representation
  learning on graphs,'' in {\em Int. Conf. Mach. Learn.}, 2020.

\bibitem{gomez2017self}
L.~Gomez, Y.~Patel, M.~Rusi{\~n}ol, D.~Karatzas, and C.~Jawahar,
  ``Self-supervised learning of visual features through embedding images into
  text topic spaces,'' in {\em IEEE Conf. Comput. Vis. Pattern Recognit.},
  pp.~4230--4239, 2017.

\bibitem{jing2020self}
L.~Jing, Y.~Chen, L.~Zhang, M.~He, and Y.~Tian, ``Self-supervised feature
  learning by cross-modality and cross-view correspondences,'' {\em arXiv
  preprint arXiv:2004.05749}, 2020.

\bibitem{jing2020selfM}
L.~Jing, Y.~Chen, L.~Zhang, M.~He, and Y.~Tian, ``Self-supervised modal and
  view invariant feature learning,'' {\em arXiv preprint arXiv:2005.14169},
  2020.

\bibitem{zhang2019unsupervised}
L.~Zhang and Z.~Zhu, ``Unsupervised feature learning for point cloud
  understanding by contrasting and clustering using graph convolutional neural
  networks,'' in {\em International Conference on 3D Vision}, pp.~395--404,
  2019.

\bibitem{yang2018foldingnet}
Y.~Yang, C.~Feng, Y.~Shen, and D.~Tian, ``Foldingnet: Point cloud auto-encoder
  via deep grid deformation,'' in {\em IEEE Conf. Comput. Vis. Pattern
  Recognit.}, pp.~206--215, 2018.

\bibitem{gadelha2018multiresolution}
M.~Gadelha, R.~Wang, and S.~Maji, ``Multiresolution tree networks for 3d point
  cloud processing,'' in {\em Eur. Conf. Comput. Vis.}, pp.~103--118, 2018.

\bibitem{zhao20193d}
Y.~Zhao, T.~Birdal, H.~Deng, and F.~Tombari, ``3d point capsule networks,'' in
  {\em IEEE Conf. Comput. Vis. Pattern Recognit.}, pp.~1009--1018, 2019.

\bibitem{sun2020test}
Y.~Sun, X.~Wang, Z.~Liu, J.~Miller, A.~A. Efros, and M.~Hardt, ``Test-time
  training with self-supervision for generalization under distribution
  shifts,'' in {\em Int. Conf. Mach. Learn.}, 2020.

\bibitem{gandelsman2022test}
Y.~Gandelsman, Y.~Sun, X.~Chen, and A.~A. Efros, ``Test-time training with
  masked autoencoders,'' {\em arXiv preprint arXiv:2209.07522}, 2022.

\bibitem{sun2021task}
J.~J. Sun, A.~Kennedy, E.~Zhan, D.~J. Anderson, Y.~Yue, and P.~Perona, ``Task
  programming: Learning data efficient behavior representations,'' in {\em IEEE
  Conf. Comput. Vis. Pattern Recognit.}, pp.~2876--2885, 2021.

\bibitem{ren2018cross}
Z.~Ren and Y.~Jae~Lee, ``Cross-domain self-supervised multi-task feature
  learning using synthetic imagery,'' in {\em IEEE Conf. Comput. Vis. Pattern
  Recognit.}, pp.~762--771, 2018.

\bibitem{saito2020universal}
K.~Saito, D.~Kim, S.~Sclaroff, and K.~Saenko, ``Universal domain adaptation
  through self supervision,'' in {\em Neural Inf. Process. Syst.}, pp.~1--11,
  2020.

\bibitem{sun2019unsupervised}
Y.~Sun, E.~Tzeng, T.~Darrell, and A.~A. Efros, ``Unsupervised domain adaptation
  through self-supervision,'' {\em arXiv preprint arXiv:1909.11825}, 2019.

\bibitem{noroozi2018boosting}
M.~Noroozi, A.~Vinjimoor, P.~Favaro, and H.~Pirsiavash, ``Boosting
  self-supervised learning via knowledge transfer,'' in {\em IEEE Conf. Comput.
  Vis. Pattern Recognit.}, pp.~9359--9367, 2018.

\bibitem{hu2020gpt}
Z.~Hu, Y.~Dong, K.~Wang, K.-W. Chang, and Y.~Sun, ``Gpt-gnn: Generative
  pre-training of graph neural networks,'' in {\em ACM SIGKDD International
  Conference on Knowledge Discovery and Data Mining}, pp.~1857--1867, 2020.

\bibitem{rong2020self}
Y.~Rong, Y.~Bian, T.~Xu, W.~Xie, Y.~Wei, W.~Huang, and J.~Huang,
  ``Self-supervised graph transformer on large-scale molecular data,'' in {\em
  Neural Inf. Process. Syst.}, 2020.

\bibitem{buchler2018improving}
U.~Buchler, B.~Brattoli, and B.~Ommer, ``Improving spatiotemporal
  self-supervision by deep reinforcement learning,'' in {\em Eur. Conf. Comput.
  Vis.}, pp.~770--786, 2018.

\bibitem{guo2020bootstrap}
D.~Guo, B.~A. Pires, B.~Piot, J.-b. Grill, F.~Altch{\'e}, R.~Munos, and M.~G.
  Azar, ``Bootstrap latent-predictive representations for multitask
  reinforcement learning,'' {\em arXiv preprint arXiv:2004.14646}, 2020.

\bibitem{hansen2020self}
N.~Hansen, Y.~Sun, P.~Abbeel, A.~A. Efros, L.~Pinto, and X.~Wang,
  ``Self-supervised policy adaptation during deployment,'' {\em arXiv preprint
  arXiv:2007.04309}, 2020.

\bibitem{gidaris2019boosting}
S.~Gidaris, A.~Bursuc, N.~Komodakis, P.~P{\'e}rez, and M.~Cord, ``Boosting
  few-shot visual learning with self-supervision,'' in {\em IEEE Int. Conf.
  Comput. Vis.}, pp.~8059--8068, 2019.

\bibitem{su2019boosting}
J.-C. Su, S.~Maji, and B.~Hariharan, ``Boosting supervision with
  self-supervision for few-shot learning,'' {\em arXiv preprint
  arXiv:1906.07079}, 2019.

\bibitem{li2021bossnas}
C.~Li, T.~Tang, G.~Wang, J.~Peng, B.~Wang, X.~Liang, and X.~Chang, ``Bossnas:
  Exploring hybrid cnn-transformers with block-wisely self-supervised neural
  architecture search,'' in {\em IEEE Int. Conf. Comput. Vis.}, 2021.

\bibitem{fan2021does}
L.~Fan, S.~Liu, P.-Y. Chen, G.~Zhang, and C.~Gan, ``When does contrastive
  learning preserve adversarial robustness from pretraining to finetuning?,''
  in {\em Neural Inf. Process. Syst.}, 2021.

\bibitem{kim2020adversarial}
M.~Kim, J.~Tack, and S.~J. Hwang, ``Adversarial self-supervised contrastive
  learning,'' in {\em Neural Inf. Process. Syst.}, pp.~1--12, 2020.

\bibitem{chen2020adversarial}
T.~Chen, S.~Liu, S.~Chang, Y.~Cheng, L.~Amini, and Z.~Wang, ``Adversarial
  robustness: From self-supervised pre-training to fine-tuning,'' in {\em IEEE
  Conf. Comput. Vis. Pattern Recognit.}, pp.~699--708, 2020.

\bibitem{lin2021self}
Y.~Lin, X.~Guo, and Y.~Lu, ``Self-supervised video representation learning with
  meta-contrastive network,'' in {\em IEEE Int. Conf. Comput. Vis.},
  pp.~8239--8249, 2021.

\bibitem{anconditional}
Y.~An, H.~Xue, X.~Zhao, and L.~Zhang, ``Conditional self-supervised learning
  for few-shot classification,'' in {\em Int. Joint Conf. Artif. Intell.},
  pp.~2140--2146, 2021.

\bibitem{pal1978computer}
S.~Pal, A.~Datta, and D.~D. Majumder, ``Computer recognition of vowel sounds
  using a self-supervised learning algorithm,'' {\em Journal of the Anatomical
  Society of India}, pp.~117--123, 1978.

\bibitem{ghosh1993self}
A.~Ghosh, N.~R. Pal, and S.~K. Pal, ``Self-organization for object extraction
  using a multilayer neural network and fuzziness mearsures,'' {\em IEEE
  Transactions on Fuzzy Systems}, pp.~54--68, 1993.

\bibitem{sharma2016vconv}
A.~Sharma, O.~Grau, and M.~Fritz, ``Vconv-dae: Deep volumetric shape learning
  without object labels,'' in {\em Eur. Conf. Comput. Vis.}, pp.~236--250,
  2016.

\bibitem{gong2017look}
K.~Gong, X.~Liang, D.~Zhang, X.~Shen, and L.~Lin, ``Look into person:
  Self-supervised structure-sensitive learning and a new benchmark for human
  parsing,'' in {\em IEEE Conf. Comput. Vis. Pattern Recognit.}, pp.~932--940,
  2017.

\bibitem{liang2018look}
X.~Liang, K.~Gong, X.~Shen, and L.~Lin, ``Look into person: Joint body parsing
  \& pose estimation network and a new benchmark,'' {\em IEEE Trans. Pattern
  Anal. Mach. Intell.}, vol.~41, no.~4, pp.~871--885, 2018.

\bibitem{zhan2020self}
X.~Zhan, X.~Pan, B.~Dai, Z.~Liu, D.~Lin, and C.~C. Loy, ``Self-supervised scene
  de-occlusion,'' in {\em IEEE Conf. Comput. Vis. Pattern Recognit.},
  pp.~3784--3792, 2020.

\bibitem{pathak2017learning}
D.~Pathak, R.~Girshick, P.~Doll{\'a}r, T.~Darrell, and B.~Hariharan, ``Learning
  features by watching objects move,'' in {\em IEEE Conf. Comput. Vis. Pattern
  Recognit.}, pp.~2701--2710, 2017.

\bibitem{wang2020self}
Y.~Wang, J.~Zhang, M.~Kan, S.~Shan, and X.~Chen, ``Self-supervised equivariant
  attention mechanism for weakly supervised semantic segmentation,'' in {\em
  IEEE Conf. Comput. Vis. Pattern Recognit.}, pp.~12275--12284, 2020.

\bibitem{chen2021aggnet}
Z.~Chen, X.~Ye, L.~Du, W.~Yang, L.~Huang, X.~Tan, Z.~Shi, F.~Shen, and E.~Ding,
  ``Aggnet for self-supervised monocular depth estimation: Go an aggressive
  step furthe,'' in {\em ACM Int. Conf. Multimedia}, pp.~1526--1534, 2021.

\bibitem{chen2021ice}
H.~Chen, B.~Lagadec, and F.~Bremond, ``Ice: Inter-instance contrastive encoding
  for unsupervised person re-identification,'' in {\em IEEE Int. Conf. Comput.
  Vis.}, pp.~14960--14969, 2021.

\bibitem{isobe2021towards}
T.~Isobe, D.~Li, L.~Tian, W.~Chen, Y.~Shan, and S.~Wang, ``Towards
  discriminative representation learning for unsupervised person
  re-identification,'' in {\em IEEE Int. Conf. Comput. Vis.}, pp.~8526--8536,
  2021.

\bibitem{li2020self}
S.~Li, X.~Wang, Y.~Cao, F.~Xue, Z.~Yan, and H.~Zha, ``Self-supervised deep
  visual odometry with online adaptation,'' in {\em IEEE Conf. Comput. Vis.
  Pattern Recognit.}, pp.~6339--6348, 2020.

\bibitem{wupointpwc}
W.~Wu, Z.~Y. Wang, Z.~Li, W.~Liu, and L.~Fuxin, ``Pointpwc-net: Cost volume on
  point clouds for (self-) supervised scene flow estimation,'' in {\em Eur.
  Conf. Comput. Vis.}, 2020.

\bibitem{xu2020knowledge}
G.~Xu, Z.~Liu, X.~Li, and C.~C. Loy, ``Knowledge distillation meets
  self-supervision,'' {\em arXiv preprint arXiv:2006.07114}, 2020.

\bibitem{walker2015dense}
J.~Walker, A.~Gupta, and M.~Hebert, ``Dense optical flow prediction from a
  static image,'' in {\em IEEE Int. Conf. Comput. Vis.}, pp.~2443--2451, 2015.

\bibitem{zhu2020vision}
F.~Zhu, Y.~Zhu, X.~Chang, and X.~Liang, ``Vision-language navigation with
  self-supervised auxiliary reasoning tasks,'' in {\em IEEE Conf. Comput. Vis.
  Pattern Recognit.}, pp.~10012--10022, 2020.

\bibitem{niu2020rhythmnet}
X.~Niu, S.~Shan, H.~Han, and X.~Chen, ``Rhythmnet: End-to-end heart rate
  estimation from face via spatial-temporal representation,'' {\em IEEE Trans.
  Image Process.}, vol.~29, pp.~2409--2423, 2020.

\bibitem{niu2020video}
X.~Niu, Z.~Yu, H.~Han, X.~Li, S.~Shan, and G.~Zhao, ``Video-based remote
  physiological measurement via cross-verified feature disentangling,'' in {\em
  Eur. Conf. Comput. Vis.}, 2020.

\bibitem{xie2020noise2same}
Y.~Xie, Z.~Wang, and S.~Ji, ``Noise2same: Optimizing a self-supervised bound
  for image denoising,'' in {\em Neural Inf. Process. Syst.}, 2020.

\bibitem{huang2021neighbor2neighbor}
T.~Huang, S.~Li, X.~Jia, H.~Lu, and J.~Liu, ``Neighbor2neighbor:
  Self-supervised denoising from single noisy images,'' in {\em IEEE Conf.
  Comput. Vis. Pattern Recognit.}, 2021.

\bibitem{yang2021instance}
C.~Yang, Z.~Wu, B.~Zhou, and S.~Lin, ``Instance localization for
  self-supervised detection pretraining,'' in {\em IEEE Conf. Comput. Vis.
  Pattern Recognit.}, pp.~3987--3996, 2021.

\bibitem{croitoru2017unsupervised}
I.~Croitoru, S.-V. Bogolin, and M.~Leordeanu, ``Unsupervised learning from
  video to detect foreground objects in single images,'' in {\em IEEE Int.
  Conf. Comput. Vis.}, pp.~4335--4343, 2017.

\bibitem{xie2021detco}
E.~Xie, J.~Ding, W.~Wang, X.~Zhan, H.~Xu, Z.~Li, and P.~Luo, ``Detco:
  Unsupervised contrastive learning for object detection,'' {\em arXiv preprint
  arXiv:2102.04803}, 2021.

\bibitem{wu2021practical}
G.~Wu, J.~Jiang, X.~Liu, and J.~Ma, ``A practical contrastive learning
  framework for single image super-resolution,'' {\em arXiv preprint
  arXiv:2111.13924}, 2021.

\bibitem{menon2020pulse}
S.~Menon, A.~Damian, S.~Hu, N.~Ravi, and C.~Rudin, ``Pulse: Self-supervised
  photo upsampling via latent space exploration of generative models,'' in {\em
  IEEE Conf. Comput. Vis. Pattern Recognit.}, pp.~2437--2445, 2020.

\bibitem{girdhar2016learning}
R.~Girdhar, D.~F. Fouhey, M.~Rodriguez, and A.~Gupta, ``Learning a predictable
  and generative vector representation for objects,'' in {\em Eur. Conf.
  Comput. Vis.}, pp.~484--499, 2016.

\bibitem{jayaraman2015learning}
D.~Jayaraman and K.~Grauman, ``Learning image representations tied to
  ego-motion,'' in {\em IEEE Int. Conf. Comput. Vis.}, pp.~1413--1421, 2015.

\bibitem{yin2018geonet}
Z.~Yin and J.~Shi, ``Geonet: Unsupervised learning of dense depth, optical flow
  and camera pose,'' in {\em IEEE Conf. Comput. Vis. Pattern Recognit.},
  pp.~1983--1992, 2018.

\bibitem{huang2021self}
L.~Huang, Y.~Liu, B.~Wang, P.~Pan, Y.~Xu, and R.~Jin, ``Self-supervised video
  representation learning by context and motion decoupling,'' in {\em IEEE
  Conf. Comput. Vis. Pattern Recognit.}, pp.~13886--13895, 2021.

\bibitem{hu2021contrast}
K.~Hu, J.~Shao, Y.~Liu, B.~Raj, M.~Savvides, and Z.~Shen, ``Contrast and order
  representations for video self-supervised learning,'' in {\em IEEE Int. Conf.
  Comput. Vis.}, pp.~7939--7949, 2021.

\bibitem{tschannen2020self}
M.~Tschannen, J.~Djolonga, M.~Ritter, A.~Mahendran, N.~Houlsby, S.~Gelly, and
  M.~Lucic, ``Self-supervised learning of video-induced visual invariances,''
  in {\em IEEE Conf. Comput. Vis. Pattern Recognit.}, pp.~13806--13815, 2020.

\bibitem{He2022Learn}
X.~He, Y.~Pan, M.~Tang, Y.~Lv, and Y.~Peng, ``Learn from unlabeled videos for
  near-duplicate video retrieval,'' in {\em International Conference on
  Research on Development in Information Retrieval}, pp.~1--10, 2022.

\bibitem{han2019video}
T.~Han, W.~Xie, and A.~Zisserman, ``Video representation learning by dense
  predictive coding,'' in {\em ICCV Workshops}, 2019.

\bibitem{han2020memory}
T.~Han, W.~Xie, and A.~Zisserman, ``Memory-augmented dense predictive coding
  for video representation learning,'' in {\em Eur. Conf. Comput. Vis.}, 2020.

\bibitem{fernando2017self}
B.~Fernando, H.~Bilen, E.~Gavves, and S.~Gould, ``Self-supervised video
  representation learning with odd-one-out networks,'' in {\em IEEE Conf.
  Comput. Vis. Pattern Recognit.}, pp.~3636--3645, 2017.

\bibitem{lee2017unsupervised}
H.-Y. Lee, J.-B. Huang, M.~Singh, and M.-H. Yang, ``Unsupervised representation
  learning by sorting sequences,'' in {\em IEEE Int. Conf. Comput. Vis.},
  pp.~667--676, 2017.

\bibitem{xu2019self}
D.~Xu, J.~Xiao, Z.~Zhao, J.~Shao, D.~Xie, and Y.~Zhuang, ``Self-supervised
  spatiotemporal learning via video clip order prediction,'' in {\em IEEE Conf.
  Comput. Vis. Pattern Recognit.}, pp.~10334--10343, 2019.

\bibitem{benaim2020speednet}
S.~Benaim, A.~Ephrat, O.~Lang, I.~Mosseri, W.~T. Freeman, M.~Rubinstein,
  M.~Irani, and T.~Dekel, ``Speednet: Learning the speediness in videos,'' in
  {\em IEEE Conf. Comput. Vis. Pattern Recognit.}, pp.~9922--9931, 2020.

\bibitem{yao2020video}
Y.~Yao, C.~Liu, D.~Luo, Y.~Zhou, and Q.~Ye, ``Video playback rate perception
  for self-supervised spatio-temporal representation learning,'' in {\em IEEE
  Conf. Comput. Vis. Pattern Recognit.}, pp.~6548--6557, 2020.

\bibitem{wang2020selfV}
J.~Wang, J.~Jiao, and Y.-H. Liu, ``Self-supervised video representation
  learning by pace prediction,'' in {\em Eur. Conf. Comput. Vis.}, 2020.

\bibitem{diba2019dynamonet}
A.~Diba, V.~Sharma, L.~V. Gool, and R.~Stiefelhagen, ``Dynamonet: Dynamic
  action and motion network,'' in {\em IEEE Int. Conf. Comput. Vis.},
  pp.~6192--6201, 2019.

\bibitem{han2020Self}
T.~Han, W.~Xie, and A.~Zisserman, ``Self-supervised co-training for video
  representation learning,'' in {\em Neural Inf. Process. Syst.}, pp.~1--12,
  2020.

\bibitem{korbar2018cooperative}
B.~Korbar, D.~Tran, and L.~Torresani, ``Cooperative learning of audio and video
  models from self-supervised synchronization,'' in {\em Neural Inf. Process.
  Syst.}, pp.~7763--7774, 2018.

\bibitem{arandjelovic2017look}
R.~Arandjelovic and A.~Zisserman, ``Look, listen and learn,'' in {\em IEEE Int.
  Conf. Comput. Vis.}, pp.~609--617, 2017.

\bibitem{sun2019videobert}
C.~Sun, A.~Myers, C.~Vondrick, K.~Murphy, and C.~Schmid, ``Videobert: A joint
  model for video and language representation learning,'' in {\em IEEE Int.
  Conf. Comput. Vis.}, pp.~7464--7473, 2019.

\bibitem{nagrani2020speech2action}
A.~Nagrani, C.~Sun, D.~Ross, R.~Sukthankar, C.~Schmid, and A.~Zisserman,
  ``Speech2action: Cross-modal supervision for action recognition,'' in {\em
  IEEE Conf. Comput. Vis. Pattern Recognit.}, pp.~10317--10326, 2020.

\bibitem{stroud2020learning}
J.~C. Stroud, D.~A. Ross, C.~Sun, J.~Deng, R.~Sukthankar, and C.~Schmid,
  ``Learning video representations from textual web supervision,'' {\em arXiv
  preprint arXiv:2007.14937}, 2020.

\bibitem{alayrac2020self}
J.-B. Alayrac, A.~Recasens, R.~Schneider, R.~Arandjelovi{\'c}, J.~Ramapuram,
  J.~De~Fauw, L.~Smaira, S.~Dieleman, and A.~Zisserman, ``Self-supervised
  multimodal versatile networks,'' {\em arXiv preprint arXiv:2006.16228}, 2020.

\bibitem{sermanet2018time}
P.~Sermanet, C.~Lynch, Y.~Chebotar, J.~Hsu, E.~Jang, S.~Schaal, and S.~Levine,
  ``Time-contrastive networks: Self-supervised learning from video,'' in {\em
  IEEE Int. Conf. Robot. Autom.}, pp.~1134--1141, 2018.

\bibitem{wang2019learning}
X.~Wang, A.~Jabri, and A.~A. Efros, ``Learning correspondence from the
  cycle-consistency of time,'' in {\em IEEE Conf. Comput. Vis. Pattern
  Recognit.}, pp.~2566--2576, 2019.

\bibitem{li2019joint}
X.~Li, S.~Liu, S.~De~Mello, X.~Wang, J.~Kautz, and M.-H. Yang, ``Joint-task
  self-supervised learning for temporal correspondence,'' in {\em Neural Inf.
  Process. Syst.}, pp.~318--328, 2019.

\bibitem{jabri2020space}
A.~Jabri, A.~Owens, and A.~A. Efros, ``Space-time correspondence as a
  contrastive random walk,'' in {\em Neural Inf. Process. Syst.},
  pp.~19545--19560, 2020.

\bibitem{lai2020mast}
Z.~Lai, E.~Lu, and W.~Xie, ``Mast: A memory-augmented self-supervised
  tracker,'' in {\em IEEE Conf. Comput. Vis. Pattern Recognit.},
  pp.~6479--6488, 2020.

\bibitem{zhang2020online}
Z.~Zhang, S.~Lathuiliere, E.~Ricci, N.~Sebe, Y.~Yan, and J.~Yang, ``Online
  depth learning against forgetting in monocular videos,'' in {\em IEEE Conf.
  Comput. Vis. Pattern Recognit.}, pp.~4494--4503, 2020.

\bibitem{luo2020video}
D.~Luo, C.~Liu, Y.~Zhou, D.~Yang, C.~Ma, Q.~Ye, and W.~Wang, ``Video cloze
  procedure for self-supervised spatio-temporal learning,'' in {\em AAAI
  Conf.Artif. Intell.}, pp.~11701--11708, 2020.

\bibitem{henaff2019data}
O.~J. H{\'e}naff, A.~Srinivas, J.~De~Fauw, A.~Razavi, C.~Doersch, S.~Eslami,
  and A.~v.~d. Oord, ``Data-efficient image recognition with contrastive
  predictive coding,'' in {\em Int. Conf. Mach. Learn.}, 2020.

\bibitem{radford2018improving}
A.~Radford, K.~Narasimhan, T.~Salimans, and I.~Sutskever, ``Improving language
  understanding by generative pre-training,'' 2018.

\bibitem{li2021efficient}
C.~Li, J.~Yang, P.~Zhang, M.~Gao, B.~Xiao, X.~Dai, L.~Yuan, and J.~Gao,
  ``Efficient self-supervised vision transformers for representation
  learning,'' {\em arXiv preprint arXiv:2106.09785}, 2021.

\bibitem{mikolov2013distributed}
T.~Mikolov, I.~Sutskever, K.~Chen, G.~S. Corrado, and J.~Dean, ``Distributed
  representations of words and phrases and their compositionality,'' in {\em
  Neural Inf. Process. Syst.}, pp.~3111--3119, 2013.

\bibitem{clark2020electra}
K.~Clark, M.-T. Luong, Q.~V. Le, and C.~D. Manning, ``Electra: Pre-training
  text encoders as discriminators rather than generators,'' in {\em Int. Conf.
  Learn. Represent.}, 2020.

\bibitem{pappas2019gile}
N.~Pappas and J.~Henderson, ``Gile: A generalized input-label embedding for
  text classification,'' {\em Transactions of the Association for Computational
  Linguistics}, vol.~7, pp.~139--155, 2019.

\bibitem{clark2020pre}
K.~Clark, M.-T. Luong, Q.~V. Le, and C.~D. Manning, ``Pre-training transformers
  as energy-based cloze models,'' {\em arXiv preprint arXiv:2012.08561}, 2020.

\bibitem{wu2020clear}
Z.~Wu, S.~Wang, J.~Gu, M.~Khabsa, F.~Sun, and H.~Ma, ``Clear: Contrastive
  learning for sentence representation,'' {\em arXiv preprint
  arXiv:2012.15466}, 2020.

\bibitem{giorgi2020declutr}
J.~Giorgi, O.~Nitski, B.~Wang, and G.~Bader, ``Declutr: Deep contrastive
  learning for unsupervised textual representations,'' {\em arXiv preprint
  arXiv:2006.03659}, 2020.

\bibitem{zhou2021preservational}
H.-Y. Zhou, C.~Lu, S.~Yang, X.~Han, and Y.~Yu, ``Preservational learning
  improves self-supervised medical image models by reconstructing diverse
  contexts,'' in {\em IEEE Int. Conf. Comput. Vis.}, pp.~3499--3509, 2021.

\bibitem{chaitanya2020contrastive}
K.~Chaitanya, E.~Erdil, N.~Karani, and E.~Konukoglu, ``Contrastive learning of
  global and local features for medical image segmentation with limited
  annotations,'' in {\em Neural Inf. Process. Syst.}, 2020.

\bibitem{zhu2020rubik}
J.~Zhu, Y.~Li, Y.~Hu, K.~Ma, S.~K. Zhou, and Y.~Zheng, ``Rubik’s cube+: A
  self-supervised feature learning framework for 3d medical image analysis,''
  {\em Medical Image Analysis}, p.~101746, 2020.

\bibitem{manas2021seasonal}
O.~Manas, A.~Lacoste, X.~Gir{\'o}-i Nieto, D.~Vazquez, and P.~Rodriguez,
  ``Seasonal contrast: Unsupervised pre-training from uncurated remote sensing
  data,'' in {\em IEEE Int. Conf. Comput. Vis.}, pp.~9414--9423, 2021.

\bibitem{wang2022advancing}
D.~Wang, Q.~Zhang, Y.~Xu, J.~Zhang, B.~Du, D.~Tao, and L.~Zhang, ``Advancing
  plain vision transformer toward remote sensing foundation model,'' {\em IEEE
  Trans. Geoscience and Remote Sensing}, vol.~61, pp.~1--15, 2022.

\bibitem{Liu2022MixMIM}
J.~Liu, X.~Huang, Y.~Liu, and H.~Li, ``Mixmim: Mixed and masked image modeling
  for efficient visual representation learning,'' {\em arXiv preprint
  arXiv:2205.13137}, 2022.

\bibitem{bau2017network}
D.~Bau, B.~Zhou, A.~Khosla, A.~Oliva, and A.~Torralba, ``Network dissection:
  Quantifying interpretability of deep visual representations,'' in {\em IEEE
  Conf. Comput. Vis. Pattern Recognit.}, pp.~6541--6549, 2017.

\bibitem{garrido2023RankMe}
Q.~Garrido, R.~Balestriero, L.~Najman, and Y.~Lecun, ``Rankme: Assessing the
  downstream performance of pretrained self-supervised representations by their
  rank,'' in {\em Int. Conf. Mach. Learn.}, pp.~10929--10974, PMLR, July 2023.

\bibitem{everingham2010pascal}
M.~Everingham, L.~Van~Gool, C.~K. Williams, J.~Winn, and A.~Zisserman, ``The
  pascal visual object classes (voc) challenge,'' {\em Int. J. Comput. Vis.},
  vol.~88, pp.~303--338, 2010.

\bibitem{lin2015microsoft}
T.-Y. Lin, M.~Maire, S.~Belongie, L.~Bourdev, R.~Girshick, J.~Hays, P.~Perona,
  D.~Ramanan, C.~L. Zitnick, and P.~Dollár, ``Microsoft coco: Common objects
  in context,'' 2015.

\bibitem{zhou2017scene}
B.~Zhou, H.~Zhao, X.~Puig, S.~Fidler, A.~Barriuso, and A.~Torralba, ``Scene
  parsing through ade20k dataset,'' in {\em IEEE Conf. Comput. Vis. Pattern
  Recognit.}, 2017.

\bibitem{zhou2019semantic}
B.~Zhou, H.~Zhao, X.~Puig, T.~Xiao, S.~Fidler, A.~Barriuso, and A.~Torralba,
  ``Semantic understanding of scenes through the ade20k dataset,'' {\em Int. J.
  Comput. Vis.}, vol.~127, no.~3, pp.~302--321, 2019.

\bibitem{kay2017kinetics}
W.~Kay, J.~Carreira, K.~Simonyan, B.~Zhang, C.~Hillier, S.~Vijayanarasimhan,
  F.~Viola, T.~Green, T.~Back, P.~Natsev, {\em et~al.}, ``The kinetics human
  action video dataset,'' {\em arXiv preprint arXiv:1705.06950}, 2017.

\bibitem{goyal2017something}
R.~Goyal, S.~Ebrahimi~Kahou, V.~Michalski, J.~Materzynska, S.~Westphal, H.~Kim,
  V.~Haenel, I.~Fruend, P.~Yianilos, M.~Mueller-Freitag, {\em et~al.}, ``The"
  something something" video database for learning and evaluating visual common
  sense,'' in {\em IEEE Int. Conf. Comput. Vis.}, pp.~5842--5850, 2017.

\bibitem{gu2018ava}
C.~Gu, C.~Sun, D.~A. Ross, C.~Vondrick, C.~Pantofaru, Y.~Li,
  S.~Vijayanarasimhan, G.~Toderici, S.~Ricco, R.~Sukthankar, {\em et~al.},
  ``Ava: A video dataset of spatio-temporally localized atomic visual
  actions,'' in {\em IEEE Conf, ComputVis.Pattern Recognit.}, pp.~6047--6056,
  2018.

\bibitem{soomro2012ucf101}
K.~Soomro, A.~R. Zamir, and M.~Shah, ``Ucf101: A dataset of 101 human actions
  classes from videos in the wild,'' {\em arXiv preprint arXiv:1212.0402},
  2012.

\bibitem{kuehne2011hmdb}
H.~Kuehne, H.~Jhuang, E.~Garrote, T.~Poggio, and T.~Serre, ``Hmdb: a large
  video database for human motion recognition,'' in {\em IEEE Int. Conf.
  Comput. Vis.}, pp.~2556--2563, IEEE, 2011.

\bibitem{wang2021enhancing}
J.~Wang, Y.~Gao, K.~Li, J.~Hu, X.~Jiang, X.~Guo, R.~Ji, and X.~Sun, ``Enhancing
  unsupervised video representation learning by decoupling the scene and the
  motion,'' in {\em Proceedings of the AAAI Conference on Artificial
  Intelligence}, vol.~35, pp.~10129--10137, 2021.

\bibitem{knights2021temporally}
J.~Knights, B.~Harwood, D.~Ward, A.~Vanderkop, O.~Mackenzie-Ross, and
  P.~Moghadam, ``Temporally coherent embeddings for self-supervised video
  representation learning,'' in {\em 2020 25th International Conference on
  Pattern Recognition (ICPR)}, pp.~8914--8921, IEEE, 2021.

\bibitem{recasens2021broaden}
A.~Recasens, P.~Luc, J.-B. Alayrac, L.~Wang, F.~Strub, C.~Tallec,
  M.~Malinowski, V.~P{\u{a}}tr{\u{a}}ucean, F.~Altch{\'e}, M.~Valko, {\em
  et~al.}, ``Broaden your views for self-supervised video learning,'' in {\em
  IEEE Int. Conf. Comput. Vis.}, pp.~1255--1265, 2021.

\bibitem{yang2020video}
C.~Yang, Y.~Xu, B.~Dai, and B.~Zhou, ``Video representation learning with
  visual tempo consistency,'' {\em arXiv preprint arXiv:2006.15489}, 2020.

\bibitem{feichtenhofer2021large}
C.~Feichtenhofer, H.~Fan, B.~Xiong, R.~Girshick, and K.~He, ``A large-scale
  study on unsupervised spatiotemporal representation learning,'' in {\em
  Proceedings of the IEEE Conf. Comput. Vis. Pattern Recognit.},
  pp.~3299--3309, 2021.

\bibitem{qian2021spatiotemporal}
R.~Qian, T.~Meng, B.~Gong, M.-H. Yang, H.~Wang, S.~Belongie, and Y.~Cui,
  ``Spatiotemporal contrastive video representation learning,'' in {\em
  Proceedings of the IEEE Conf. Comput. Vis. Pattern Recognit.},
  pp.~6964--6974, 2021.

\bibitem{robinson2021can}
J.~Robinson, L.~Sun, K.~Yu, K.~Batmanghelich, S.~Jegelka, and S.~Sra, ``Can
  contrastive learning avoid shortcut solutions?,'' in {\em Neural Inf.
  Process. Syst.}, pp.~4974--4986, 2021.

\bibitem{wei2022contrastive}
Y.~Wei, H.~Hu, Z.~Xie, Z.~Zhang, Y.~Cao, J.~Bao, D.~Chen, and B.~Guo,
  ``Contrastive learning rivals masked image modeling in fine-tuning via
  feature distillation,'' {\em arXiv preprint arXiv:2205.14141}, 2022.

\bibitem{chen2021Intriguing}
T.~Chen, C.~Luo, and L.~Li, ``Intriguing properties of contrastive losses,'' in
  {\em Neural Inf. Process. Syst.}, vol.~34, pp.~11834--11845, Curran
  Associates, Inc., 2021.

\bibitem{tian2021understanding}
Y.~Tian, X.~Chen, and S.~Ganguli, ``Understanding self-supervised learning
  dynamics without contrastive pairs,'' in {\em Int. Conf. Mach. Learn.},
  pp.~10268--10278, 2021.

\bibitem{garrido2023Duality}
Q.~Garrido, Y.~Chen, A.~Bardes, L.~Najman, and Y.~LeCun, ``On the duality
  between contrastive and non-contrastive self-supervised learning,'' in {\em
  Int. Conf. Learn. Represent.}, 2023.

\bibitem{lavoie2023simplicial}
S.~Lavoie, C.~Tsirigotis, M.~Schwarzer, A.~Vani, M.~Noukhovitch, K.~Kawaguchi,
  and A.~Courville, ``Simplicial embeddings in self-supervised learning and
  downstream classification,'' in {\em Int. Conf. Learn. Represent.}, 2023.

\bibitem{tao2022Exploring}
C.~Tao, H.~Wang, X.~Zhu, J.~Dong, S.~Song, G.~Huang, and J.~Dai, ``Exploring
  the equivalence of siamese self-supervised learning via a unified gradient
  framework,'' in {\em Proceedings of the IEEE Conf. Comput. Vis. Pattern
  Recognit.}, pp.~14431--14440, 2022.

\bibitem{wang2021dense}
X.~Wang, R.~Zhang, C.~Shen, T.~Kong, and L.~Li, ``Dense contrastive learning
  for self-supervised visual pre-training,'' in {\em IEEE Conf. Comput. Vis.
  Pattern Recognit.}, pp.~3024--3033, 2021.

\bibitem{wang2022image}
W.~Wang, H.~Bao, L.~Dong, J.~Bjorck, Z.~Peng, Q.~Liu, K.~Aggarwal, O.~K.
  Mohammed, S.~Singhal, S.~Som, {\em et~al.}, ``Image as a foreign language:
  Beit pretraining for all vision and vision-language tasks,'' {\em arXiv
  preprint arXiv:2208.10442}, 2022.

\bibitem{newell2020useful}
A.~Newell and J.~Deng, ``How useful is self-supervised pretraining for visual
  tasks?,'' in {\em IEEE Conf. Comput. Vis. Pattern Recognit.}, pp.~7345--7354,
  2020.

\bibitem{chicco2021siamese}
D.~Chicco, ``Siamese neural networks: An overview,'' {\em Artificial neural
  networks}, pp.~73--94, 2021.

\bibitem{cao2020parametric}
Y.~Cao, Z.~Xie, B.~Liu, Y.~Lin, Z.~Zhang, and H.~Hu, ``Parametric instance
  classification for unsupervised visual feature learning,'' in {\em Neural
  Inf. Process. Syst.}, pp.~1--11, 2020.

\bibitem{hou2022milan}
Z.~Hou, F.~Sun, Y.-K. Chen, Y.~Xie, and S.-Y. Kung, ``Milan: Masked image
  pretraining on language assisted representation,'' {\em arXiv preprint
  arXiv:2208.06049}, 2022.

\bibitem{fang2023EVA}
Y.~Fang, W.~Wang, B.~Xie, Q.~Sun, L.~Wu, X.~Wang, T.~Huang, X.~Wang, and
  Y.~Cao, ``Eva: Exploring the limits of masked visual representation learning
  at scale,'' in {\em Proceedings of the IEEE Conf. Comput. Vis. Pattern
  Recognit.}, pp.~19358--19369, 2023.

\end{thebibliography}

%

 
\clearpage
\appendix

\section{Analysis of CL}

\textbf{Connection to principal component analysis}: Tian \cite{tian2022deep} demonstrated that CL with loss functions like InfoNCE can be formulated as a max-min problem. The max function aims to maximize the contrast between feature representations, while the min function assigns weights to pairs of examples with similar representations. In the context of deep linear networks, Tian showed that the max function in representation learning is equivalent to principal component analysis (PCA), and most local minima correspond to global minima, thus recovering optimal PCA solutions. Experimental results revealed that this formulation, when extended to include new contrastive losses beyond InfoNCE, achieves comparable or even superior performance on datasets like STL-10 and CIFAR10. Furthermore, Tian extended his theoretical analysis to 2-layer rectified linear unit (ReLU) networks, emphasizing the substantial differences between linear and nonlinear scenarios and highlighting the essential role of data augmentation during the training process. It is noteworthy that PCA aims to maximize the inter-example distances within a low-dimensional subspace, making it a specific instance of instance discrimination.

\textbf{Connection to spectral clustering}: 
Chen et al. \cite{haochen2021provable} established a connection between CL and spectral clustering, showing that the representations obtained from CL correspond to embeddings of a positive pair graph in spectral clustering. Specifically, the authors introduced a population augmentation graph, where nodes represent augmented data from the population distribution, and the presence of an edge between nodes is determined by whether they originate from the same original example. Their key assumption is that different classes exhibit only a limited number of connections, resulting in a sparser partition for such a graph. Empirical evidence has confirmed this characteristic, illustrating the data continuity within the same class \cite{wei2020theoretical}.

Specifically, spectral decomposition is employed on the adjacency matrix to construct a matrix, where each row denotes the representation of an example. Through a linear transformation, they demonstrated that the corresponding feature extractor could be retrieved by minimizing an unconventional contrastive loss given as
\begin{eqnarray}\label{equ:14_0}
  \begin{array}{l}
\mathcal{L}(f)=-2 \cdot \mathbb{E}_{x, x^{+}}\left[f(x)^{\top} f\left(x^{+}\right)\right]\\+\mathbb{E}_{x, x^{\prime}}\left[\left(f(x)^{\top} f\left(x^{\prime}\right)\right)^{2}\right],
\end{array}
\end{eqnarray}
where $\left(x, x^{+}\right)$ is a pair of augmentations of the same data, $\left(x, x^{\prime}\right)$ is a pair of independently random augmented data, and $f$ is a parameterized function from augmented data to $\mathbb{R}^k$.
It is worth noting that in cases where the dimensionality of the representation surpasses the maximum count of disjoint subgraphs, the utilization of learned representations in linear classification is guaranteed to yield minimal error. 

\textbf{Connection to supervised learning}: 
Recent research has highlighted the remarkable efficacy of self-supervised pre-training using CL for downstream tasks involving categorization. However, its effectiveness may vary when applied to other task domains. Thus, there is a compelling need to investigate the potential of contrastive pre-training in augmenting supervised learning, particularly in terms of surpassing the accuracy achieved through traditional supervised learning.

Newell et al. \cite{newell2020useful} conducted a comprehensive investigation into the potential effects of pre-training on model performance. Their study explored three key hypotheses as follows. Firstly, whether pre-training consistently leads to performance improvements. Secondly, whether pre-training achieves higher accuracy when faced with limited labeled data, but eventually levels off at a performance comparable to the baseline when sufficient labeled data is available. Thirdly, whether pre-training converges to baseline performance before reaching its plateau in accuracy. To address these hypotheses, the authors conducted experiments on the synthetic COCO dataset with rendering, allowing for the availability of a large number of labels. The results revealed that self-supervised pre-training adheres to the assumption outlined in the third hypothesis. This suggests that SSL does not surpass supervised learning in terms of learning capability, but does perform effectively when dealing with limited labeled data.

\section{Complexity Analysis}

Table \ref{tab:complexity} shows the computational complexity of different SSL methods.
Note that the primary sources of time complexity and memory consumption are the neural network other than SSL components, e.g., the calculation of the cross-correlation matrix in Barlow Twins. Hence, we categorize the SSL methods into different groups based on the architecture.

\begin{table}[ht]
  \centering
  \caption{Summary of the computational complexity of different SSL methods. We categorize the SSL methods into different groups based on the architecture. We denote methods that require large batch training or employ multi-crop with an asterisk (*), which demands substantial memory resources.
Here, \textbf{Encoder Only} refers to models with only an encoder, \textbf{Encoder \& Decoder} denotes models employing a decoder with significant computational load, \textbf{Encoder \& Tokenizer} indicates the use 
of an additional encoder for obtaining semantic tokens, and \textbf{Momentum Encoder} refers to one branch of the siamese model being an exponentially moving average version of the other. \textbf{Siamese Model} essentially refers to two identical neural networks \cite{chicco2021siamese}. In terms of computational load, the hierarchy is as follows: Encoder Only $<$  Encoder \& Decoder $<$ Momentum Encoder $\approx$ Encoder \& Tokenizer $<$ Siamese Model.}
\resizebox{0.5\textwidth}{!}{
\begin{tabular}{c|c|ll}
  \hline
  \textbf{Complexity} & \textbf{Tower Type} & \textbf{Model Type} & \textbf{Methods} \\
  \hline
  \multirow{23}{*}{$\Bigg\downarrow$} & \multirow{8}{*}{one-tower model} & \multirow{7}{*}{Encoder Only} & Jigsaw \cite{noroozi2016unsupervised} \\
  & & & Colorization \cite{zhang2016colorful} \\
  & & & Rotation \cite{gidaris2018unsupervised} \\
  & & & Examplar \cite{dosovitskiy2014discriminative} \\
  & & & Instdisc \cite{wu2018unsupervised} \\
  & & & PIC \cite{cao2020parametric} \\
  & & & SimMIM \cite{xie2021simmim} \\
  \cline{3-4}
  & & \multirow{1}{*}{Encoder \& Decoder} & MAE \cite{he2021masked} \\
  \cline{2-4}
  & \multirow{17}{*}{dual-tower model} & \multirow{4}{*}{Encoder \& Tokenizer} & BEiT \cite{bao2021beit} \\
  & & & BEiT v2 \cite{peng2022beitv2} \\
  & & & MILAN \cite{hou2022milan} \\
  & & & EVA \cite{fang2023EVA} \\
  \cline{3-4}
  &  & \multirow{4}{*}{Momentum Encoder} & MoCo v1 \cite{he2020momentum} \\
  & & & BYOL \cite{grill2020bootstrap} \\
  & & & DINO* \cite{caron2021emerging} \\
  & & & MoCo v2 \cite{chen2020improved} \\
  \cline{3-4}
  &  & \multirow{9}{*}{Siamese Model \cite{chicco2021siamese}} & SimCLR* \cite{chen2020simple} \\
  & & & MoCo v3* \cite{chen2021empirical} \\
  & & & SwAV* \cite{caron2020unsupervised} \\
  & & & SimSiam \cite{chen2020exploring} \\
  & & & Barlow Twins \cite{zbontar2021barlow} \\
  & & & VICReg \cite{bardes2021vicreg} \\
  & & & data2vec \cite{baevski2022data2vec} \\
  & & & iBOT \cite{zhou2021ibot} \\
  \hline
\end{tabular}
}
  \label{tab:complexity}
\end{table}

\begin{IEEEbiography}[{\includegraphics[width=1in,height=1.25in,clip,keepaspectratio]{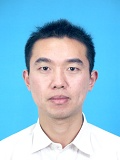}}]{Jie Gui} (SM'16) is currently a professor at the School of Cyber Science and Engineering, Southeast University. He received a BS degree in Computer Science from Hohai University, Nanjing, China, in 2004, an MS degree in Computer Applied Technology from the Hefei Institutes of Physical Science, Chinese Academy of Sciences, Hefei, China, in 2007, and a PhD degree in Pattern Recognition and Intelligent Systems from the University of Science and Technology of China, Hefei, China, in 2010. He has published more than 60 papers in international journals and conferences such as IEEE TPAMI, IEEE TNNLS, IEEE TCYB, IEEE TIP, IEEE TCSVT, IEEE TSMCS, KDD, and ACM MM. He is the Area Chair, Senior PC Member, or PC Member of many conferences such as NeurIPS and ICML. He is an Associate Editor of IEEE Transactions on Circuits and Systems for Video Technology (T-CSVT), Artificial Intelligence Review, Neural Networks, and Neurocomputing. His research interests include machine learning, pattern recognition, and image processing.
\end{IEEEbiography}

\begin{IEEEbiography}[{\includegraphics[width=1in,height=1.25in,clip,keepaspectratio]{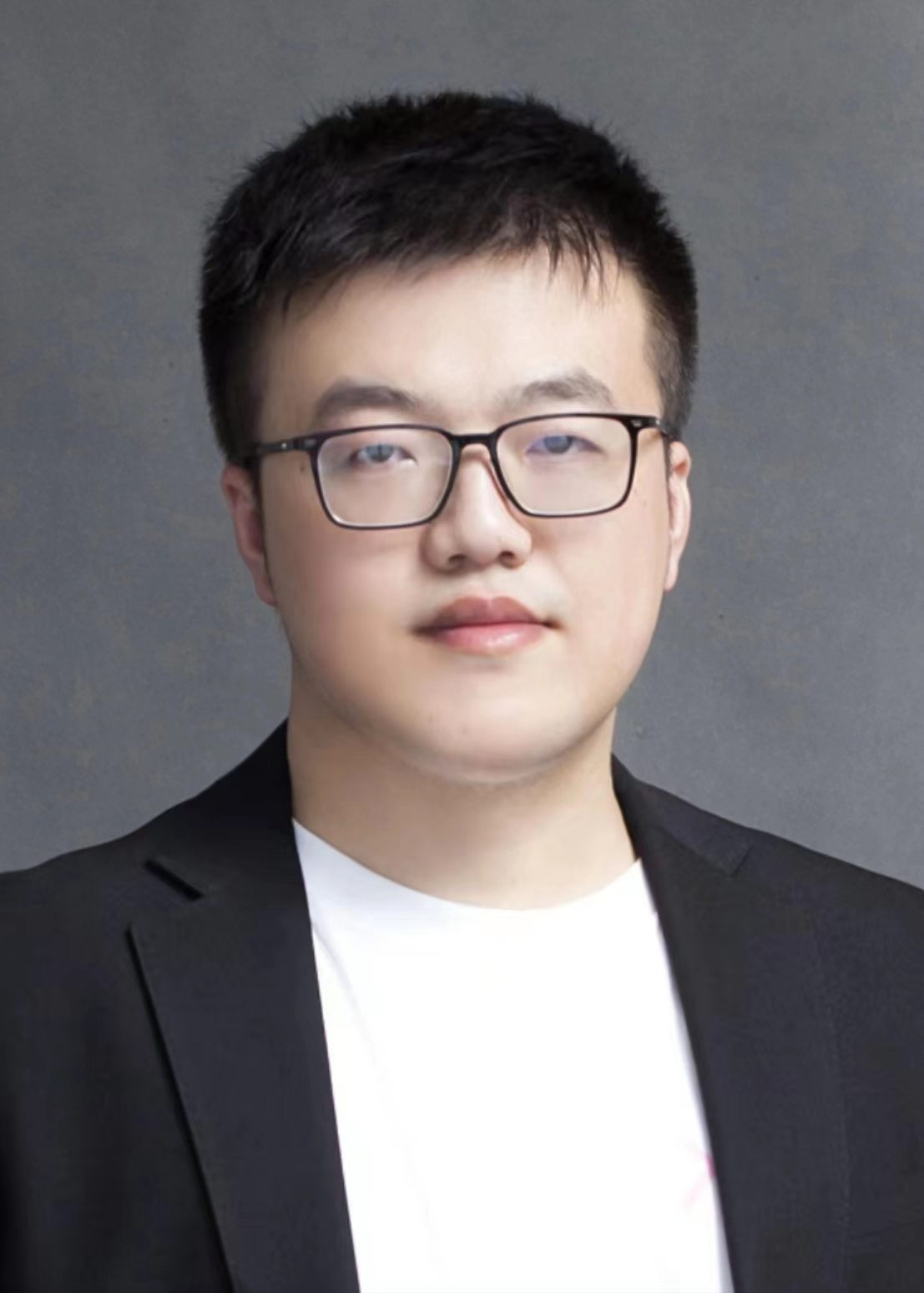}}]{Tuo Chen} is a PhD student with
the Department of Electronic Information, Southeast University. He received his bachelor’s degree from the Department of Information Security, Lanzhou University. His main research interests include Self-supervised learning, representation learning, and adversarial robustness.
\end{IEEEbiography}

\begin{IEEEbiography}[{\includegraphics[width=1in,height=1.25in,clip,keepaspectratio]{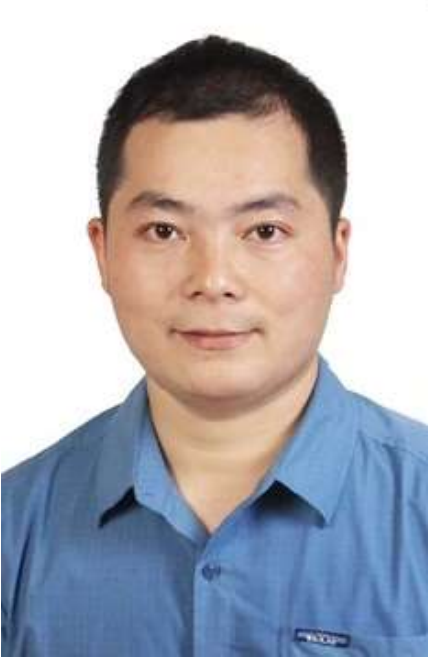}}]{Jing Zhang} (Senior Member, IEEE) is currently a Research Fellow at the School of Computer Science, The University of Sydney. He has authored over 80 papers in prestigious conferences and journals, including CVPR, ICCV, ECCV, NeurIPS, ICLR, IEEE TPAMI, and IJCV. His research focuses on computer vision and deep learning. Additionally, he is an Area Chair for ICPR, a Senior Program Committee member for AAAI and IJCAI, and a guest editor for IEEE TBD. He regularly reviews for numerous prestigious journals and conferences.
\end{IEEEbiography}

\begin{IEEEbiography}[{\includegraphics[width=1in,height=1.25in,clip,keepaspectratio]{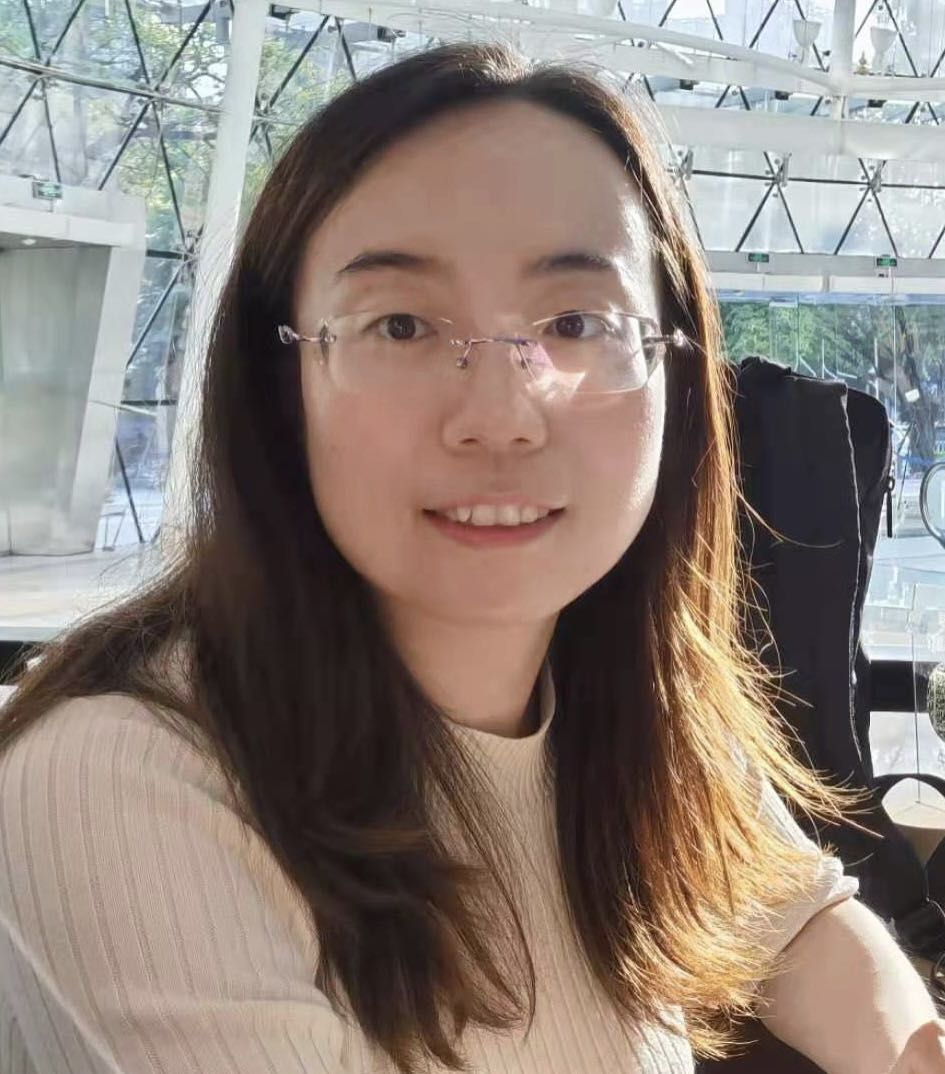}}]{Qiong Cao} is a Research Scientist at JD Explore Academy. Before that, she was a Senior Researcher at Tencent. Prior to joining Tencent, she was a Postdoctoral Researcher at the Department of Engineering Science, University of Oxford. She obtained her PhD in Computer Science from the University of Exeter.
\end{IEEEbiography}

\begin{IEEEbiography}[{\includegraphics[width=1in,height=1.25in,clip,keepaspectratio]{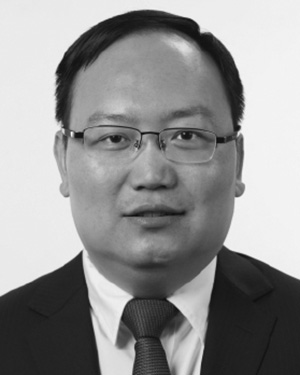}}]{Zhenan Sun} (SM'18) received a B.E. degree in industrial automation from Dalian University of Technology, Dalian, China, in 1999, an M.S. degree in system engineering from Huazhong University of Science and Technology, Wuhan, China, in 2002, and a PhD degree in pattern recognition and intelligent systems from the Institute of Automation, Chinese Academy of Sciences (CASIA), Beijing, China, in 2006.
  
Since 2006, he has been a Faculty Member with the National Laboratory of Pattern Recognition, CASIA, and he is currently a professor with the Center for Research on Intelligent Perception and Computing. He has authored/coauthored over 200 technical papers. His current research interests include biometrics, pattern recognition, and CV.
  
Prof. Sun is an Associate Editor of IEEE Transactions on Biometrics, Behavior, and Identity Science. He is a member of the IEEE Computer Society and IEEE Signal Processing Society and a fellow of IAPR.
  \end{IEEEbiography}

\begin{IEEEbiography}[{\includegraphics[width=1in,height=1.25in,clip,keepaspectratio]{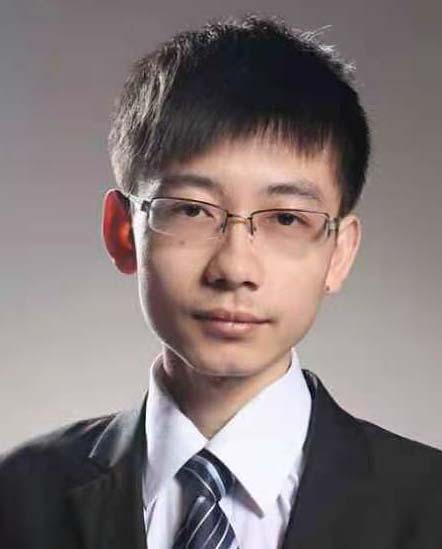}}]{Hao Luo} received B.S. and PhD degrees from Zhejiang University, China, in 2015 and 2020, respectively. He is currently working at the Alibaba DAMO Academy. His research interests include person re-identification, vision transformer, self-supervised, computer vision, and deep learning.
\end{IEEEbiography}

\begin{IEEEbiography}[{\includegraphics[width=1in,height=1.25in,clip,keepaspectratio]{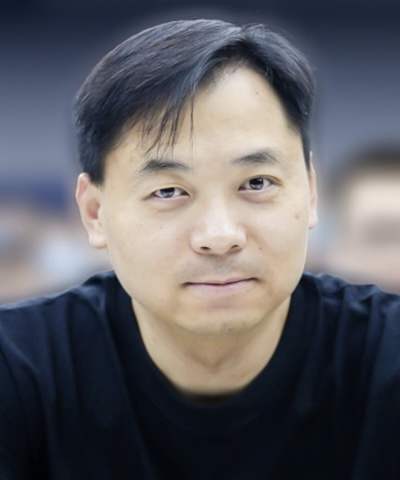}}]{Dacheng Tao} (Fellow, IEEE) is currently a Distinguished University Professor in the College of Computing \& Data Science at Nanyang Technological University. He mainly applies statistics and mathematics to artificial intelligence and data science, and his research is detailed in one monograph and over 200 publications in prestigious journals and proceedings at leading conferences, with best paper awards, best student paper awards, and test-of-time awards. His publications have been cited over 112K times and he has an h-index 160+ in Google Scholar. He received the 2015 and 2020 Australian Eureka Prize, the 2018 IEEE ICDM Research Contributions Award, and the 2021 IEEE Computer Society McCluskey Technical Achievement Award. He is a Fellow of the Australian Academy of Science, AAAS, ACM and IEEE.
\end{IEEEbiography}




\end{document}